\documentclass[10pt]{article} 
\usepackage[accepted]{tmlr}

\usepackage{url}
\usepackage[
    colorlinks=true,
    breaklinks=true,
    linkcolor=green!50!black,
    citecolor=green!50!black,
    urlcolor=green!50!black,
    pdfborder={0 0 0}
]{hyperref}

\usepackage{amsmath}
\usepackage{amsfonts}
\usepackage{amssymb}
\usepackage{mathtools}
\usepackage{commath}
\usepackage{bbm}

\usepackage{algorithm}
\usepackage{algpseudocode}
\usepackage{array}

\usepackage{varwidth}
\usepackage{setspace}

\usepackage[
    overcommands, 
    subscripts,
]{overarrows}

\usepackage{pgfplots}
\pgfplotsset{compat=1.18}
\usepgfplotslibrary{groupplots}
\usetikzlibrary{shapes.geometric}
\usepgfplotslibrary{fillbetween}

\DeclareMathOperator*{\argmin}{arg\,min}
\DeclareMathOperator*{\argmax}{arg\,max}
\DeclareMathOperator*{\minimize}{minimize}
\DeclareMathOperator*{\maximize}{maximize}
\DeclareMathOperator*{\subject}{subject\,\,to}
\DeclareMathOperator*{\const}{const}
\DeclareMathOperator*{\tr}{tr}

\DeclareMathOperator*{\given}{\,|\,}
\DeclareMathOperator*{\ggiven}{\,||\,}

\DeclareMathOperator*{\T}{\scalebox{0.65}{$T$}}
\DeclareMathOperator*{\sT}{\scalebox{0.5}{$T$}}

\NewOverArrowCommand{rh}{%
    start=\relbar,
    end=\rightharpoonup,
    detect subscripts=true,
    center arrow,
    shift leftright=-2,
}

\NewOverArrowCommand{lh}{%
    start=\leftharpoonup,
    end=\relbar,
    detect subscripts=true,
    center arrow,
    shift leftright=-2,
}

\newcommand{\rhk}[2]{\ensuremath \rh{#1}_{#2}}
\newcommand{\lhk}[2]{\ensuremath \lh{#1}_{#2}}

\newcommand{\rhi}[2]{\ensuremath \rh{#1}^{[#2]}}
\newcommand{\lhi}[2]{\ensuremath \lh{#1}^{[#2]}}

\newcommand{\rhix}[2]{\ensuremath \rh{#1}^{\,[#2]}}
\newcommand{\lhix}[2]{\ensuremath \lh{#1}^{\,[#2]}}

\newcommand{\rhki}[3]{\ensuremath \rh{#1}_{#2}^{\,\,[#3]}}
\newcommand{\lhki}[3]{\ensuremath \lh{#1}_{#2}^{\,\,[#3]}}

\newcommand{\rhkx}[2]{\ensuremath \rh{#1}_{\!#2}}
\newcommand{\lhkx}[2]{\ensuremath \lh{#1}_{\!#2}}

\newcommand{\rhkix}[3]{\ensuremath \rh{#1}_{\!#2}^{\,[#3]}}
\newcommand{\lhkix}[3]{\ensuremath \lh{#1}_{\!#2}^{\,[#3]}}

\newcommand{\rhkxx}[2]{\ensuremath \rh{#1}_{\!\!#2}}
\newcommand{\lhkxx}[2]{\ensuremath \lh{#1}_{\!\!#2}}

\newcommand{\rhkixx}[3]{\ensuremath \rh{#1}_{\!\!#2}^{\,[#3]}}
\newcommand{\lhkixx}[3]{\ensuremath \lh{#1}_{\!\!#2}^{\,[#3]}}

\newtheorem{proposition}{Proposition}
\newtheorem{lemma}{Lemma}
\newtheorem{corollary}{Corollary}
\newtheorem{definition}{Definition}
\newtheorem{assumption}{Assumption}
\newtheorem{remark}{Remark}

\allowdisplaybreaks
\mathtoolsset{showonlyrefs}

\title{Recursive Entropic Variational Inference \\ for Nonlinear State-Space Models}

\author{\name Hany Abdulsamad \email h.abdulsamad@uva.nl \\
        \addr Amsterdam Machine Learning Lab \\
        University of Amsterdam, Netherlands
        \AND
        \name \'Angel F. Garc\'ia-Fern\'andez \email angel.garcia.fernandez@upm.es \\
        \addr Information Processing and Telecommunications Center \\
        Universidad Polit\'ecnica de Madrid, Spain
        \AND
        \name Simo S\"arkk\"a \email simo.sarkka@aalto.fi \\
        \addr ELLIS Institute Finland \\
        \addr Department of Electrical Engineering and Automation \\
        Aalto University, Finland
}

\def\month{11}
\def\year{2026}
\def\openreview{\url{https://openreview.net/forum?id=Pt3zdEdg5y}}

\begin{document}

\maketitle

\begin{abstract}
We present a class of algorithms for state estimation in nonlinear, non-Gaussian state-space models. Our approach is based on a variational Lagrangian formulation that casts Bayesian inference as a sequence of entropic trust-region updates subject to dynamic consistency constraints. This framework gives rise to a family of forward-backward algorithms whose structure is determined by the chosen factorization of the variational posterior. By focusing on Gauss--Markov approximations, we derive recursive schemes with favorable computational complexity. For general nonlinear, non-Gaussian models, we close the recursions using generalized statistical linear regression and Fourier--Hermite moment matching.
\end{abstract}

\section{Introduction}

Accurate state estimation in partially observable dynamical phenomena is a fundamental problem across scientific and engineering disciplines, including robotics~\citep{barfoot2024state}, economics~\citep{jacquier2002bayesian}, and biology~\citep{Murray2002}. Bayesian inference in state-space models (SSMs) provides a principled framework, including techniques such as Kalman filtering, sequential Monte Carlo, and message passing~\citep{sarkka2023bayesian}. Additionally, an important link between inference and optimization was established early in the field~\citep{kalman1960new, cox1964estimation, bell1994iterated}, giving rise to a suite of inference-as-optimization algorithms. More recently, this relationship has been deepened by approximate inference methods, such as variational inference \citep[VI,][]{wainwright2008graphical} and expectation propagation \citep[EP,][]{minka2001expectation}, which extend the optimization perspective beyond point estimates to structured posterior approximations.

For state-space models with conjugate prior-likelihood pairs, the corresponding Bayesian posterior distributions admit closed-form solutions via efficient recursive algorithms, such as the Rauch--Tung--Striebel (RTS) smoother for linear-Gaussian SSMs~\citep{rauch1965maximum}. However, general nonlinear, non-Gaussian models do not usually admit closed-form recursive solutions. While sampling-based methods such as sequential Monte Carlo~\citep{chopin2020intro} and Markov chain Monte Carlo~\citep{brooks2011handbook} provide asymptotically exact solutions, they do so at the price of high computational complexity and limited scalability, particularly for sequences in high-dimensional spaces and long time horizons~\citep{beskos2014stability}. Approximate Bayesian techniques, though often suboptimal, offer a trade-off that sacrifices exactness for computational efficiency. This trade-off motivates our focus on this class of methods.

Variational inference and expectation propagation methods have been adapted for Bayesian inference in state-space models~\citep{deisenroth2012expectation, chang2020fast, wilkinson2020state}. However, despite the strong connection of both frameworks to numerical optimization~\citep{wilkinson2023bayes}, many existing approaches extend VI and EP to state-space representations in an \emph{ad hoc} manner. Such methods often incorporate variational principles into the structure of the Rauch--Tung--Striebel smoother template rather than deriving recursive algorithms from an explicit dynamic optimization problem. As a result, they may not fully leverage the potential benefits of a systematic approach to state estimation.

In this work, we cast approximate Bayesian inference in nonlinear, non-Gaussian state-space models as an iterative \emph{dynamic} optimization problem over the space of candidate posteriors. Our key contributions are:
\begin{itemize}
    \setlength{\itemsep}{1.5pt}
    \item \emph{Dynamic optimization formulation}: We introduce a principled variational framework for Bayesian inference that formulates inference as a sequence of entropic trust-region updates in probability-density space, using a structured proximal regularizer.
    \item \emph{Unified recursive algorithms}: We derive a family of efficient forward-backward recursion schemes whose structure is determined entirely by the factorization of the variational posterior --- forward-, reverse-, and hybrid-Markov factorizations.
    \item \emph{Adaptive step size}: We show that the step size between iterative updates corresponds to a Lagrangian multiplier whose optimal value is determined by a dual R\'enyi divergence objective.
    \item \emph{Generalization of classical smoothers}: The proposed approach generalizes algorithms in the linear-Gaussian setting, providing a bridge between classical and modern approximate inference techniques.
    \item \emph{Flexible posterior approximations}: We provide practical realizations using Gauss--Markov approximations, instantiated via generalized statistical linear regression and Fourier--Hermite moment matching, allowing for efficient inference in nonlinear and non-Gaussian models.
\end{itemize}

The paper is organized as follows. Section~\ref{sec:problem-statement} introduces the approximate inference problem in state-space models. Section~\ref{sec:proximal-optimization} provides a brief review of proximal variational optimization, which serves as the foundation for our method. In Section~\ref{sec:proximal-smoothing}, we present our dynamic Lagrangian framework for approximate Bayesian inference. Section~\ref{sec:practical-algorithms} specializes this framework to the Gaussian setting and derives practical recursive algorithms. We then connect our approach to related work in Section~\ref{sec:related-work}. Finally, Section~\ref{sec:experiments} provides a numerical evaluation of the proposed algorithms. Proofs are deferred to the appendix.

\section{Problem Statement}
\label{sec:problem-statement}

We consider the problem of Bayesian state estimation in nonlinear, non-Gaussian state-space models. Let $\{x_k \in \mathbb{R}^{d}\}_{k=0}^{\T}$ be a latent discrete-time Markov process and $\{y_k \in \mathbb{R}^{m}\}_{k=1}^{\T}$ a corresponding sequence of noisy observations. The system is governed by the state-space dynamics:
\begin{equation}
    \label{eq:state-space-model}
    x_{0} \sim p_{0}(\cdot), \qquad x_{k+1} \sim f_{k}(\cdot \given x_{k}), \qquad y_{k+1} \sim h_{k+1}(\cdot \given x_{k+1}),
\end{equation}
where $p_{0}(x_{0})$ is the initial prior probability density, $f_{k}(x_{k+1} \given x_{k})$ is the transition density of the latent Markov dynamics, and $h_{k}(y_{k} \given x_{k})$ is the conditional likelihood of the measurements. State estimation refers to the problems of online filtering and offline smoothing, which aim to reconstruct the posterior distributions $p(x_{k} \given y_{1:k})$ and $p(x_{k} \given y_{1:\T})$, for all $0 \leq k \leq T$, respectively. For arbitrary transition densities $f_{k}(x_{k+1} \given x_{k})$ and measurement likelihoods $h_{k}(y_{k} \given x_{k})$, these posterior densities are generally intractable.

In this work, we focus on finding the best approximation $q(x_{0:\T} \given y_{1:\T})$ of the smoothing posterior density within a certain variational family of joint distributions
\begin{equation}
    q(x_{0:\T} \given y_{1:\T}) \approx p(x_{0:\T} \given y_{1:\T}) \propto p_{0}(x_{0}) \prod_{k=0}^{T-1} f_{k}(x_{k+1} \given x_{k}) \, h_{k+1}(y_{k+1} \given x_{k+1}).
\end{equation}
More specifically, to leverage the temporal structure of state-space models, we consider structured Markov approximations of the smoothing posterior, leading to the following possibilities:

\begin{assumption}[Forward-Markov factorization]
    \label{asm:forward-markov}
    For state-space models of the form~\eqref{eq:state-space-model}, we restrict the approximate smoothing posterior to the forward-Markov variational family
    \begin{equation}
        \rh{q}(x_{0:\T} \given y_{1:\T}) = \rhk{q}{0}(x_{0} \given y_{1:\T}) \prod_{k=0}^{\T-1} \rhk{q}{k}(x_{k+1} \given x_{k}, y_{k+1:\T}).
    \end{equation}
\end{assumption}

\begin{assumption}[Reverse-Markov factorization]
    \label{asm:reverse-markov}
    For state-space models of the form~\eqref{eq:state-space-model}, we restrict the approximate smoothing posterior to the reverse-Markov variational family
    \begin{equation}
        \lh{q}(x_{0:\T} \given y_{1:\T}) = \lhk{q}{\T}(x_{\T} \given y_{1:\T}) \prod_{k=1}^{\T} \lhk{q}{k}(x_{k-1} \given x_{k}, y_{1:k-1}).
    \end{equation}
\end{assumption}
We use $\overrightharpoonup{q}$ and $\overleftharpoonup{q}$ to distinguish forward- and reverse-Markov quantities, respectively. Although Assumption~\ref{asm:forward-markov} and Assumption~\ref{asm:reverse-markov} explicitly state the dependence of the approximate conditionals on the measurements, we omit this dependence in subsequent sections for notational simplicity.

\section{Entropic Proximal Variational Optimization}
\label{sec:proximal-optimization}

Before introducing the proposed approach to variational smoothing in state-space models, we briefly review the principles of entropic proximal variational optimization in a simpler, static setting. Consider a hidden variable $x \in \mathbb{R}^{d}$, observed data $y \in \mathbb{R}^{m}$, a likelihood $p(y \given x)$, and a prior $p(x)$. The goal is to approximate the posterior $p(x \given y)$ by a distribution $q$, belonging to a variational family of probability distributions $\mathcal{Q}$, that minimizes the Kullback--Leibler (KL) divergence:
\begin{equation}
    q^\star(x) = \argmin_{q \in \mathcal{Q}} \,\, \mathbb{D}_{\mathrm{KL}} \left[ q(x) \ggiven p(x \given y) \right] = \argmin_{q \in \mathcal{Q}} \,\, \mathbb{E}_{q} \left[ \log \frac{q(x)}{p(x \given y)} \right].
\end{equation}
Equivalently, since the marginal likelihood $p(y)$ does not depend on $q(x)$, this problem can be written as the maximization of the evidence lower bound (ELBO)~\citep{blei2017variational}:
\begin{equation}\label{eq:static-elbo}
    q^\star(x) = \argmax_{q\in\mathcal{Q}} \,\, \mathcal{L}(q) \coloneqq \mathbb{E}_{q} \Big[ \log p(x, y) \Big] - \mathbb{E}_{q} \Big[ \log q(x) \Big] \leq \log p(y),
\end{equation}
where $p(y)$ is the marginal likelihood or evidence. In conjugate models, the ELBO can be optimized exactly~\citep{bishop2006pattern}, but for more general settings, approximate solutions are required. Black-box variational inference~\citep{ranganath2014black} provides a flexible framework for such cases. However, alternative methods such as natural and Riemannian gradient approaches~\citep{honkela2010approximate} and proximal variational inference~\citep{chretien2002kullback, khan2015proximal, theis2015trust} explicitly account for the geometry of the variational family and often lead to more stable and efficient optimization~\citep{amari1998natural}.

In this work, we focus on the entropic proximal variational optimization framework to derive recursive algorithms for structured approximate inference in nonlinear non-Gaussian state-space models. However, before directing our attention to the case of state-space models, we use the static case in~\eqref{eq:static-elbo} to illustrate the mechanics of entropic proximal variational optimization. We start by formulating the approximate inference problem within the iterative entropic proximal optimization framework proposed by~\citet{teboulle1992entropic, iusem1994entropy}. Starting from the ELBO in \eqref{eq:static-elbo}, we introduce an entropic-proximal constraint on the subsequent posterior iterate in the form of a Kullback--Leibler divergence, leading to the following nonlinear program:
\begin{equation}
    \label{eq:static-proximal}
    \begin{aligned}
        \maximize_{q(x)} & \quad \mathbb{E}_{q} \Big[ \log p(x, y) \Big] - \mathbb{E}_{q} \Big[ \log q(x) \Big], \\
        \subject & \quad \mathbb{D}_{\mathrm{KL}} \left[ q(x) \ggiven q^{[i]}(x) \right] \leq \varepsilon \quad \text{and} \quad \int q(x) \dif x = 1,
    \end{aligned}
\end{equation}
where $q^{[i]}(x)$ is the approximate posterior at iteration $i$ and $\varepsilon \geq 0$ is a hyperparameter that controls the information bottleneck between the current iterate $q^{[i]}(x)$ and the solution $q^{[i+1]}(x)$. Additionally, problem~\eqref{eq:static-proximal} includes a distributional normalization constraint for $q^{[i+1]}(x)$, whereas the positivity of $q^{[i+1]}(x)$ is implied by the logarithmic function embedded within the KL constraint. The following proposition derives the solution to~\eqref{eq:static-proximal}, the variational distribution iterate $q^{[i+1]}(x)$, by constructing the Lagrangian. Throughout, we assume that the target posterior is absolutely continuous with respect to the current iterate.

\begin{proposition}[Damped Gibbs posterior]
    \label{prop:static-proximal}
    The solution to the constrained nonlinear program~\eqref{eq:static-proximal}, corresponding to a Bayesian model with likelihood $p(y \given x)$ and prior $p(x)$, is given by the Gibbs posterior
    \begin{equation}
        q^{[i+1]}(x) = \left[ \mathcal{Z}^{[i+1]}(\beta) \right]^{-1} \Big[ p(y \given x) \, p(x) \Big]^{1 - \beta} \Big[ q^{[i]}(x) \Big]^{\beta},
    \end{equation}
    with a damping parameter $\beta \in [0, 1)$ and a normalizing constant
    \begin{equation}
        \mathcal{Z}^{[i+1]}(\beta) = \int \Big[ p(y \given x) \, p(x) \Big]^{1 - \beta} \Big[ q^{[i]}(x) \Big]^{\beta} \dif x.
    \end{equation}
    The damping parameter $\beta$ is a reparameterization of the Lagrangian multiplier $\alpha \geq 0$ associated with the Kullback--Leibler divergence constraint in~\eqref{eq:static-proximal}, so that $ \beta = \alpha / (1 + \alpha)$. Furthermore, the optimal $\beta$ is a minimizer of the dual problem
    \begin{equation}
        \label{eq:static-proximal-dual}
        \minimize_{\beta} \,\, \mathcal{G}(\beta) =  \frac{\beta \varepsilon}{1 - \beta} + \frac{1}{1 - \beta} \log \mathcal{Z}^{[i+1]}(\beta), \quad \subject \,\, 0 \leq \beta < 1.
    \end{equation}
\end{proposition}
\textit{Proof.} See Appendix~\ref{app:proof-static-proximal}.

\begin{remark}
    The posterior iterate $q^{[i+1]}(x)$ can be interpreted as a geometric interpolation in the space of probability densities between the exact posterior $p(x \given y) \propto p(y \given x) \, p(x)$ and the previous iterate $q^{[i]}(x)$. As $\beta \to 0$, the update recovers the exact posterior, whereas as $\beta \to 1$, the update approaches the previous iterate $q^{[i]}(x)$ and corresponds to an infinitesimal proximal update.
\end{remark}

\begin{remark}
    The dual objective over $\beta$ in~\eqref{eq:static-proximal-dual} can be interpreted through the
    variational R\'enyi bound of~\citet{li2016renyi}:
    \begin{equation}
        \frac{1}{1 - \beta} \log \mathcal{Z}^{[i+1]}(\beta) = \log p(y) - \mathbb{D}_{\beta} \left[ q^{[i]}(x) \ggiven p(x \given y) \right],
    \end{equation}
    where $\mathbb{D}_{\beta} \left[ \cdot \ggiven \cdot \right]$ denotes the R\'enyi divergence with order $\beta$. As the damping parameter $\beta$ varies, the corresponding R\'enyi divergence traces a continuum of $\alpha$-divergence geometries, each emphasizing different regions of the posterior distribution. Thus, varying $\beta$ smoothly alters the global geometric structure of the optimization problem but preserves the local notion of distance defined by the Fisher information metric induced by the primal Kullback--Leibler divergence objective~\citep{amari2016information}.
\end{remark}
Having reviewed the principles of entropic proximal optimization, we now turn to state-space models. In the next section, we extend this framework into the dynamic setting and derive recursive algorithms for approximate Bayesian inference that exploit the structure of the assumed posterior.

\section{Entropic Proximal Bayesian Smoothing}
\label{sec:proximal-smoothing}

For a state-space model~\eqref{eq:state-space-model}, we adapt the ELBO from~\eqref{eq:static-elbo} to the joint smoothing distribution as follows
\begin{equation}\label{eq:smoothing-elbo}
    \mathcal{L}(q) = \mathbb{E}_{q} \Big[ \log p(x_{0:\T}, y_{1:\T}) \Big] - \mathbb{E}_{q} \Big[ \log q(x_{0:\T}) \Big] \leq \log p(y_{1:\T}),
\end{equation}
where $p(x_{0:\T}, y_{1:\T})$ is the joint state-measurement distribution given by
\begin{equation}
    p(x_{0:\T}, y_{1:\T}) = p_{0}(x_{0}) \prod_{k=0}^{\T-1} f_{k}(x_{k+1} \given x_{k}) \, h_{k+1}(y_{k+1} \given x_{k+1}).
\end{equation}
We now formulate the entropic proximal optimization problem over the joint smoothing posterior $q(x_{0:\T})$:
\begin{equation}
    \label{eq:smoothing-proximal}
    \begin{aligned}[t]
        \maximize_{q\in\mathcal{Q}} & \quad \mathbb{E}_{q} \Big[ \log p(x_{0:\T}, y_{1:\T}) \Big] - \mathbb{E}_{q} \Big[ \log q(x_{0:\T}) \Big], \\
        \subject & \quad \mathbb{D}_{\mathrm{KL}} \left[ q(x_{0:\T}) \ggiven q^{[i]}(x_{0:\T}) \right] \leq \varepsilon \quad \text{and} \quad \int q(x_{0:\T}) \dif x_{0:\T} = 1.
    \end{aligned}
\end{equation}
This formulation does not yet impose any assumptions on the structure of the approximate posterior $q(x_{0:\T})$. While \eqref{eq:smoothing-proximal} can be solved in a manner similar to \eqref{eq:static-proximal}, such an approach can lead to significant computational complexity due to the high dimensionality of $q(x_{0:\T})$ as it extends over the state and time dimensions. To address this challenge, we introduce variations to \eqref{eq:smoothing-proximal} that take advantage of sparsity induced by the forward- and reverse Markov structure from Assumption~\ref{asm:forward-markov} and Assumption~\ref{asm:reverse-markov}, enabling recursive inference algorithms with linear time complexity in the horizon $T$.

\subsection{Damped Forward-Markov Posterior}
\label{sec:forward-markov}

Here, we restrict the admissible posterior family in~\eqref{eq:smoothing-proximal} to the forward-Markov decomposition stated in Assumption~\ref{asm:forward-markov}. Under this factorization, the joint smoothing posterior is represented by an initial marginal density and a sequence of forward transition conditionals. The entropic proximal update can then be decomposed into a recursive message-passing scheme, where a backward recursion first accumulates the contribution of future measurements into potential functions, and a subsequent forward recursion constructs the updated forward-Markov posterior and its marginal smoothing distributions. This leads to a generic \emph{backward-forward} algorithm for approximate Bayesian smoothing.

\begin{proposition}[Optimal forward-Markov posterior]
    \label{prop:forward-markov}
    For state-space models of the form~\eqref{eq:state-space-model}, and under Assumption~\ref{asm:forward-markov}, the approximate forward-Markov smoothing posterior that solves problem~\eqref{eq:smoothing-proximal} within the forward-Markov variational family is characterized by the following tilted distributions:
    \begin{align}
        \rhki{q}{0}{i+1}(x_{0}) & = \left[ \rhki{\mathcal{Z}}{0}{i+1} \right]^{-1} \Big[ \rhki{q}{0}{i}(x_{0}) \Big]^{\beta} \Big[ \exp \left\{ \rhkixx{V}{0}{i+1}(x_{0}) \right\} \Big]^{1 - \beta}, \label{eq:forward-markov-optimal-marginal} \\
        \rhki{q}{k}{i+1}(x_{k+1} \given x_{k}) & = 
        \begin{aligned}[t] \label{eq:forward-markov-optimal-conditionals}
            \left[ \rhki{\psi}{k}{i+1}(x_{k}) \right]^{-1} & \Big[ \rhki{q}{k}{i}(x_{k+1} \given x_{k}) \Big]^{\beta} \\ 
            & \times \Big[ f_{k}(x_{k+1} \given x_{k}) \exp \left\{ \rhkixx{V}{k+1}{i+1}(x_{k+1}) \right\} \Big]^{1 - \beta},
        \end{aligned}
        \quad 0 \leq k < T,
    \end{align}
    where $\beta \in [0, 1)$ is the damping parameter associated with the Lagrangian multiplier $\alpha \geq 0$, so that $\beta = \alpha / (1 + \alpha)$. The quantities $\rhki{\mathcal{Z}}{0}{i+1}$ and $\rhki{\psi}{k}{i+1}(x_{k})$ are the corresponding normalizing factors
    \begin{equation}
        \label{eq:forward-markov-normalizers}
        \begin{aligned}[t]
            \rhki{\mathcal{Z}}{0}{i+1} & = \int \Big[ \rhki{q}{0}{i}(x_{0}) \Big]^{\beta} \Big[  \exp \left\{ \rhkixx{V}{0}{i+1}(x_{0}) \right\} \Big]^{1 - \beta} \dif x_{0}, \\
            \rhki{\psi}{k}{i+1}(x_{k}) & = \int \Big[ \rhki{q}{k}{i}(x_{k+1} \given x_{k}) \Big]^{\beta} \Big[ f_{k}(x_{k+1} \given x_{k}) \exp \left\{ \rhkixx{V}{k+1}{i+1}(x_{k+1}) \right\} \Big]^{1 - \beta} \dif x_{k+1}, \quad 0 \leq k < T.
        \end{aligned}
    \end{equation}
    The potential functions $\rhkixx{V}{k}{i+1}(x_{k})$, for all $0 \leq k \leq T$, are computed recursively backwards via
    \begin{equation}
        \label{eq:forward-markov-potentials}
        \rhkixx{V}{k}{i+1}(x_{k}) = 
        \begin{cases}
            \, \log h_{\T}(y_{\T} \given x_{\T}) & \mathrm{if} \,\, k = T, \\[0.5em]
            \, \log h_{k}(y_{k} \given x_{k}) + 1 / (1 - \beta) \log \rhki{\psi}{k}{i+1}(x_{k}) & \mathrm{if} \,\, 0 < k < T, \\[0.5em]
            \, \log p_{0}(x_{0}) + 1 / (1 - \beta) \log \rhki{\psi}{0}{i+1}(x_{0}) & \,\, \mathrm{if} \,\, k = 0.
        \end{cases}
    \end{equation}
    Here, each $\rhki{\psi}{k}{i+1}(x_{k})$ depends on the next potential $\rhkixx{V}{k+1}{i+1}(x_{k+1})$, so that the recursion is initialized at $k=T$ and propagated backwards to $k=0$. Finally, the optimal damping $\beta$ is the minimizer of the dual objective
    \begin{equation}
        \label{eq:forward-markov-dual}
        \minimize_{\beta} \,\, \rh{\mathcal{G}}(\beta) = \frac{\beta \varepsilon}{1 - \beta} + \frac{1}{1 - \beta} \log \rhki{\mathcal{Z}}{0}{i+1}(\beta), \quad \subject \,\, 0 \leq \beta < 1.
    \end{equation}
\end{proposition}
\textit{Proof.} See Appendix~\ref{app:forward-markov-proof}.

\begin{remark}
    \label{rem:forward-markov-potentials}
    The potential functions $\rhkxx{V}{k}(x_{k})$ can be interpreted as damped log-space backward messages $\log q(y_{k:\T} \given x_k)$, that accumulate the contribution of the future measurements $y_{k:\T}$ conditioned on $x_{k}$, while the log-normalizers $\log \rhk{\psi}{k}(x_{k})$ and $\log \rhk{\mathcal{Z}}{0}$ correspond to $\log q(y_{k+1:\T} \given x_{k})$ and $\log q(y_{1:\T})$, respectively.
\end{remark}

\begin{remark}
    \label{rem:forward-markov-marginals}
    Under Assumption~\ref{asm:forward-markov}, the marginal smoothing distributions $\rhki{q}{k+1}{i+1}(x_{k+1})$, for all $0 \leq k < T$, are computed via forward propagation starting from $\rhki{q}{0}{i+1}(x_{0})$
    \begin{equation}
        \label{eq:forward-markov-marginals}
        \rhki{q}{k+1}{i+1}(x_{k+1}) = \int \rhki{q}{k}{i+1}(x_{k}) \, \rhki{q}{k}{i+1}(x_{k+1} \given x_{k}) \dif x_{k},
    \end{equation}
    where $\rhki{q}{k}{i+1}(x_{k+1} \given x_{k})$ and $\rhki{q}{0}{i+1}(x_{0})$ are given by Proposition~\ref{prop:forward-markov}.
\end{remark}

\subsection{Damped Reverse-Markov Posterior}
\label{sec:reverse-markov}

Next, we restrict the admissible posterior family in~\eqref{eq:smoothing-proximal} to the reverse-Markov decomposition stated in Assumption~\ref{asm:reverse-markov}. Under this factorization, the approximate joint smoothing posterior is parameterized by a terminal marginal density $\lh{q}_{\T}(x_{\T})$ and a sequence of reverse conditionals $\lhk{q}{k}(x_{k-1} \given x_{k})$. The proximal update can therefore be decomposed into local tilted updates for the terminal marginal and for each reverse conditional. These local updates are coupled through potential functions that accumulate information from the initial condition and past measurements. The resulting scheme first computes these potentials by a forward recursion and then propagates the updated posterior marginals backward in time. This yields a generic \emph{forward-backward} recursive algorithm for approximate Bayesian smoothing.

\begin{proposition}[Optimal reverse-Markov posterior]
    \label{prop:reverse-markov}
    For state-space models of the form~\eqref{eq:state-space-model}, and under Assumption~\ref{asm:reverse-markov}, the approximate reverse-Markov smoothing posterior that solves problem~\eqref{eq:smoothing-proximal} within the reverse-Markov variational family is characterized by the following tilted conditional and marginal distributions:
    \begin{align}
        \lhki{q}{k}{i+1}(x_{k-1} \given x_{k}) & = 
        \begin{aligned}[t] \label{eq:reverse-markov-optimal-conditionals} 
            \Big[ \lhki{\psi}{k}{i+1}(x_{k}) \Big]^{-1} & \Big[ \lhki{q}{k}{i}(x_{k-1} \given x_{k}) \Big]^{\beta} \\
            & \times \Big[ f_{k-1}(x_{k} \given x_{k-1}) \exp \left\{ \lhkixx{V}{k-1}{i+1}(x_{k-1}) \right\} \Big]^{1 - \beta},
        \end{aligned}
        \quad 0 < k \leq T, \\
        \lhki{q}{\T}{i+1}(x_{\T}) & = \Big[ \lhki{\mathcal{Z}}{\T}{i+1} \Big]^{-1} \Big[ \lhki{q}{\T}{i}(x_{\T}) \Big]^{\beta} \Big[ \exp \left\{ \lhkixx{V}{\T}{i+1}(x_{\T}) \right\} \Big]^{1 - \beta}, \label{eq:reverse-markov-optimal-marginal}
    \end{align}
    where $\beta \in [0, 1)$ is the damping parameter associated with the Lagrangian multiplier $\alpha \geq 0$, so that $\beta = \alpha / (1 + \alpha)$. The quantities $\lhki{\psi}{k}{i+1}(x_{k})$ and $\lhki{\mathcal{Z}}{\T}{i+1}$ are the corresponding normalizing factors
    \begin{equation}\label{eq:reverse-markov-normalizers}
        \begin{aligned}[t]
            \lhki{\psi}{k}{i+1}(x_{k}) & = \int \Big[ \lhki{q}{k}{i}(x_{k-1} \given x_{k}) \Big]^{\beta} \Big[ f_{k-1}(x_{k} \given x_{k-1}) \exp \left\{ \lhkixx{V}{k-1}{i+1}(x_{k-1}) \right\} \Big]^{1 - \beta} \! \dif x_{k-1}, \quad 0 < k \leq T, \\
            \lhki{\mathcal{Z}}{\T}{i+1} & = \int \Big[ \lhki{q}{\T}{i}(x_{\T}) \Big]^{\beta} \Big[ \exp \left\{ \lhkixx{V}{\T}{i+1}(x_{\T}) \right\} \Big]^{1 - \beta} \dif x_{\T}.
        \end{aligned}
    \end{equation}
    The potential functions $\lhkixx{V}{k}{i+1}(x_{k})$, for all $0 \leq k \leq T$, are computed recursively forward via
    \begin{equation}\label{eq:reverse-markov-potentials}
        \lhkixx{V}{k}{i+1}(x_{k}) =
        \begin{cases}
            \, \log p_{0}(x_{0}) & \mathrm{if} \,\, k = 0, \\[0.5em]
            \, \log h_{k}(y_{k} \given x_{k}) + 1 / (1 - \beta) \log \lhki{\psi}{k}{i+1}(x_{k}) & \mathrm{if} \,\, 0 < k \leq T.
        \end{cases}
    \end{equation}
    Here, each $\lhki{\psi}{k}{i+1}(x_{k})$ depends on the previous potential $\lhkixx{V}{k-1}{i+1}(x_{k-1})$, with the recursion initialized at $k=0$ and propagated forwards to $k=T$. Finally, the optimal damping $\beta$ is the minimizer of the dual objective
    \begin{equation}
        \label{eq:reverse-markov-dual}
        \minimize_{\beta} \quad \lh{\mathcal{G}}(\beta) = \frac{\beta \varepsilon}{1 - \beta} + \frac{1}{1 - \beta} \log \lhki{\mathcal{Z}}{\T}{i+1}(\beta), \quad \subject \,\, 0 \leq \beta < 1.
    \end{equation}
\end{proposition}
\textit{Proof.} See Appendix~\ref{app:reverse-markov-proof}.

\begin{remark}
    \label{rem:reverse-markov-potentials}
    The potential functions $\lhkxx{V}{k}(x_{k})$ can be interpreted as damped log-space forward messages $\log q(x_{k} \given y_{1:k})$, that accumulate the contribution of the initial condition and past measurements $y_{1:k}$ up to state $x_{k}$, while the log-normalizers $\log \lhk{\psi}{k}(x_{k})$ and $\log \lhk{\mathcal{Z}}{\T}$ correspond to $\log q(x_{k} \given y_{1:k-1})$ and $\log q(y_{1:\T})$, respectively.
\end{remark}

\begin{remark}
    \label{rem:reverse-markov-marginals}
    Under Assumption~\ref{asm:reverse-markov}, the marginal smoothing distributions $\lhki{q}{k-1}{i+1}(x_{k-1})$, for all $1 \leq k \leq T$, are computed via backward propagation starting from $\lhki{q}{\T}{i+1}(x_{\T})$
    \begin{equation}
        \label{eq:reverse-markov-marginals}
        \lhki{q}{k-1}{i+1}(x_{k-1}) = \int \lhki{q}{k}{i+1}(x_{k}) \, \lhki{q}{k}{i+1}(x_{k-1} \given x_{k}) \dif x_{k},
    \end{equation}
    where $\lhki{q}{k}{i+1}(x_{k-1} \given x_{k})$ and $\lhki{q}{\T}{i+1}(x_{\T})$ are given by Proposition~\ref{prop:reverse-markov}.
\end{remark}

\subsection{Hybrid Posterior Marginals}
\label{sec:hybrid-markov}

The recursions in Proposition~\ref{prop:forward-markov} and Proposition~\ref{prop:reverse-markov} arise from two complementary factorizations of the approximate smoothing posterior. The forward-Markov factorization in Proposition~\ref{prop:forward-markov} represents the joint trajectory through an initial marginal and forward transition conditionals, leading to a backward recursion for future-measurement potentials followed by forward marginal propagation. Conversely, the reverse-Markov factorization in Proposition~\ref{prop:reverse-markov} represents the same trajectory through a terminal marginal and reverse transition conditionals, leading to a forward recursion for past-measurement potentials followed by backward marginal propagation.

These two constructions need not be treated as mutually exclusive. When the forward- and reverse-Markov conditionals are associated with the same joint smoothing iterate $q^{[i]}(x_{0:\T} \given y_{1:\T})$, they provide two alternative approximations of the same posterior. In this case, the backward potential recursion from Proposition~\ref{prop:forward-markov} and the forward potential recursion from Proposition~\ref{prop:reverse-markov} can be combined at the marginal level. This yields a hybrid update in which each intermediate marginal receives information propagated from both temporal directions, while retaining the same entropic damping with respect to the previous marginal iterate.

\begin{corollary}[Optimal hybrid marginals]
    \label{cor:hybrid-markov}
    For state-space models of the form~\eqref{eq:state-space-model}, suppose that the forward-Markov conditionals $\rhki{q}{k}{i}(x_{k+1} \given x_{k})$ and reverse-Markov conditionals $\lhki{q}{k}{i}(x_{k-1} \given x_{k})$ are the corresponding conditionals of a common joint smoothing distribution $q^{[i]}(x_{0:\T} \given y_{1:\T})$, and therefore induce the same marginals $q_{k}^{[i]}(x_{k})$. Then, the updated marginals $q_{k}^{[i+1]}(x_{k})$ associated with problem~\eqref{eq:smoothing-proximal} can be computed by combining the backward recursion from Proposition~\ref{prop:forward-markov} with the forward recursion from Proposition~\ref{prop:reverse-markov}        
    \begin{equation}
        q_{k}^{[i+1]}(x_{k}) \propto
        \begin{cases}
            \, \Big[ q_{0}^{[i]}(x_{0}) \Big]^{\beta} \Big[ \exp \left\{ \rhkixx{V}{0}{i+1}(x_{0}) \right\} \Big]^{1 - \beta} & \mathrm{if} \,\, k = 0, \\[0.5em]
            \, \Big[ q_{k}^{[i]}(x_{k}) \Big]^{\beta} \Big[ \exp \left\{ \lhkixx{V}{k}{i+1}(x_{k}) \right\} \, \rhki{\psi}{k}{i+1}(x_{k}) \Big]^{1 - \beta} & \mathrm{if} \,\, 0 < k < T, \\[0.5em]
            \, \Big[ q_{\T}^{[i]}(x_{\T}) \Big]^{\beta} \Big[ \exp \left\{ \lhkixx{V}{\T}{i+1}(x_{\T}) \right\} \Big]^{1 - \beta} & \,\, \mathrm{if} \,\, k = T.
        \end{cases}
    \end{equation}
    For $0 < k < T$, the potential $\lhkixx{V}{k}{i+1}(x_{k})$ carries the forward-in-time filtering contribution accumulated by the reverse-Markov recursion, while the predictive $\rhki{\psi}{k}{i+1}(x_{k})$ carries the backward-in-time likelihood contribution accumulated by the forward-Markov recursion. Therefore, the hybrid update forms a damped geometric interpolation between the previous marginal $q_{k}^{[i]}(x_{k})$ and the product of the two directional message contributions. Equivalently, one may combine the reverse-Markov predictive $\lhki{\psi}{k}{i+1}(x_{k})$ with the forward-Markov potential $\rhkixx{V}{k}{i+1}(x_{k})$ to reach an equivalent construction of the marginals.    
\end{corollary}

The forward–backward recursion schemes derived in Proposition~\ref{prop:forward-markov}, Proposition~\ref{prop:reverse-markov}, and Corollary~\ref{cor:hybrid-markov} share a general algorithmic structure that does not rely on specific assumptions about the dynamics $f_{k}(x_{k+1} \given x_{k})$, the measurement model $h_{k}(y_{k} \given x_{k})$, or the forms of the approximate posteriors $\overrightharpoonup{q}(x_{0:\T})$ and $\overleftharpoonup{q}(x_{0:\T})$. In the sections that follow, we present concrete implementations of these schemes by adopting conditionally Gaussian approximations for the forward- and reverse-Markov posterior distributions.

\subsection{Optimal Damping Parameter}
\label{sec:optimal-damping}

The recursive posterior updates in Proposition~\ref{prop:forward-markov}, Proposition~\ref{prop:reverse-markov}, and Corollary~\ref{cor:hybrid-markov} depend on the damping parameter $\beta \in [0,1)$. This parameter is a reparameterization of the Lagrange multiplier $\alpha \geq 0$ associated with the Kullback--Leibler constraint, where $\beta=\alpha/(1+\alpha)$. At each iteration, the optimal damping parameter is obtained by minimizing the corresponding dual objective in~\eqref{eq:forward-markov-dual} or~\eqref{eq:reverse-markov-dual}. 

For a fixed multiplier $\alpha$, let $q_\alpha^{[i+1]}(x_{0:\T})$ denote the joint proximal posterior from Proposition~\ref{prop:forward-markov} or Proposition~\ref{prop:reverse-markov}. Evaluating the Lagrangian of~\eqref{eq:smoothing-proximal} at this posterior yields the reduced dual objective
\begin{equation}
    \mathcal{R}(\alpha)
    = \mathcal{L}(q^{[i+1]}_\alpha)
    + \alpha \left[ \varepsilon - \mathbb{D}_{\mathrm{KL}}\left[ q^{[i+1]}_\alpha \ggiven q^{[i]} \right] \right].
\end{equation}
Differentiating this reduced objective with respect to $\alpha$, we obtain 
\begin{equation}
    \label{eq:temperature-grad}
    \mathcal{R}'(\alpha) = \varepsilon - \mathbb{D}_{\mathrm{KL}}\left[ q^{[i+1]}_\alpha \ggiven q^{[i]} \right] \coloneq \Delta_{\mathrm{KL}}(\alpha).
\end{equation}
Hence, the derivative measures the slack in the trust-region constraint. If the constraint is active and $\alpha^\star>0$, then $\Delta_{\mathrm{KL}}(\alpha^\star)=0$ and the corresponding posterior lies on the boundary of the KL-ball. If the constraint is inactive, then the undamped update is feasible, so that $\alpha^\star=0$ and $\beta^\star=0$.

\section{Practical Recursive Inference Algorithms}
\label{sec:practical-algorithms}

To transform the recursive schemes introduced in Section~\ref{sec:forward-markov} and Section~\ref{sec:reverse-markov} into concrete and tractable algorithms, we now impose additional structure on the forward- and reverse-Markov posterior updates. The abstract recursions in Proposition~\ref{prop:forward-markov} and Proposition~\ref{prop:reverse-markov} characterize the optimal tilted conditionals in~\eqref{eq:forward-markov-optimal-conditionals} and~\eqref{eq:reverse-markov-optimal-conditionals}, together with the corresponding potential functions in~\eqref{eq:forward-markov-potentials} and~\eqref{eq:reverse-markov-potentials}. These objects define the exact proximal update within the chosen Markov factorization, but they generally involve nonlinear integral transformations that are not analytically tractable. In this section, we obtain practical recursive algorithms by restricting the approximate joint smoothing posterior to the Gaussian family and the Markov conditionals to the Gauss--Markov family. Under this restriction, the tilted updates can be represented through finite-dimensional parameters, and the potential functions can be propagated by closed-form recursions.

\begin{assumption}[Forward Gauss--Markov approximation]
    \label{asm:forward-gauss-markov}
    Given a state-space model~\eqref{eq:state-space-model} and the forward-Markov factorization in Assumption~\ref{asm:forward-markov}, we restrict the approximate posterior $\overrightharpoonup{q}(x_{0:\T} \given y_{1:\T})$ to the family of forward Gauss--Markov densities. Specifically, we use the parameterization
    \begin{equation}
        \rh{q}(x_{0:\T} \given y_{1:\T}) = \mathcal{N}(x_{0} \given \rhk{m}{0}, \rhkx{P}{0}) \prod_{k=0}^{\T-1} \mathcal{N}(x_{k+1} \given \rhkx{F}{k} \, x_{k} + \rhk{d}{k}, \rhk{\Sigma}{k}).
    \end{equation}
\end{assumption}

\begin{assumption}[Reverse Gauss--Markov approximation]
    \label{asm:reverse-gauss-markov}
    Given a state-space model~\eqref{eq:state-space-model} and the reverse-Markov factorization in Assumption~\ref{asm:reverse-markov}, we restrict the approximate posterior $\overleftharpoonup{q}(x_{0:\T} \given y_{1:\T})$ to the family of reverse Gauss--Markov densities. Specifically, we use the parameterization
    \begin{equation}
        \lh{q}(x_{0:\T} \given y_{1:\T}) = \mathcal{N}(x_{\T} \given \lhk{m}{\T}, \lhkx{P}{\T}) \prod_{k=1}^{\T} \mathcal{N}(x_{k-1} \given \lhkx{F}{k} \, x_{k} + \lhk{d}{k}, \lhk{\Sigma}{k}).
    \end{equation}
\end{assumption}
In the following sections, we develop tractable recursive inference schemes by constructing local quadratic approximations to the potential functions and deriving update rules that preserve the Gauss--Markov structure of the approximate posterior across iterations.

\subsection{Statistical Function Approximations}
\label{sec:statistical-approximations}

A key step in constructing efficient smoothing algorithms is to approximate the potential functions in~\eqref{eq:forward-markov-potentials} and~\eqref{eq:reverse-markov-potentials} with tractable forms that preserve the recursive structure and admit closed-form updates. We achieve this by introducing second-order \emph{statistical expansions} of the log-density functions associated with the latent dynamics $\log f_{k}(x_{k+1} \given x_{k})$ and the measurement model $\log h_{k}(y_{k} \given x_{k})$. 

Given the iterative nature of the proposed optimization procedure~\eqref{eq:smoothing-proximal}, it is natural to construct these expansions locally around the current iterate $q^{[i]}(x_{0:\T})$. This ensures that the approximation is tailored to the current posterior belief and captures relevant structure in the vicinity of the current iterate. Crucially, the KL-based trust-region constraint used in the entropic proximal updates controls the step size of each iteration. It prevents the distribution $q^{[i+1]}$ from deviating too far from $q^{[i]}$, thereby keeping the updates within the region where the local expansions are intended to be accurate. This interaction between local approximations and bounded updates stabilizes the optimization and maintains the fidelity of the recursive inference scheme~\citep{teboulle1992entropic, iusem1994entropy, chretien2002kullback}.

\begin{definition}[Statistical second-order expansion]
    \label{def:statistical-expansion}
    Let $z \sim \mathcal{N}(m, P)$ and let $g(z)$ be a twice-differentiable scalar function. A second-order statistical expansion of $g(z)$ with respect to the random variable $z \in \mathbb{R}^{d_z}$ takes a quadratic form $g(z) \approx - \frac{1}{2} \, z^{\top} U \, z + z^{\top} u + \eta$, where $U \in \mathbb{R}^{d_z \times d_z}$, $u \in \mathbb{R}^{d_z}$, and $\eta$ is a scalar. 
\end{definition}
Definition~\ref{def:statistical-expansion} specifies the structure but not the computation of the expansion parameters. We return to this in Sections~\ref{sec:generalized-regression} and \ref{sec:fourier-hermite}, where we describe two approximation strategies for computing $(U, u, \eta)$.

\begin{definition}[Quadratic expansion of log-densities]
    \label{def:statistical-expansion-log-densities}
    Let $(x_{k+1}, x_{k}) \sim q_{k}^{[i]}(x_{k+1}, x_{k})$. The statistical expansion of $\ell_{f}^{[i]}(x_{k+1}, x_{k}) \approx \log f_{k}(x_{k+1} \given x_{k})$, for all $0 \leq k < T$, is defined as
    \begin{equation}
        \label{eq:transition-quadratic-form}
        \ell_{f}^{[i]}(x_{k+1}, x_{k}) =
        \begin{aligned}[t]
            - \frac{1}{2} 
            \begin{bmatrix} x_{k+1}^{\top} & x_{k}^{\top} \end{bmatrix}
            \begin{bmatrix} 
                C_{\bar x \bar x, k}^{[i]} & - C_{\bar x x, k}^{[i]} \\[0.5em]
                - C_{x \bar x, k}^{[i]} & C_{xx, k}^{[i]}
            \end{bmatrix}
            \begin{bmatrix} x_{k+1} \\[0.75em] x_{k} \end{bmatrix} 
            + \begin{bmatrix} x_{k+1}^{\top} & x_{k}^{\top} \end{bmatrix}
            \begin{bmatrix} c_{\bar x, k}^{[i]} \\[0.5em] c_{x, k}^{[i]} \end{bmatrix} + \kappa_{k}^{[i]},
        \end{aligned}
    \end{equation}
    where $C_{\bar x \bar x,k}^{[i]}, C_{xx,k}^{[i]} \in \mathbb{R}^{d \times d}$, $C_{\bar x x,k}^{[i]}, C_{x \bar x,k}^{[i]} \in \mathbb{R}^{d \times d}$, $c_{\bar x,k}^{[i]}, c_{x,k}^{[i]} \in \mathbb{R}^{d}$, and $\kappa_k^{[i]} \in \mathbb{R}$. Further, for each $0 < k \leq T$, the statistical expansion of $\ell_h^{[i]}(x_k) \approx \log h_k(y_k \given x_k)$ is defined as
    \begin{equation}
        \label{eq:likelihood-quadratic-form}
        \ell_{h}^{[i]}(x_{k}) = - \frac{1}{2} x_{k}^{\top} L_{k}^{[i]} \, x_{k} + x_{k}^{\top} l_{k}^{[i]} + \nu_{k}^{[i]},
    \end{equation}
    where $L_k^{[i]} \in \mathbb{R}^{d \times d}$, $l_k^{[i]} \in \mathbb{R}^{d}$, and $\nu_k^{[i]} \in \mathbb{R}$. Finally, for $x_{0} \sim q_{0}^{[i]}(x_{0})$, the statistical expansion of $\ell_{p}^{[i]}(x_{0}) \approx \log p_{0}(x_{0})$ is defined as
    \begin{equation}
        \ell_{p}^{[i]}(x_{0}) = - \frac{1}{2} x_{0}^{\top} L_{0}^{[i]} \, x_{0} + x_{0}^{\top} l_{0}^{[i]} + \nu_{0}^{[i]},
    \end{equation}
    where $L_0^{[i]} \in \mathbb{R}^{d \times d}$, $l_0^{[i]} \in \mathbb{R}^{d}$, and $\nu_0^{[i]} \in \mathbb{R}$.
\end{definition}

\subsection{Recursive Quadratic Potentials}
\label{sec:quadratic-potentials}

In this section, we derive tractable schemes for computing the potential functions defined in~\eqref{eq:forward-markov-potentials} and~\eqref{eq:reverse-markov-potentials}, leveraging the statistical approximations introduced in Definition~\ref{def:statistical-expansion-log-densities}. We show that these approximations lead to tractable recursions over quadratic forms of $\rhkxx{V}{k}(x_{k})$ and $\lhkxx{V}{k}(x_{k})$ from Proposition~\ref{prop:forward-markov} and Proposition~\ref{prop:reverse-markov}. The use of quadratic potentials is consistent with the interpretations in Remark~\ref{rem:forward-markov-potentials} and Remark~\ref{rem:reverse-markov-potentials}: the forward-Markov potentials act as log-space backward messages, whereas the reverse-Markov potentials act as log-space forward messages. Under the Gaussian approximation, such messages are naturally parameterized by quadratic functions, which motivates the form adopted in our smoothing framework.

Before introducing these recursions, we first define the following parametric forms for the potential functions and the associated log-normalizing functions. Let
\begin{equation}
    \label{eq:potentials-quadratic-form}
    V_{k}^{[i+1]}(x_{k}) = - \frac{1}{2} x_{k}^{\top} {R}_{k}^{[i+1]} \, x_{k} + x_{k}^{\top} r_{k}^{[i+1]} + \rho_{k}^{[i+1]}, 
\end{equation}
where $R_{k}^{[i+1]} \in \mathbb{R}^{d \times d}$, $r_{k}^{[i+1]} \in \mathbb{R}^{d}$, and $\rho_{k}^{[i+1]} \in \mathbb{R}$. Similarly, we define
\begin{equation}
    \label{eq:normalizer-quadratic-form}
    \log \psi_{k}^{[i+1]}(x_{k}) = - \frac{1}{2} x_{k}^{\top} {S}_{k}^{[i+1]} \, x_{k} + x_{k}^{\top} s_{k}^{[i+1]} + \xi_{k}^{[i+1]},
\end{equation}
where $S_{k}^{[i+1]} \in \mathbb{R}^{d \times d}$, $s_{k}^{[i+1]} \in \mathbb{R}^{d}$, and $\xi_{k}^{[i+1]} \in \mathbb{R}$. These quadratic parameterizations enable efficient message-passing recursions while preserving the Gaussian structure of the approximate posterior.

\begin{proposition}[Recursive forward Gauss--Markov potentials]
    \label{prop:forward-gauss-markov-potentials}
    Let $\rhki{q}{k}{i}(x_{k+1} \given x_{k})$ be a forward Gauss--Markov conditional as defined in Assumption~\ref{asm:forward-gauss-markov}, and let $\ell_{f}^{[i]}(x_{k+1}, x_{k})$, $\ell_{h}^{[i]}(x_{k})$, and $\ell_{p}^{[i]}(x_{0})$ be the second-order approximations of the log-densities from Definition~\ref{def:statistical-expansion-log-densities}. Then, the potentials $\rhkixx{V}{k}{i+1}(x_{k})$ in~\eqref{eq:forward-markov-potentials} are quadratic functions of the form~\eqref{eq:potentials-quadratic-form}. The recursion is initialized at $k=T$ by
    \begin{equation}
        \rhki{R}{\T}{i+1} = L_{\T}^{[i]}, \qquad
        \rhki{r}{\T}{i+1} = l_{\T}^{[i]}.
    \end{equation}
    For $0 \leq k < T$, the backward recursion is given by
    \begin{equation}
        \label{eq:forward-gauss-markov-potential-functions}
        \begin{aligned}[t]
            \rhki{R}{k}{i+1} & = L_{k}^{[i]} + \frac{1}{1 - \beta} \, \rhkix{S}{k}{i+1}, \\  
            \rhki{r}{k}{i+1} & = l_{k}^{[i]} + \frac{1}{1 - \beta} \, \rhki{s}{k}{i+1},
        \end{aligned}
    \end{equation}
    where the log-normalizing function $\log \rhki{\psi}{k}{i+1}(x_k)$ has quadratic form~\eqref{eq:normalizer-quadratic-form}. Its parameters are
    \begin{equation}
        \label{eq:forward-gauss-markov-conditionals-normalizers}
        \begin{aligned}
            \rhkix{S}{k}{i+1} & = \rhki{G}{xx, k}{i+1} - \left[ \rhki{G}{\bar x x, k}{i+1} \right]^{\top} \left[ \rhki{G}{\bar x \bar x, k}{i+1} \right]^{-1} \rhki{G}{\bar x x, k}{i+1}, \\
            \rhki{s}{k}{i+1} & = \rhki{g}{x, k}{i+1} + \left[ \rhki{G}{\bar x x, k}{i+1} \right]^{\top} \left[ \rhki{G}{\bar x \bar x, k}{i+1} \right]^{-1} \rhki{g}{\bar x, k}{i+1},
        \end{aligned}
    \end{equation}
    with intermediate quantities
    \begin{equation}
        \label{eq:forward-gauss-markov-joint-quadratic}
        \begin{aligned}
            \rhki{G}{\bar x \bar x, k}{i+1} & \coloneqq (1 - \beta) \left[C_{\bar x \bar x, k}^{[i]} + \rhki{R}{k+1}{i+1} \right] + \beta \left[ \rhki{\Sigma}{k}{i} \right]^{-1}, \\
            \rhki{G}{xx, k}{i+1} & \coloneqq (1 - \beta) \, C_{xx, k}^{[i]} + \beta \left[ \rhkix{F}{k}{i} \right]^{\top} \left[ \rhki{\Sigma}{k}{i} \right]^{-1} \rhkix{F}{k}{i}, \\
            \rhki{G}{\bar x x, k}{i+1} & \coloneqq (1 - \beta) \, C_{\bar x x, k}^{[i]} + \beta \left[ \rhki{\Sigma}{k}{i} \right]^{-1} \rhkix{F}{k}{i}, \\
            \rhki{g}{\bar x, k}{i+1} & \coloneqq (1 - \beta) \left[ c_{\bar x, k}^{[i]} + \rhki{r}{k+1}{i+1} \right] + \beta \left[ \rhki{\Sigma}{k}{i} \right]^{-1} \rhki{d}{k}{i}, \\
            \rhki{g}{x, k}{i+1} & \coloneqq (1 - \beta) \, c_{x, k}^{[i]} - \beta \left[ \rhkix{F}{k}{i} \right]^{\top} \left[ \rhki{\Sigma}{k}{i} \right]^{-1} \rhki{d}{k}{i}.
        \end{aligned}
    \end{equation}
    Finally, given the boundary potential $\rhkixx{V}{0}{i+1}(x_{0})$ and the Gaussian density $\rhki{q}{0}{i}(x_{0}) = \mathcal{N}(x_{0} \given \rhki{m}{0}{i}, \rhkix{P}{0}{i})$, the log-normalizing constant $\log \rhki{\mathcal{Z}}{0}{i+1}$ in~\eqref{eq:forward-markov-normalizers} is computed according to
    \begin{equation}
        \label{eq:forward-gauss-markov-marginal-normalizer}
        \log \rhki{\mathcal{Z}}{0}{i+1} = - \frac{1}{2} \Big[ \rhki{m}{0}{i} \Big]^{\top} \rhix{U}{i+1} \, \rhki{m}{0}{i} + \Big[ \rhki{m}{0}{i} \Big]^{\top} \rhix{u}{i+1} + \rhix{\eta}{i+1},
    \end{equation}
    where
    \begin{equation}
        \label{eq:forward-gauss-markov-normalizer-factors}
        \begin{aligned}[t]
            \rhix{U}{i+1} & = \beta \, (1 - \beta) \left[ \rhkix{P}{0}{i} \right]^{-1} \left[ (1 - \beta) \, \rhki{R}{0}{i+1} + \beta \left[ \rhkix{P}{0}{i} \right]^{-1} \right]^{-1} \rhki{R}{0}{i+1}, \\
            \rhix{u}{i+1} & = \beta \, (1 - \beta) \left[ \rhkix{P}{0}{i} \right]^{-1} \left[ (1 - \beta) \, \rhki{R}{0}{i+1} + \beta \left[ \rhkix{P}{0}{i} \right]^{-1} \right]^{-1} \rhki{r}{0}{i+1}.
        \end{aligned}
    \end{equation}
\end{proposition}
\textit{Proof.} See Appendix~\ref{app:forward-gauss-markov-potentials-proof}.

\begin{proposition}[Recursive reverse Gauss--Markov potentials]
    \label{prop:reverse-gauss-markov-potentials}
    Let $\lhki{q}{k}{i}(x_{k-1} \given x_{k})$ be a reverse Gauss--Markov conditional as defined in Assumption~\ref{asm:reverse-gauss-markov} and let $\ell_{f}^{[i]}(x_{k}, x_{k-1})$, $\ell_{h}^{[i]}(x_{k})$, and $\ell_{p}^{[i]}(x_{0})$ be the second-order approximations of the log-densities in Definition~\ref{def:statistical-expansion-log-densities}. Then the potentials $\lhkixx{V}{k}{i+1}(x_{k})$ from~\eqref{eq:reverse-markov-potentials} are quadratic functions of the form~\eqref{eq:potentials-quadratic-form}, computed by a forward recursion starting from $k=0$ with
    \begin{equation}
        \lhki{R}{0}{i+1} = L_{0}^{[i]}, \qquad
        \lhki{r}{0}{i+1} = l_{0}^{[i]},
    \end{equation}
    For $0 < k \leq T$, the updates follow:
    \begin{equation}
        \label{eq:reverse-gauss-markov-potential-functions}
        \begin{aligned}[t]
            \lhki{R}{k}{i+1} & = L_{k}^{[i]} + \frac{1}{1 - \beta} \, \lhkix{S}{k}{i+1}, \\
            \lhki{r}{k}{i+1} & = l_{k}^{[i]} + \frac{1}{1 - \beta} \, \lhki{s}{k}{i+1},       
        \end{aligned}
    \end{equation}
    where the log-normalizing function $\log \lhki{\psi}{k}{i+1}(x_{k})$ has quadratic form~\eqref{eq:normalizer-quadratic-form}. Its parameters are
    \begin{equation}\label{eq:reverse-gauss-markov-conditionals-normalizers}
        \begin{aligned}
            \lhkix{S}{k}{i+1} & = \lhki{G}{\bar x \bar x, k}{i+1} - \left[ \lhki{G}{x \bar x, k}{i+1} \right]^{\top} \left[ \lhki{G}{xx, k}{i+1} \right]^{-1} \lhki{G}{x \bar x, k}{i+1}, \\
            \lhki{s}{k}{i+1} & = \lhki{g}{\bar x, k}{i+1} + \left[ \lhki{G}{x \bar x, k}{i+1} \right]^{\top} \left[ \lhki{G}{xx, k}{i+1} \right]^{-1} \lhki{g}{x, k}{i+1},
        \end{aligned}
    \end{equation}
    with intermediate quantities
    \begin{equation}
        \label{eq:reverse-gauss-markov-joint-quadratic}
        \begin{aligned}
            \lhki{G}{\bar x \bar x, k}{i+1} & \coloneqq (1 - \beta) \, C_{\bar x \bar x, k-1}^{[i]} + \beta \left[ \lhkix{F}{k}{i} \right]^{\top} \left[ \lhki{\Sigma}{k}{i} \right]^{-1} \lhkix{F}{k}{i}, \\
            \lhki{G}{xx, k}{i+1} & \coloneqq (1 - \beta) \, \left[ C_{xx, k-1}^{[i]} + \lhki{R}{k-1}{i+1} \right] + \beta \left[ \lhki{\Sigma}{k}{i} \right]^{-1}, \\
            \lhki{G}{x \bar x, k}{i+1} & \coloneqq (1 - \beta) \, C_{x \bar x, k-1}^{[i]} + \beta \left[ \lhki{\Sigma}{k}{i} \right]^{-1} \lhkix{F}{k}{i}, \\
            \lhki{g}{\bar x, k}{i+1} & \coloneqq (1 - \beta) \, c_{\bar x, k-1}^{[i]} - \beta \left[ \lhkix{F}{k}{i} \right]^{\top} \left[ \lhki{\Sigma}{k}{i} \right]^{-1} \lhki{d}{k}{i}, \\
            \lhki{g}{x, k}{i+1} & \coloneqq (1 - \beta) \, \left[ c_{x, k-1}^{[i]} + \lhki{r}{k-1}{i+1} \right] + \beta \left[ \lhki{\Sigma}{k}{i} \right]^{-1} \lhki{d}{k}{i}.
        \end{aligned}
    \end{equation}
    Finally, given a quadratic potential function $\lhkixx{V}{\T}{i+1}(x_{\T})$ of the form~\eqref{eq:potentials-quadratic-form} and Gaussian posterior density $\lhki{q}{\T}{i}(x_{\T}) = \mathcal{N}(x_{\T} \given \lhki{m}{\T}{i}, \lhkix{P}{\T}{i})$, the log-normalizing constant $\log \lhki{\mathcal{Z}}{\T}{i+1}$ is computed according to \eqref{eq:reverse-markov-normalizers}
    \begin{equation}
        \label{eq:reverse-gauss-markov-marginal-normalizer}
        \log \lhki{\mathcal{Z}}{\T}{i+1} = - \frac{1}{2} \left[ \lhki{m}{\T}{i} \right]^{\top} \lhix{U}{i+1} \, \lhki{m}{\T}{i} + \left[ \lhki{m}{\T}{i} \right]^{\top} \lhix{u}{i+1} + \lhix{\eta}{i+1},
    \end{equation}
    where
    \begin{equation}
        \label{eq:reverse-gauss-markov-normalizer-factors}
        \begin{aligned}[t]
            \lhix{U}{i+1} & = \beta \, (1 - \beta) \left[ \lhkix{P}{\T}{i} \right]^{-1} \left[ (1 - \beta) \, \lhki{R}{\T}{i+1} + \beta \left[ \lhkix{P}{\T}{i} \right]^{-1} \right]^{-1} \lhki{R}{\T}{i+1}, \\
            \lhix{u}{i+1} & = \beta \, (1 - \beta) \left[ \lhkix{P}{\T}{i} \right]^{-1} \left[ (1 - \beta) \, \lhki{R}{\T}{i+1} + \beta \left[ \lhkix{P}{\T}{i} \right]^{-1} \right]^{-1} \lhki{r}{\T}{i+1}.
        \end{aligned}
    \end{equation}
\end{proposition}
\textit{Proof.} See Appendix~\ref{app:reverse-gauss-markov-potentials-proof}.

The results in Proposition~\ref{prop:forward-gauss-markov-potentials} and Proposition~\ref{prop:reverse-gauss-markov-potentials} rely on general second-order statistical expansions of the log-transition and log-likelihood functions. We now turn to the practical question of how these expansions can be computed, and present two distinct strategies for obtaining the required quadratic coefficients.

\subsubsection{Generalized Statistical Linear Regression}
\label{sec:generalized-regression}

Our first approach to computing the second-order statistical expansions is based on generalized statistical linear regression \citep[GSLR,][]{garcia2015posterior, tronarp2018iterative}. These \emph{enabling approximations}~\citep{sarkka2023bayesian} construct affine-Gaussian surrogate models for the transition density $f_{k}(x_{k+1} \given x_{k})$ and the measurement likelihood $h_{k}(y_{k} \given x_{k})$ by matching the relevant conditional moments under the current Gaussian iterate. These surrogates directly provide the coefficients required by Definition~\ref{def:statistical-expansion-log-densities} for the recursive computation of the potentials and log-normalizing functions.

\begin{definition}[Generalized statistical linear regression]
    \label{def:gslr}
    Let $x$ and $y$ be two random variables, and suppose that the first and second moments $\mathbb{E}\left[ x \right]$, $\mathbb{V}\left[ x \right]$, $\mathbb{E}\left[ y \right]$, $\mathbb{V}\left[ y \right]$, and $\mathbb{C}\left[ y, x \right]$ are known. Statistical linear regression approximates the relation between $x$ and $y$ by $y \approx A \, x + b + \omega$, with $\omega \sim \mathcal{N}(0, \Omega)$, where
    \begin{equation}
        A = \mathbb{C}\left[ y, x \right] \mathbb{V}\left[ x \right]^{-1}, \quad b = \mathbb{E}\left[ y \right] - A \, \mathbb{E}\left[ x \right], \quad \Omega = \mathbb{V}\left[y \right] - A \, \mathbb{V}\left[ x \right] A^{\top}.
    \end{equation}
    The parameters $\{A, b\}$ minimize the mean squared error objective $\mathrm{MSE}(A, b) = \mathbb{E}\big[(y - A \, x - b)^{\top} (y - A \, x - b) \big]$, and the covariance matrix satisfies $\Omega = \mathbb{V}\left[ y - A \, x - b \right]$. When only the conditional moments $\mathbb{E}\left[ y \given x \right]$ and $\mathbb{V}\left[ y \given x \right]$ are available, generalized statistical linear regression estimates the required marginal moments via the law of total expectation and the law of total variance:
    \begin{equation}
        \mathbb{E} \left[ y \right] = \mathbb{E} \left[ \mathbb{E} \left[ y \given x \right] \right], \quad 
        \mathbb{V} \left[ y \right] = \mathbb{E} \left[ \mathbb{V} \left[ y \given x \right] \right] + \mathbb{V} \left[ \mathbb{E} \left[ y \given x \right] \right], \quad 
        \mathbb{C} \left[ y, x \right] = \mathbb{C} \left[ \mathbb{E} \left[ y \given x \right], x \right],
    \end{equation}
    where the outer expectations are taken with respect to a Gaussian marginal distribution $q(x)$ and are evaluated analytically when closed-form expressions are available, or can be efficiently approximated using numerical quadrature rules~\citep{sarkka2023bayesian}.
\end{definition}

Based on the recipe from Definition~\ref{def:gslr}, and given the conditional moments $\mathbb{E}\left[ x_{k+1} \given x_{k} \right]$, $\mathbb{V}\left[ x_{k+1} \given x_{k} \right]$, $\mathbb{E}\left[ y_{k} \given x_{k} \right]$, and $\mathbb{V}\left[ y_{k} \given x_{k} \right]$, we can replace the conditional densities $f_{k}(x_{k+1} \given x_{k})$ and $h_{k}(y_{k} \given x_{k})$, at each iteration $[i+1]$, by their affine-Gaussian approximations
\begin{equation}
    \label{eq:gslr-approximations}
    \begin{aligned}
        x_{k+1} & \approx A_{k}^{[i]} \, x_{k} + b_{k}^{[i]} + \omega_{k}^{[i]}, \quad & & \omega_{k}^{[i]} \sim \mathcal{N}(0, \Omega_{k}^{[i]}), & & & 0 \leq k < T, \\
        y_{k} & \approx H_{k}^{[i]} \, x_{k} + e_{k}^{[i]} + \delta_{k}^{[i]}, \quad & & \delta_{k}^{[i]} \sim \mathcal{N}(0, \Delta_{k}^{[i]}), & & & 0 < k \leq T,
    \end{aligned}
\end{equation}
which corresponds to generalized statistical linear regression with respect to the Gaussian marginal distributions $q_{k}^{[i]}(x_{k})$ from iteration $[i]$
\[
    \begin{array}{ccc}
        \begin{array}{ccl}
            A_{k}^{[i]} & = & \mathbb{C}^{[i]} \left[ x_{k+1}, x_{k} \right] \mathbb{V}^{[i]} \left[ x_{k} \right]^{-1}, \\[0.5em]
            b_{k}^{[i]} & = & \mathbb{E}^{[i]} \left[ x_{k+1} \right] - A_{k}^{[i]} \, \mathbb{E}^{[i]} \left[ x_{k} \right], \\[0.5em]
            \Omega_{k}^{[i]} & = & \mathbb{V}^{[i]} \left[ x_{k+1} \right] - A_{k}^{[i]} \, \mathbb{V}^{[i]} \left[ x_{k} \right] \left[ A_{k}^{[i]} \right]^{\top},
        \end{array}
        & \vline &
        \begin{array}{ccl}
            H_{k}^{[i]} & = & \mathbb{C}^{[i]} \left[ y_{k}, x_{k} \right] \mathbb{V}^{[i]} \left[ x_{k} \right]^{-1}, \\[0.5em]
            e_{k}^{[i]} & = & \mathbb{E}^{[i]} \left[ y_{k} \right] - H_{k}^{[i]} \, \mathbb{E}^{[i]} \left[ x_{k} \right], \\[0.5em]
            \Delta_{k}^{[i]} & = & \mathbb{V}^{[i]} \left[ y_{k} \right] - H_{k}^{[i]} \, \mathbb{V}^{[i]} \left[ x_{k} \right] \left[ H_{k}^{[i]} \right]^{\top}.
        \end{array}
    \end{array}
\]
Finally, the prior density $p_{0}(x_{0})$ is likewise approximated by a Gaussian density by matching the first and second moments
\begin{equation}
    \label{eq:gslr-prior}
    p_{0}(x_{0}) \approx \mathcal{N}(\mu_{0},  \Lambda_{0}), \quad \textrm{with} \quad \mu_{0} = \mathbb{E} \left[ x_{0} \right], \quad \,\, \Lambda_{0} = \mathbb{V} \left[ x_{0} \right].
\end{equation}

Given the GSLR approximations~\eqref{eq:gslr-approximations} and~\eqref{eq:gslr-prior}, we obtain the quadratic log-density expansions required by Definition~\ref{def:statistical-expansion-log-densities}, enabling closed-form computation in Proposition~\ref{prop:forward-gauss-markov-potentials} and Proposition~\ref{prop:reverse-gauss-markov-potentials}. Specifically, for all $0 \leq k < T$, the resulting coefficients for $\ell_f^{[i]}(x_{k+1}, x_k)$ are
\begin{equation}
    \begin{aligned}
        C_{\bar x \bar x, k}^{[i]} & = \left[ \Omega_{k}^{[i]} \right]^{-1}, &
        C_{\bar x x, k}^{[i]} & = \left[ \Omega_{k}^{[i]} \right]^{-1} A_{k}^{[i]}, &
        c_{\bar x, k}^{[i]} & = \left[ \Omega_{k}^{[i]} \right]^{-1} b_{k}^{[i]}, \\
        C_{xx, k}^{[i]} & = \left[ A_{k}^{[i]} \right]^{\top} \left[ \Omega_{k}^{[i]} \right]^{-1} A_{k}^{[i]}, &
        C_{x \bar x, k}^{[i]} & = \left[ A_{k}^{[i]} \right]^{\top} \left[ \Omega_{k}^{[i]} \right]^{-1}, &
        c_{x, k}^{[i]} & = - \left[ A_{k}^{[i]} \right]^{\top} \left[ \Omega_{k}^{[i]} \right]^{-1} b_{k}^{[i]}.
    \end{aligned}
\end{equation}
For all $0 < k \leq T$, the coefficients of $\ell_{h}^{[i]}(x_{k})$ are
\begin{equation}
    L_{k}^{[i]} = \left[ H_{k}^{[i]} \right]^{\top} \left[ \Delta_{k}^{[i]} \right]^{-1} H_{k}^{[i]}, \qquad l_{k}^{[i]} = \left[ H_{k}^{[i]} \right]^{\top} \left[ \Delta_{k}^{[i]} \right]^{-1} \left[ y_{k} - e_{k}^{[i]} \right],
\end{equation}
and, finally, the coefficients associated with $\ell_{p}^{[i]}(x_{0})$ are
\begin{equation}
    L_{0}^{[i]} = \Lambda_{0}^{-1}, \qquad l_{0}^{[i]} = \Lambda_{0}^{-1} \mu_{0}.
\end{equation}

\subsubsection{Fourier--Hermite Series Expansion}
\label{sec:fourier-hermite}

The statistical approximation provided by statistical linear regression comes with two main limitations: it imposes an explicit additive Gaussian noise model on the approximated dynamics and measurements, and neglects second-order curvature information. Alternatively, we can obtain second-order statistical approximations using Fourier--Hermite series~\citep{sarmavuori2011fourier, hassan2023fourier}, which leverage Hermite polynomial bases in a Hilbert space $\mathcal{H}$ to capture higher-order effects~\citep{janson1997gaussian}.

\begin{definition}[Fourier--Hermite expansion]
    \label{def:fourier-hermite}
    Let $s\sim\mathcal{N}(0,I_d)$ and let $g$ be a scalar-valued function in the Hilbert space $\mathcal{H}=L^2(\mathbb{R}^d,\mathcal{N}(0,I_d))$ equipped with the inner product $\langle f,g\rangle_{\mathcal{H}}=\mathbb{E}[f(s)g(s)]$. A second-order Fourier--Hermite expansion of $g(s)$ is given by
    \begin{equation}
        g(s) \approx \mathbb{E} \left[ g(s) \right] + \mathbb{E} \left[ g(s) \, H_{1}(s) \right]^{\top} H_{1}(s) + \frac{1}{2} \tr \left\{ \mathbb{E} \left[ g(s) \, H_{2}(s) \right] H_{2}(s) \right\},
    \end{equation}
    where $H_{1}(s)$ and $H_{2}(s)$ are the first- and second-order Hermite polynomials defined as 
    \begin{equation}
        H_{1}(s) = s, \qquad H_{2}(s) = s s^{\top} - I_d.
    \end{equation}
    This expansion generalizes to any Gaussian $\mathcal{N}(m, P)$ by letting $z = R s + m$ and $P = R R^\top$, so that
    \begin{align}
        g(z) & \approx
        \begin{aligned}[t]
            & \mathbb{E} \left[ g(z) \right] + \mathbb{E} \left[ g(z) \, H_{1}(R^{-1} (z - m)) \right]^{\top} H_{1}(R^{-1} (z - m)) \\
            & + \frac{1}{2} \tr \left\{ \mathbb{E} \left[ g(z) \, H_{2}(R^{-1} (z - m)) \right] \left[ R^{-1} (z - m) (z - m)^{\top} R^{-\top} - I \right] \right\}
        \end{aligned} \\
        & = -\frac{1}{2} z^{\top} U z + z^{\top} u + \eta.
    \end{align}
    The coefficients of this quadratic form are
    \begin{align}
        U & \coloneqq - \mathbb{E} \left[ \nabla^2_{\!z} \, g(z) \right], \quad u \coloneqq \mathbb{E} \left[ \nabla_{\!z} \, g(z) \right] - \mathbb{E} \left[ \nabla^2_{\!z} \, g(z) \right] m, \\
        \eta & \coloneqq 
        \begin{aligned}[t]
            & \mathbb{E} \left[ g(z) \right] - \mathbb{E} \left[ \nabla_{\!z} \, g(z) \right]^{\top} m + \frac{1}{2} m^{\top} \mathbb{E} \left[ \nabla^2_{\!z} \, g(z) \right] m - \frac{1}{2} \tr \left\{ R^{\top} \, \mathbb{E} \left[ \nabla^2_{\!z} \, g(z) \right] R \right\}.
        \end{aligned}
    \end{align}
    Assuming $g$ is twice continuously differentiable, and that $g$, the gradient $\nabla_{\!z}\,g(z)$, and Hessian $\nabla^2_{\!z}\,g(z)$ all have finite second moments under $\mathcal{N}(m,P)$, the following identities hold~\citep{hassan2023fourier}:
    \begin{equation}
        \mathbb{E} \left[ g(z) \, H_{1}(R^{-1} (z - m)) \right] = R^{\top} \mathbb{E} \left[ \nabla_{\!z} \, g(z) \right], \qquad \mathbb{E} \left[ g(z) \, H_{2}(R^{-1} (z - m)) \right] = R^{\top} \mathbb{E} \left[ \nabla^2_{\!z} \, g(z) \right] R.
    \end{equation}
    The expectations defining these coefficients can be evaluated analytically when closed-form expressions are available, or approximated using numerical quadrature rules~\citep{sarkka2023bayesian}.
\end{definition}

Given a set of marginal distributions $q_{k}^{[i]}(x_{k})$, we can apply the expansion in Definition~\ref{def:fourier-hermite} to the log-transition and log-measurement functions to obtain second-order approximations consistent with Definition~\ref{def:statistical-expansion-log-densities}. This approach provides an alternative to GSLR that does not impose an explicit additive Gaussian noise assumption and incorporates second-order information, offering potentially higher-fidelity approximations.

\subsection{Gauss--Markov Posterior Updates}

\label{sec:gauss-markov-updates}
Having established the quadratic approximations that enable recursive computation of the potential functions, we now turn our attention to the computation of the tilted distributions introduced in Proposition~\ref{prop:forward-markov} and Proposition~\ref{prop:reverse-markov}. As specified in Assumption~\ref{asm:forward-gauss-markov} and Assumption~\ref{asm:reverse-gauss-markov}, our goal is to maintain the Gauss--Markov structure of the forward and reverse variational posteriors throughout the iterative optimization process. To ensure that the updated variational distributions remain within the Gauss--Markov family, we project the corresponding tilted distributions onto a marginal or conditional Gaussian form. 

We begin by deriving a general moment-matching rule for tilted marginal distributions of the form in~\eqref{eq:forward-markov-optimal-marginal} and~\eqref{eq:reverse-markov-optimal-marginal}. These results then directly yield closed-form updates for the parameters $\rhki{m}{0}{i+1}$, $\rhkix{P}{0}{i+1}$, $\lhki{m}{\T}{i+1}$, and $\lhkix{P}{\T}{i+1}$, preserving the tractability and structure of the overall smoothing algorithms.

\begin{lemma}[Tilted Gaussian moment matching]
    \label{lem:gauss-markov-marginal-update}
    Let $q^{[i+1]}(x)$ be a tilted distribution of the form
    \begin{equation}
        q^{[i+1]}(x) = \left[ \mathcal{Z}^{[i+1]} \right]^{-1} \Big[ q^{[i]}(x) \Big]^{\beta} \Big[ \exp \left\{ V^{[i+1]}(x) \right\} \Big]^{1 - \beta},
    \end{equation}
    with $q^{[i]}(x) = \mathcal{N}(x \given m^{[i]}, P^{[i]})$, and a normalizing constant
    \begin{equation}
        \mathcal{Z}^{[i+1]} = \int \Big[ q^{[i]}(x) \Big]^{\beta} \Big[ \exp \left\{ V^{[i+1]}(x) \right\} \Big]^{1 - \beta} \dif x.
    \end{equation}
    Then, for $\beta\in(0,1)$, the first and second moments of $q^{[i+1]}(x)$ are given by
    \begin{equation}
        \begin{aligned}
            \mathbb{E}^{[i+1]} \left[ x \right]  = m^{[i]} + \frac{1}{\beta} \, P^{[i]} \, \frac{\partial \log \mathcal{Z}^{[i+1]}}{\partial \, m^{[i]}}, \qquad 
            \mathbb{V}^{[i+1]} \left[ x \right]  = \frac{1}{\beta} \, P^{[i]} + \frac{1}{\beta^{2}} \, P^{[i]} \frac{\partial^{2} \log \mathcal{Z}^{[i+1]}}{\partial \, m^{[i]} \, \partial \left[ m^{[i]} \right]^{\top}} \, P^{[i]}.
        \end{aligned}
    \end{equation}
\end{lemma}
\textit{Proof.} See Appendix~\ref{app:gauss-markov-marginal-update-proof}.

\begin{corollary}[Forward Gauss--Markov boundary]
    \label{cor:forward-gauss-markov-marginal-update}
    Let $\rhki{q}{0}{i}(x_0)$ be a Gaussian marginal distribution and let $\rhkixx{V}{0}{i+1}(x_{0})$ and $\log \rhki{\mathcal{Z}}{0}{i+1}$ be the quadratic potential function and log-normalizing constant in~\eqref{eq:forward-gauss-markov-potential-functions} and~\eqref{eq:forward-gauss-markov-marginal-normalizer}, respectively. Then, the tilted marginal $\rhki{q}{0}{i+1}(x_0)$ from~\eqref{eq:forward-markov-optimal-marginal} admits a Gaussian form:
    \begin{equation}
        \begin{aligned}
            \rhkix{P}{0}{i+1} &= \left[ (1 - \beta) \, \rhki{R}{0}{i+1} + \beta \left[ \rhkix{P}{0}{i} \right]^{-1} \right]^{-1}, \\
            \rhki{m}{0}{i+1} &= \rhkix{P}{0}{i+1} \, \left[ (1 - \beta) \, \rhki{r}{0}{i+1} + \beta \left[ \rhkix{P}{0}{i} \right]^{-1} \rhki{m}{0}{i} \right].
        \end{aligned}
    \end{equation}
    These expressions follow from Lemma~\ref{lem:gauss-markov-marginal-update} for $\beta\in(0,1)$ and from directly completing the square for $\beta=0$.
\end{corollary}

\begin{corollary}[Reverse Gauss--Markov boundary]
    \label{cor:reverse-gauss-markov-marginal-update}
    Let $\lhki{q}{\T}{i}(x_{\T})$ be a Gaussian marginal distribution and let $\lhkixx{V}{\T}{i+1}(x_{\T})$ and $\log \lhki{\mathcal{Z}}{\T}{i+1}$ be the quadratic potential function and log-normalizing constant in~\eqref{eq:reverse-gauss-markov-potential-functions} and~\eqref{eq:reverse-gauss-markov-marginal-normalizer}, respectively. Then, the tilted marginal $\lhki{q}{\T}{i+1}(x_{\T})$ from~\eqref{eq:reverse-markov-optimal-marginal} admits a Gaussian form:
    \begin{equation}
        \begin{aligned}
            \lhkix{P}{\T}{i+1} &= \left[ (1 - \beta) \, \lhki{R}{\T}{i+1} + \beta \left[ \lhkix{P}{\T}{i} \right]^{-1} \right]^{-1}, \\
            \lhki{m}{\T}{i+1} &= \lhkix{P}{\T}{i+1} \left[ (1 - \beta) \, \lhki{r}{\T}{i+1} + \beta \left[ \lhkix{P}{\T}{i} \right]^{-1} \lhki{m}{\T}{i} \right].
        \end{aligned}
    \end{equation}
    These expressions follow from Lemma~\ref{lem:gauss-markov-marginal-update} for $\beta\in(0,1)$ and from directly completing the square for $\beta=0$.
\end{corollary}
Projecting the tilted conditionals in~\eqref{eq:forward-markov-optimal-conditionals} and~\eqref{eq:reverse-markov-optimal-conditionals} onto the affine-Gaussian parametric family defined in Assumption~\ref{asm:forward-gauss-markov} and Assumption~\ref{asm:reverse-gauss-markov} is more involved, as it requires deriving the conditional moments of the forward and reverse distributions $\mathbb{E} \left[ x_{k+1} \given x_{k} \right]$, $\mathbb{V} \left[ x_{k+1} \given x_{k} \right]$, $\mathbb{E} \left[ x_{k-1} \given x_{k} \right]$, and $\mathbb{V} \left[ x_{k-1} \given x_{k} \right]$. In what follows, we derive closed-form expressions for these moments and use them to update the parameters of the forward and reverse Gauss--Markov posteriors.

\begin{proposition}[Forward Gauss--Markov conditionals]
    \label{prop:forward-gauss-markov-conditionals-update}
    Let $\rhki{q}{k}{i}(x_{k+1} \given x_{k})$ be the forward Gauss--Markov conditional in Assumption~\ref{asm:forward-gauss-markov}. Additionally, suppose that $\ell_{f}^{[i]}(x_{k+1}, x_{k})$ admits a second-order expansion as in~\eqref{eq:transition-quadratic-form}, and that the log-message $\log \rhki{\psi}{k}{i+1}(x_k)$ is quadratic as in~\eqref{eq:forward-gauss-markov-conditionals-normalizers}. Then the tilted conditional $\rhki{q}{k}{i+1}(x_{k+1} \given x_k)$ is linear-Gaussian with conditional moments
    \begin{equation}
        \mathbb{E}^{[i+1]} \left[ x_{k+1} \given x_{k} \right] = \Big[ \rhki{G}{\bar x \bar x, k}{i+1} \Big]^{-1} \left[ \rhki{G}{\bar x x, k}{i+1} \, x_{k} + \rhki{g}{\bar x, k}{i+1} \right], \qquad \mathbb{V}^{[i+1]} \left[ x_{k+1} \given x_{k} \right] = \Big[ \rhki{G}{\bar x \bar x, k}{i+1} \Big]^{-1},
    \end{equation}
    and Gauss--Markov parameters
    \begin{equation}
        \rhki{\Sigma}{k}{i+1} = \Big[ \rhki{G}{\bar x \bar x, k}{i+1} \Big]^{-1}, \qquad
        \rhkix{F}{k}{i+1} = \Big[ \rhki{G}{\bar x \bar x, k}{i+1} \Big]^{-1} \, \rhki{G}{\bar x x, k}{i+1}, \qquad
        \rhki{d}{k}{i+1} = \Big[ \rhki{G}{\bar x \bar x, k}{i+1} \Big]^{-1} \, \rhki{g}{\bar x, k}{i+1}.
    \end{equation}
    Substituting the parameters from~\eqref{eq:forward-gauss-markov-joint-quadratic} gives the explicit update
    \begin{equation}
        \begin{aligned}
            \rhki{\Sigma}{k}{i+1} & = \left[ (1 - \beta) \left[ C_{\bar x \bar x, k}^{[i]} + \rhki{R}{k+1}{i+1} \right] + \beta \left[ \rhki{\Sigma}{k}{i} \right]^{-1} \right]^{-1}, \\
            \rhkix{F}{k}{i+1} & = \rhki{\Sigma}{k}{i+1} \left[ (1 - \beta) \, C_{\bar x x, k}^{[i]} + \beta \left[ \rhki{\Sigma}{k}{i} \right]^{-1} \rhkix{F}{k}{i} \right], \\
            \rhki{d}{k}{i+1} & = \rhki{\Sigma}{k}{i+1} \left[ (1 - \beta) \left[ c_{\bar x, k}^{[i]} + \rhki{r}{k+1}{i+1} \right] + \beta \left[ \rhki{\Sigma}{k}{i} \right]^{-1} \rhki{d}{k}{i} \right].
        \end{aligned}
    \end{equation}
\end{proposition}
\textit{Proof.} See Appendix~\ref{app:forward-gauss-markov-conditionals-update-proof}.

\begin{proposition}[Reverse Gauss--Markov conditionals]
    \label{prop:reverse-gauss-markov-conditionals-update}
    Let $\lhki{q}{k}{i}(x_{k-1} \given x_{k})$ be the reverse Gauss--Markov conditional in Assumption~\ref{asm:reverse-gauss-markov}. Additionally, suppose that $\ell_{f}^{[i]}(x_{k}, x_{k-1})$ admits a second-order expansion as in~\eqref{eq:transition-quadratic-form}, and that the log-message $\log \lhki{\psi}{k}{i+1}(x_k)$ is quadratic as in~\eqref{eq:reverse-gauss-markov-conditionals-normalizers}. Then the tilted conditional $\lhki{q}{k}{i+1}(x_{k-1} \given x_{k})$ is linear-Gaussian with conditional moments
    \begin{equation}
        \mathbb{E}^{[i+1]} \left[ x_{k-1} \given x_{k} \right] = \Big[ \lhki{G}{xx, k}{i+1} \Big]^{-1} \left[ \lhki{G}{x \bar x, k}{i+1} \, x_{k} + \lhki{g}{x, k}{i+1} \right], \qquad \mathbb{V}^{[i+1]} \left[ x_{k-1} \given x_{k} \right] = \Big[ \lhki{G}{xx, k}{i+1} \Big]^{-1},
    \end{equation}
    and Gauss--Markov parameters
    \begin{equation}
        \lhki{\Sigma}{k}{i+1} = \Big[ \lhki{G}{xx, k}{i+1} \Big]^{-1}, \qquad
        \lhkix{F}{k}{i+1} = \Big[ \lhki{G}{xx, k}{i+1} \Big]^{-1} \, \lhki{G}{x \bar x, k}{i+1}, \qquad
        \lhki{d}{k}{i+1} = \Big[ \lhki{G}{xx, k}{i+1} \Big]^{-1} \, \lhki{g}{x, k}{i+1}.
    \end{equation}
    Substituting the parameters from~\eqref{eq:reverse-gauss-markov-joint-quadratic} gives the explicit update
    \begin{equation}
        \begin{aligned}
            \lhki{\Sigma}{k}{i+1} & = \left[ (1 - \beta) \, \left[ C_{xx, k-1}^{[i]} + \lhki{R}{k-1}{i+1} \right] + \beta \left[ \lhki{\Sigma}{k}{i} \right]^{-1} \right]^{-1}, \\
            \lhkix{F}{k}{i+1} & = \lhki{\Sigma}{k}{i+1} \left[ (1 - \beta) \, C_{x \bar x, k-1}^{[i]} + \beta \left[ \lhki{\Sigma}{k}{i} \right]^{-1} \lhkix{F}{k}{i} \right], \\
            \lhki{d}{k}{i+1} & = \lhki{\Sigma}{k}{i+1} \left[ (1 - \beta) \, \left[ c_{x, k-1}^{[i]} + \lhki{r}{k-1}{i+1} \right] + \beta \left[ \lhki{\Sigma}{k}{i} \right]^{-1} \lhki{d}{k}{i} \right].
        \end{aligned}
    \end{equation}
\end{proposition}
\textit{Proof.} See Appendix~\ref{app:reverse-gauss-markov-conditionals-update-proof}.

The boundary and conditional updates above admit a (local) natural gradient interpretation with step size $1 - \beta$. Appendix~\ref{app:natural-gradient} establishes this correspondence via the Fourier--Hermite coefficients.

\subsection{Recursive Gaussian Marginals}

Given the Gauss--Markov parameterization of the posterior in Assumption~\ref{asm:forward-gauss-markov} and Assumption~\ref{asm:reverse-gauss-markov}, together with the update rules introduced in Section~\ref{sec:gauss-markov-updates}, the smoothing posterior marginals at each iteration $[i+1]$ can be computed efficiently. Specifically, we use a forward recursion to evaluate the marginals of the forward Gauss--Markov posterior as described in Remark~\ref{rem:forward-markov-marginals}, and a backward recursion to evaluate the marginals of the reverse Gauss--Markov posterior as described in Remark~\ref{rem:reverse-markov-marginals}. These marginals play a central role in the proposed iterative smoothing framework, since they are used to construct the second-order statistical expansions of the log-densities associated with the dynamics and measurement model, as discussed in Section~\ref{sec:quadratic-potentials}.

\begin{corollary}[Forward Gauss--Markov marginals]
    \label{cor:forward-gauss-markov-marginals}
    Given the forward Gauss--Markov posterior from Corollary~\ref{cor:forward-gauss-markov-marginal-update} and Proposition~\ref{prop:forward-gauss-markov-conditionals-update}, the forward marginal smoothing distributions $\rhki{q}{k+1}{i+1}(x_{k+1})$, for all $0 \leq k < T$, are computed in closed form via the forward recursion
    \begin{align}
        \rhki{m}{k+1}{i+1} &= \rhkix{F}{k}{i+1} \, \rhki{m}{k}{i+1} + \rhki{d}{k}{i+1}, \\
        \rhkix{P}{k+1}{i+1} &= \rhkix{F}{k}{i+1} \, \rhkix{P}{k}{i+1} \, \left[ \rhkix{F}{k}{i+1} \right]^{\top} + \rhki{\Sigma}{k}{i+1},
    \end{align}
    with the initial condition $\rhki{q}{0}{i+1}(x_{0}) = \mathcal{N}(x_{0} \given \rhki{m}{0}{i+1}, \rhkix{P}{0}{i+1})$.
\end{corollary}

\begin{corollary}[Reverse Gauss--Markov marginals]
    \label{cor:reverse-gauss-markov-marginals}
    Given the reverse Gauss--Markov posterior from Corollary~\ref{cor:reverse-gauss-markov-marginal-update} and Proposition~\ref{prop:reverse-gauss-markov-conditionals-update}, the reverse marginal smoothing distributions $\lhki{q}{k-1}{i+1}(x_{k-1})$, for all $1 \leq k \leq T$, are computed in closed form via the backward recursion
    \begin{align}
        \lhki{m}{k-1}{i+1} &= \lhkix{F}{k}{i+1} \, \lhki{m}{k}{i+1} + \lhki{d}{k}{i+1}, \\
        \lhkix{P}{k-1}{i+1} &= \lhkix{F}{k}{i+1} \, \lhkix{P}{k}{i+1} \, \left[ \lhkix{F}{k}{i+1} \right]^{\top} + \lhki{\Sigma}{k}{i+1},
    \end{align}
    with the terminal condition $\lhki{q}{\T}{i+1}(x_{\T}) = \mathcal{N}(x_{\T} \given \lhki{m}{\T}{i+1}, \lhkix{P}{\T}{i+1})$.
\end{corollary}

\begin{corollary}[Hybrid Gauss--Markov marginals]
    \label{cor:hybrid-gauss-markov-marginals}
    Given the recursions from Proposition~\ref{prop:forward-gauss-markov-potentials} and Proposition~\ref{prop:reverse-gauss-markov-potentials}, suppose that the current marginal is Gaussian, $q_{k}^{[i]}(x_{k}) = \mathcal{N}(x_{k} \given m_{k}^{[i]}, P_{k}^{[i]})$. Then, the interior hybrid marginal smoothing distributions $q_{k}^{[i+1]}(x_{k})$, for all $0 < k < T$, are computed according to Corollary~\ref{cor:hybrid-markov} in closed form by completing the square:
    \begin{align}
        P_{k}^{[i+1]} & = \left[ (1 - \beta) \left[ \lhki{R}{k}{i+1} + \rhki{S}{k}{i+1} \right] + \beta \left[ P_{k}^{[i]} \right]^{-1} \right]^{-1}, \\
        m_{k}^{[i+1]} & = P_{k}^{[i+1]} \left[ (1 - \beta) \left[ \lhki{r}{k}{i+1} + \rhki{s}{k}{i+1} \right] + \beta \left[ P_{k}^{[i]} \right]^{-1} m_{k}^{[i]} \right].
    \end{align}
    The boundary marginals are given by the corresponding forward and reverse boundary updates, namely $q_{0}^{[i+1]}(x_{0}) = \mathcal{N}(x_{0} \given \rhki{m}{0}{i+1}, \rhkix{P}{0}{i+1})$ and $q_{\T}^{[i+1]}(x_{\T}) = \mathcal{N}(x_{\T} \given \lhki{m}{\T}{i+1}, \lhkix{P}{\T}{i+1})$.
\end{corollary}

\subsection{Positive Definiteness and Bisection}
\label{sec:bisection}

We now address the practical search procedure for the damping parameter $\beta$. At each outer iteration, the local quadratic approximations are constructed under the current iterate $q^{[i]}$ and held fixed while optimizing over $\alpha$, where $\beta=\alpha/(1+\alpha)$. The Gauss--Markov update is recomputed for each candidate value of $\beta$.

According to the general optimization problem described in Section~\ref{sec:optimal-damping}, the optimal parameter $\alpha^\star$ can be found by searching for a root of the trust-region slack $\Delta_{\mathrm{KL}}(\alpha)$ in~\eqref{eq:temperature-grad}. However, this search must account for the fact that the quadratic approximations used in Proposition~\ref{prop:forward-gauss-markov-potentials} and Proposition~\ref{prop:reverse-gauss-markov-potentials} may fail to preserve the positive definiteness of the precision matrices required by the Gaussian updates. In particular, for the forward Gauss--Markov recursion, this requires
\begin{equation}
    \rhki{G}{\bar{x}\bar{x}, k}{i+1} \succ 0
    \quad \text{for all} \quad 0 \leq k < T,
    \qquad \text{and} \qquad
    (1-\beta)\,\rhki{R}{0}{i+1}
    + \beta \left[\rhkix{P}{0}{i}\right]^{-1} \succ 0,
\end{equation}
while for the reverse Gauss--Markov recursion, we require
\begin{equation}
    \lhki{G}{xx, k}{i+1} \succ 0
    \quad \text{for all} \quad 0 < k \leq T,
    \qquad \text{and} \qquad
    (1 - \beta) \, \lhki{R}{\T}{i+1}
    + \beta \left[ \lhkix{P}{\T}{i} \right]^{-1} \succ 0.
\end{equation}
The damping parameter plays a central role in satisfying these conditions. Since the current iterate $q^{[i]}$ is a well-defined Gaussian, the Gauss--Markov updates approach the current iterate as $\beta\to1$. Consequently, for any $\varepsilon>0$, sufficiently strong damping yields a positive-definite update within the KL radius. 

To realize the selection procedure, we adopt a simple bisection method to search for a root of the trust-region slack and reject candidates that violate the positive-definiteness conditions. An outline of this approach is presented in Algorithm~\ref{alg:forward-gauss-markov-optimal-damping} and Algorithm~\ref{alg:reverse-gauss-markov-optimal-damping}. 

\subsection{Proximal Variational Bayesian Smoothing Algorithms}

The preceding sections provide the components needed to instantiate the general results of Proposition~\ref{prop:forward-markov}, Proposition~\ref{prop:reverse-markov}, and Corollary~\ref{cor:hybrid-markov} for Gaussian variational posterior approximations. In this section, we present an overview of the resulting three iterative Bayesian smoothing algorithms.

The \textit{Forward Entropic Variational Smoother (FEVS)}, described in Algorithm~\ref{alg:forward-gauss-markov-iterated-smoother}, iteratively refines the posterior estimate through a backward-forward recursion. First, a backward pass computes the potential functions, log-normalizing terms, and updated forward conditional smoothing distributions, as detailed in Algorithm~\ref{alg:forward-gauss-markov-backward-pass}. This is followed by a forward pass that propagates the smoothed marginal distributions, as outlined in Algorithm~\ref{alg:forward-gauss-markov-forward-pass}. At each iteration, the damping parameter is adapted using Algorithm~\ref{alg:forward-gauss-markov-optimal-damping}, which enforces the proximal trust-region constraint induced by the entropic regularization.

The \textit{Reverse Entropic Variational Smoother (REVS)}, presented in Algorithm~\ref{alg:reverse-gauss-markov-iterated-smoother}, follows the inverse recursion order. A forward pass computes the potential functions, log-normalizing terms, and updated reverse conditional smoothing distributions, as given in Algorithm~\ref{alg:reverse-gauss-markov-forward-pass}. This is followed by a backward pass that propagates the smoothed marginal distributions, according to Algorithm~\ref{alg:reverse-gauss-markov-backward-pass}. As in the forward smoother, the damping parameter is adaptively selected at each iteration using Algorithm~\ref{alg:reverse-gauss-markov-optimal-damping}.

The \textit{Hybrid Entropic Variational Smoother (HEVS)}, shown in Algorithm~\ref{alg:hybrid-gauss-markov-iterated-smoother}, combines the two directional message recursions. In this scheme, the potential functions and log-normalizing terms are computed using both the backward pass from the forward smoother in Algorithm~\ref{alg:forward-gauss-markov-backward-pass} and the forward pass from the reverse smoother in Algorithm~\ref{alg:reverse-gauss-markov-forward-pass}. The resulting representations are then combined at the marginal level to compute the smoothed marginal distributions, as described in Corollary~\ref{cor:hybrid-markov}. The damping parameter can be adapted using either the forward or reverse dual objective to enforce the proximal constraint.

All proposed methods share several favorable properties. First, the recursions are formulated through log-potentials and log-normalizing functions, which improves numerical stability and mitigates underflow over long time horizons. Second, each recursion is damped through a KL trust-region constraint, which controls the deviation between successive posterior iterates and reduces sensitivity to local approximation error. Finally, for fixed state dimension and fixed cost per local approximation step, all algorithms have linear time complexity with respect to the horizon $T$, making them scalable for long sequences.

\subsection{Properties and Limitations}
\label{sec:limitations}

Trust-region methods typically perform a step using a local model of the objective and restrict that step to a region where the approximation remains valid~\citep{nocedal2006numerical}. Our practical Gauss--Markov algorithms follow a similar principle. At each iteration, we use the statistical expansions in Section~\ref{sec:statistical-approximations} to approximate the objective around $q^{[i]}$, then optimize the parameter $\beta$ to ensure that the new iterate $q^{[i+1]}$ is a valid Gaussian posterior within the KL radius $\varepsilon$.

By construction, optimizing the exact constrained problem~\eqref{eq:smoothing-proximal} cannot decrease the ELBO. However, under approximation, a step that improves the local model may fail to improve the true ELBO. Classical trust-region methods handle this by comparing the actual improvement of the true objective with the improvement predicted by the local model before accepting or rejecting the step~\citep{nocedal2006numerical}, somewhat analogous to line-search methods.

A similar acceptance test could be incorporated into our framework to adapt the radius $\varepsilon$, but it requires an accurate evaluation of the true ELBO, for example via numerical integration. Our implementation does not perform such a test and cannot guarantee monotonic improvement of the  ELBO. Nevertheless, our experiments show stable improvement of the ELBO when sufficiently small trust-region radii are selected to produce cautious updates of the variational posterior.

\section{Connection to Existing Bayesian Inference Algorithms}
\label{sec:related-work}

This section provides an overview of recent research that helps situate our contribution. We highlight related research in signal processing, Markovian Gaussian processes, and approximate Bayesian inference.

\subsection{Forward-Backward Smoothing Algorithms}
For the forward Gauss--Markov decomposition introduced in Proposition~\ref{prop:forward-markov} and Proposition~\ref{prop:forward-gauss-markov-potentials}, a direct conceptual connection can be made to the classical smoother proposed by \citet{cox1964estimation}. In his work, \citeauthor{cox1964estimation} formulates Bayesian smoothing in state-space models with additive Gaussian noise as a maximum a posteriori (MAP) optimization problem. Subject to certain non-singularity conditions, he derives a dynamic programming solution that propagates adjoint state functions backward in time, which aligns closely with the backward recursion over potential functions used in our framework. This method was later extended to nonlinear systems via iterative linearization techniques~\citep{mortensen1968maximum}. Our approach generalizes \citeauthor{cox1964estimation}’s formulation by moving beyond MAP estimation to target a full Gaussian approximation of the smoothing posterior. In doing so, we accommodate a broader class of nonlinear, non-Gaussian state-space models. Moreover, by incorporating entropic regularization, our method introduces a principled mechanism for trust-region control, ensuring stable and well-posed updates over the space of variational densities.

In contrast, the reverse Gauss--Markov decomposition described in Proposition~\ref{prop:reverse-markov} and Proposition~\ref{prop:reverse-gauss-markov-potentials} yields a recursion that closely parallels the structure of the Rauch--Tung--Striebel (RTS) smoother~\citep{rauch1965maximum}. The RTS smoother operates by first performing a forward filtering pass to accumulate measurement information, followed by a backward recursion that propagates smoothed marginals through the implicit computation of reverse posterior conditionals. Our iterated reverse Gauss--Markov smoother generalizes this two-pass structure to a significantly broader class of models. It lifts the standard assumptions of linear dynamics and Gaussian noise, enabling application to nonlinear, non-Gaussian systems. Additionally, by operating directly in the log-domain of the filtered densities and incorporating entropic proximal regularization, our method enhances numerical stability and provides a principled means of controlling the update step size throughout the recursion.

Finally, the hybrid posterior decomposition introduced in Corollary~\ref{cor:hybrid-markov} and Corollary~\ref{cor:hybrid-gauss-markov-marginals} bears a strong resemblance to log-space two-filter smoothers~\citep{mayne1966solution, fraser1969optimum}. Originally developed for MAP-based smoothing in linear-Gaussian state-space models, these methods were later extended to nonlinear settings through Gaussian-sum filters~\citep{kitagawa1987non}. The canonical structure involves independent forward and backward filtering recursions that are subsequently combined to form the smoothing solution. Our hybrid smoother adopts this two-filter architecture by jointly leveraging both the forward and reverse Gauss--Markov decompositions of the variational posterior. In doing so, it generalizes the classical two-filter approach to accommodate nonlinear and non-Gaussian models within a variational inference framework. To the best of our knowledge, this is the first attempt to formulate an iterated two-filter smoother that integrates proximal updates and operates in the log-domain of the densities, akin to the forward and reverse variants. A notable advantage of this hybrid scheme is the structural independence of the forward and backward recursions, which allows for parallel execution. Although this design increases the overall computational cost relative to the single-pass algorithms, it offers practical benefits in terms of speed and modularity when implemented on modern parallel hardware.

\subsection{Posterior-Linearization Bayesian Smoothing}
Our work is primarily inspired by posterior-linearization algorithms \citep{garcia2015posterior, garcia2016iterated, tronarp2018iterative}, which perform approximate Gaussian Bayesian smoothing in nonlinear state-space models by alternating between posterior linearization and a Rauch--Tung--Striebel smoothing pass. These algorithms can be viewed as generalizations of the maximum a posteriori iterated smoother proposed by \citet{bell1994iterated}. While these methods offer advantages over traditional extended and unscented smoothers~\citep{sarkka2023bayesian}, they have two significant limitations.

First, they impose linearity and additive noise assumptions on the dynamics and measurement likelihoods, which can be problematic in state-space models with multiplicative noise, as highlighted by \citet{corenflos2023variational}. Second, these algorithms, in their original form, fall within the class of undamped Gauss--Newton optimization methods that rely on linear approximations of the dynamics and measurement models. Crucially, Gauss--Newton techniques require a full-rank Jacobian to ensure convergence~\citep{nocedal2006numerical}, a condition that is not generally met in nonlinear settings.

To address the latter issue, a partial remedy was proposed by \citet{raitoharju2018damped}, who introduced an \emph{ad hoc} damping mechanism for the mean updates in the iterated posterior-linearization filter, albeit at the cost of a computationally expensive nested optimization loop. In contrast, \citet{lindqvist2021posterior} proposed a posterior-linearization smoother that implements damping through Levenberg--Marquardt regularization and line-search procedures with convergence guarantees. Both \citet{raitoharju2018damped} and \citet{lindqvist2021posterior}, however, fail to exploit the information-geometric structure of the statistical manifold on which the approximate smoothing distribution is assumed to reside.

Our approach generalizes these algorithms and resolves both weaknesses. By using Fourier--Hermite expansions, we overcome the limitations of linear-Gaussian approximations, while the introduction of entropic proximal constraints provides a principled way to dampen optimization over the space of densities, leading to techniques akin to trust-region approaches~\citep{nocedal2006numerical, teboulle1992entropic}. This improvement, however, comes at the cost of increased algorithmic complexity. 

\subsection{Approximate Bayesian Inference in State-Space Models}

The field of approximate Bayesian inference has seen significant progress in adapting standard techniques to exploit the temporal structure of state-space models, resulting in specialized algorithms for Bayesian smoothing. One influential approach builds on expectation propagation (EP), introduced by \citet{minka2001expectation}. In particular, \citet{deisenroth2012expectation} applied EP to nonlinear state-space models with additive Gaussian noise. Their method iteratively linearizes the dynamics and measurement models around the current posterior estimate, yielding tractable forward–backward message-passing recursions. However, the construction is largely heuristic and lacks a clearly defined global objective, which complicates convergence analysis, especially in nonlinear or non-Gaussian regimes. In contrast, our method is grounded in a well-posed variational optimization problem, enabling us to draw on a body of work from convex and information-theoretic optimization~\citep{teboulle1992entropic, chretien2002kullback}, and ensuring a more principled treatment of smoothing in complex models.

A parallel line of work has explored variational inference (VI)~\citep{wainwright2008graphical} as a foundation for approximate smoothing. For instance, \citet{courts2021gaussian} proposed a variational Gaussian smoother that optimizes the evidence lower bound through constrained nonlinear optimization. Their formulation parameterizes pairwise Gaussian marginals independently and introduces equality constraints to enforce consistency over shared states. This results in a sparse batch optimization with linear complexity in the time horizon. Moreover, the updates are computed using an off-the-shelf constrained solver, without explicitly accounting for the information geometry of the Gaussian family~\citep{amari1998natural}. 

Related work by \citet{campbell2021online} developed an online variational method for filtering and parameter learning in state-space models. Their approach optimizes an evidence lower bound using backward decompositions of the joint posterior approximation together with Bellman-type recursions for the objective and its gradients. This provides another connection between variational inference and recursive state estimation, although their focus is online filtering and learning rather than proximal Bayesian smoothing. 

In concurrent work, \citet{tronarp2025recursive} introduced a recursive variational framework that derives forward-backward algorithms similar to ours, albeit without entropic regularization. Their approach invokes dynamic programming directly to formulate variational state estimation through forward and backward value-function recursions. In contrast, our regularized recursions emerge from an entropic proximal update, which provides explicit control of the change between successive updates of the variational posterior.

Our framework is closely related to the exactly sparse Gaussian variational inference method of \citet{barfoot2020exactly}, which builds on \citet{opper2009variational}. Their method applies natural gradient updates to a joint Gaussian posterior, exploiting block-tridiagonal structure to achieve linear complexity in the time horizon. In Appendix~\ref{app:natural-gradient}, we show that the forward- and reverse-Markov Gaussian updates are recursive natural gradient steps for fixed Fourier--Hermite coefficients. The difference is in how the step size is enforced and how the updates are computed. \citet{barfoot2020exactly} use a sparse batch optimization with a natural-gradient step size, whereas our framework provides forward, reverse, and hybrid recursive implementations and selects the damping parameter to satisfy a KL trust-region constraint.

\subsection{Approximate Bayesian Inference in Temporal Gaussian Processes}

There is a well-established connection between \emph{temporal} Gaussian processes (GP) and \emph{linear} state-space models~\citep{ohagan1978curve}. \citet{hartikainen2010kalman} showed that temporal GP models with Mat\'ern kernels can be reformulated exactly as linear-Gaussian SSMs. This enables using Kalman filtering and smoothing with linear time complexity to perform inference in temporal GPs, a significant improvement over the cubic complexity in standard GP inference.

Building on this foundation, \citet{chang2020fast} proposed a variational method that combines conjugate-computation variational inference~\citep{khan2017conjugate} with Rauch–Tung–Striebel recursions, enabling principled inference in models with linear dynamics and nonlinear measurement models. \citet{wilkinson2020state} extended this framework by unifying variational inference and expectation propagation within Kalman smoothing. \citet{wilkinson2023bayes} further generalized these approaches via the Bayes--Newton framework, which interprets VI, EP, and posterior linearization as optimization algorithms akin to Newton’s method. Despite these advances, a common critique of this class of algorithms is their \emph{ad hoc} treatment of the temporal dynamics. While inference is formulated from an optimization perspective, the temporal structure is handled implicitly using linear RTS smoothing, rather than being explicitly integrated into the inference objective.

In the continuous-time setting, several variational methods have been developed. \citet{archambeau2007variational} proposed one of the earliest approaches for inference in partially-observed diffusion processes with linear dynamics by approximating the posterior with a linear stochastic differential equation. This approximation can be viewed as the continuous-time limit of a Gauss--Markov factorization. \citet{ala2015gaussian} extended this line of work to nonlinear dynamics through sigma-point approximations. More recently, \citet{wildner2021moment} and \citet{verma2024variational} proposed natural-gradient-based generalizations of the method introduced by \citet{archambeau2007variational}. \citet{wildner2021moment} developed a moment-based recursive inference algorithm that reformulates the smoothing problem as an optimal control problem, thereby explicitly incorporating the dynamic structure of the setting. In contrast, \citet{verma2024variational} adopt a site-based approach inspired by \citet{minka2001expectation}, introducing a non-recursive algorithm that leverages iterative posterior linearization to handle nonlinear dynamics. 

Recently, \citet{bartosh2025sde} proposed a continuous-time simulation-free approach based on score matching techniques that scales to high dimensional settings. Finally, \citet{hu2026sing} propose a natural gradient method for inference in stochastic differential equations with a discretized Gaussian Markovian posterior and a block-tridiagonal precision matrix. This approach closely parallels the information-geometric sparse Gaussian  perspective of \citet{barfoot2020exactly}, albeit developed in the continuous-time setting.

\section{Numerical Evaluation}
\label{sec:experiments}

In this section, we empirically evaluate the proposed Bayesian inference algorithms on four representative benchmarks. The evaluation examines different posterior factorizations and approximation schemes and compares our methods with damped and undamped posterior-linearization smoothers and structured Gaussian variational methods. Additionally, we provide an open-source implementation of all algorithms at \url{https://github.com/hanyas/variational-iterated-smoothers}.

\paragraph{Linear-Gaussian.} We begin by empirically validating that the recursions are correct. For this, we work in a controlled setting that admits exact inference. We consider the canonical linear-Gaussian setting. Concretely, the latent state $x_k \in \mathbb{R}^{2}$ and the observation $y_k \in \mathbb{R}^{2}$ evolve, for $k = 1, \dots, T$, as
\begin{equation}
    x_0 \sim \mathcal{N}(\mu_0, \Sigma_0), \qquad x_k = A\,x_{k-1} + \omega_k, \qquad y_k = H\,x_k + \delta_k,
\end{equation}
with $\omega_k \sim \mathcal{N}(0, \Omega)$ and $\delta_k \sim \mathcal{N}(0, \Delta)$. The transition is a contractive rotation $A = 0.985 \cdot R(0.16)$, where $R(\theta)$ is the planar rotation matrix for $\theta$ radians. The observation map is $H = I_2$, where $I_2$ is the identity matrix in $\mathbb{R}^{2\times2}$. The noise covariances are $\Omega = 25 \times 10^{-4} \cdot I_2$ and $\Delta = 6.25 \times 10^{-2} \cdot I_2$, and the state is initialized at $\mu_0 = (4,0)^\top$ with $\Sigma_0 = 1 \times 10^{-2} \cdot I_2$. We simulate sequences of length $T = 100$.

\begin{figure}[t]
    \centering
    \pgfplotsset{
discard if not/.style 2 args={
  x filter/.code={%
    \edef\tempa{\thisrow{#1}}\edef\tempb{#2}%
    \ifx\tempa\tempb\else\def\pgfmathresult{nan}\fi
  }
},
}
\pgfplotsset{filter discard warning=false}

\begin{tikzpicture}
\begin{groupplot}[
  group style={group size=2 by 2, horizontal sep=1.9cm, vertical sep=0.3cm},
  width=0.47\linewidth, height=5cm,
  tick label style={font=\footnotesize}, label style={font=\small},
  title style={font=\small},
  legend style={font=\scriptsize, draw=none, fill=none}, legend cell align=left,
]

\nextgroupplot[ymode=log, xlabel={}, ylabel={Avg.\ KL to RTS},
               title={(a) Convergence to RTS}, xtick=\empty, xmin=-1, xmax=16]
  \foreach[count=\j from 0] \e/\lbl in
      {10.0/10, 20.0/20, 40.0/40, 80.0/80, 160.0/160} {
    \pgfmathtruncatemacro{\blk}{80 - 10*\j}
    \edef\opts{black!\blk, mark=*, mark size=0.7pt, line width=1.2pt, discard if not={eps}{\e}}%
    \expandafter\addplot\expandafter[\opts]
      table[col sep=comma, x=iter, y expr={max(\thisrow{kl_rts}, 1e-3)}]{figures/results_lg_proximal.csv};
  }
  \def\bx{0.52}\def\by{0.92}      
  \pgfmathsetmacro{\cw}{0.09}     
  \pgfmathsetmacro{\ch}{0.05}     
  \pgfmathsetmacro{\boxa}{\bx-0.035}\pgfmathsetmacro{\boxb}{\by-0.10}
  \pgfmathsetmacro{\boxc}{\bx+5*\cw+0.04}\pgfmathsetmacro{\boxd}{\by+\ch+0.075}
  \edef\bg{\noexpand\fill[white, opacity=0.25] (rel axis cs:\boxa,\boxb) rectangle (rel axis cs:\boxc,\boxd);}\bg
  \foreach[count=\c from 0] \lbl in {10,20,40,80,160}{
    \pgfmathtruncatemacro{\g}{80-10*\c}
    \pgfmathsetmacro{\xa}{\bx+\cw*\c}
    \pgfmathsetmacro{\xb}{\bx+\cw*(\c+1)}
    \pgfmathsetmacro{\xm}{\bx+\cw*(\c+0.5)}
    \pgfmathsetmacro{\yt}{\by+\ch}\pgfmathsetmacro{\yl}{\by-0.01}
    \edef\cell{\noexpand\fill[black!\g] (rel axis cs:\xa,\by) rectangle (rel axis cs:\xb,\yt);
               \noexpand\node[font=\noexpand\tiny, anchor=north, inner sep=1pt]
                 at (rel axis cs:\xm,\yl) {\lbl};}%
    \cell
  }
  \pgfmathsetmacro{\fx}{\bx+5*\cw}\pgfmathsetmacro{\fy}{\by+\ch}
  \pgfmathsetmacro{\ex}{\bx+2.5*\cw}\pgfmathsetmacro{\ey}{\by+\ch+0.005}
  \edef\frm{\noexpand\draw[line width=0.3pt] (rel axis cs:\bx,\by) rectangle (rel axis cs:\fx,\fy);}\frm

\nextgroupplot[xlabel={}, ylabel={Posterior mean $m_0$},
               title={(b) Mean interpolation}, legend pos=north east, xtick=\empty]
  \foreach \i in {0,...,22} {
    \pgfmathtruncatemacro{\blk}{75 - round(63*\i/22)}
    \edef\opts{black!\blk, line width=0.5pt, forget plot, discard if not={bi}{\i}}%
    \expandafter\addplot\expandafter[\opts]
      table[col sep=comma, x=k, y=m0]{figures/results_lg_interpolation.csv};
  }
  \addplot[black, line width=1.2pt, discard if not={series}{rts}]
    table[col sep=comma, x=k, y=m0]{figures/results_lg_interpolation.csv};
  \addlegendentry{RTS ($\beta{=}0$)}
  \addplot[black, dashed, line width=1.2pt, discard if not={series}{init}]
    table[col sep=comma, x=k, y=m0]{figures/results_lg_interpolation.csv};
  \addlegendentry{Init ($\beta{\approx}1$)}

\nextgroupplot[ymode=log, xlabel={Iteration}, ylabel={Damping $\beta$}, xmin=-1, xmax=16]
  \foreach[count=\j from 0] \e/\lbl in
      {10.0/10, 20.0/20, 40.0/40, 80.0/80, 160.0/160} {
    \pgfmathtruncatemacro{\blk}{80 - 10*\j}
    \edef\opts{black!\blk, mark=*, mark size=0.7pt, line width=1.2pt, discard if not={eps}{\e}}%
    \expandafter\addplot\expandafter[\opts]
      table[col sep=comma, x=iter, y expr={\thisrow{damping}>0 ? \thisrow{damping} : 1e-3}]{figures/results_lg_proximal.csv};
  }

\nextgroupplot[xlabel={Time step $k$}, ylabel={Posterior mean $m_1$}]
  \foreach \i in {0,...,22} {
    \pgfmathtruncatemacro{\blk}{75 - round(63*\i/22)}
    \edef\opts{black!\blk, line width=0.5pt, forget plot, discard if not={bi}{\i}}%
    \expandafter\addplot\expandafter[\opts]
      table[col sep=comma, x=k, y=m1]{figures/results_lg_interpolation.csv};
  }
  \addplot[black, line width=1.2pt, forget plot, discard if not={series}{rts}]
    table[col sep=comma, x=k, y=m1]{figures/results_lg_interpolation.csv};
  \addplot[black, dashed, line width=1.2pt, forget plot, discard if not={series}{init}]
    table[col sep=comma, x=k, y=m1]{figures/results_lg_interpolation.csv};

\end{groupplot}
\end{tikzpicture}
    \vspace{-0.25cm}
    \caption{Linear-Gaussian problem. In this setting, the Rauch--Tung--Striebel (RTS) posterior is exact, jointly Gaussian, and lies in the variational family. \textbf{(a)} Trust-region convergence for several trust-region radii $\varepsilon$. For each radius, the iterated FEVS approaches the true RTS posterior as the damping $\beta$ vanishes. Larger values of $\varepsilon$ permit larger updates and reach the posterior in fewer iterations. \textbf{(b)} Geometric $\beta$-interpolation. On a two-dimensional oscillator, the damped update interpolates between the exact posterior mean ($\beta=0$, solid) and the initialization ($\beta \approx 1$, dashed).}    
    \vspace{-0.30cm}
    \label{fig:linear-gaussian}
\end{figure}

Because both the dynamics and the likelihood are linear-Gaussian, the joint distribution over $(x_{0:\T}, y_{1:\T})$ is Gaussian and the smoothing posterior $p(x_{0:\T} \given y_{1:\T})$ is available in closed form through the Rauch--Tung--Striebel (RTS) recursion. The Gauss--Markov variational family contains this posterior, so there is no variational approximation error. We use this property to verify, in turn, that the recursions are correct and that the entropic iterations converge to the exact posterior.

For $\beta > 0$, the trust region governs how quickly the iterate approaches the true target, from an arbitrary initialization. Figure~\ref{fig:linear-gaussian} illustrates this behavior for the forward-Markov variational smoother (FEVS) with a generalized statistical linear regression (GSLR) approximation. Fourier--Hermite (FH) yields the same results because both approximation schemes represent the linear-Gaussian model exactly. The reverse-Markov and hybrid factorizations also yield indistinguishable results. Panel~(a) shows that, for every trust-region radius $\varepsilon$, the constrained iteration reduces the average marginal KL divergence to the true RTS posterior to zero, with larger radii converging in fewer iterations. Panel~(b) visualizes the geometric interpolation, showing how the posterior mean evolves from the initialization ($\beta \approx 1$) toward the exact posterior ($\beta = 0$).

\paragraph{Stochastic Volatility.} Having established correctness when the posterior is exact, we turn to a nonlinear, non-Gaussian setting, in which the quality of approximations matters significantly. We consider a stochastic volatility model~\citep{kim1998stochastic}, in which a latent log-volatility $x_k \in \mathbb{R}$ drives the variance of a zero-mean observation $y_k \in \mathbb{R}$. For $k = 1, \dots, T$,
\begin{equation}
    x_0 \sim \mathcal{N} (\mu, \sigma^2 /(1-\phi^2) ), \qquad
    x_k = \mu + \phi\,(x_{k-1} - \mu) + \sigma\,\rho_k, \qquad
    y_k = \exp(x_k/2)\,\zeta_k,
\end{equation}
with $\rho_k, \zeta_k \sim \mathcal{N}(0,1)$. Equivalently, $y_k \given x_k \sim \mathcal{N}(0,\, e^{x_k})$. We set $\mu = -0.5$, $\phi = 0.98$, sweep over $\sigma \in [0.1, 0.3]$, and simulate sequences of length $T = 1000$.

Here, the observation is conditionally Gaussian but informative only through its variance. Its mean $\mathbb{E}[y_k \given x_k] = 0$ is state-independent, and all information about $x_k$ enters through the second moment $e^{x_k}$. The parameter $\sigma$ controls the noise scale of the latent process. As $\sigma$ grows, the log-volatility covers a wider range, making the nonlinearity in the likelihood more significant and the inference problem more challenging.

This highlights a fundamental difference between the two approximation regimes. GSLR linearizes the observation by matching the conditional mean $\mathbb{E}[y_k \given x_k] = 0$, which is constant in $x_k$. The resulting linear-Gaussian surrogate has observation matrix $H = 0$ and is therefore entirely uninformative, so the smoother ignores the data and reproduces the prior. The Fourier--Hermite (FH) variant, in contrast, forms a quadratic approximation to the log-likelihood and retains the information carried by the observation variance. Panel~(c) in Figure~\ref{fig:stochastic-volatility} illustrates this issue on a representative trajectory. The GSLR variant collapses to the stationary prior dynamics, whereas the FH variant tracks the latent log-volatility with adequate uncertainty bands.

Panels~(a) and (b) in Figure~\ref{fig:stochastic-volatility} quantify this effect. For this comparison, we use the reverse-Markov variational smoother (REVS) with a third-order Gauss--Hermite numerical integration rule, noting that the forward- and hybrid-Markov variants give nearly identical outcomes. We report the negative log posterior density (NLPD), defined as the log score of the true latent trajectory $x_{0:\T}^\star$ under the approximate posterior, together with the root mean squared error (RMSE), which measures the error with respect to $x_{0:\T}^\star$. Both metrics degrade more gradually for the FH variant than for the GSLR variant, highlighting the advantage of the FH approximation in capturing higher-order terms.

\begin{figure}[t]
    \centering
    \begin{tikzpicture}
  \pgfplotsset{
    every axis/.append style={
      tick label style={font=\footnotesize}, label style={font=\small},
      title style={font=\small},
      legend style={font=\scriptsize, draw=none, fill=none}, legend cell align=left,
    },
    sweepGSLR/.style={color=black!50, mark=square*, mark size=1.0pt, line width=1.0pt,
                      error bars/.cd, y dir=both, y explicit},
    sweepFH/.style={color=black, mark=*, mark size=1.0pt, line width=1.0pt,
                    error bars/.cd, y dir=both, y explicit},
  }

  \begin{axis}[name=rmseax, width=0.48\linewidth, height=4.5cm,
      xlabel={Parameter $\sigma$}, ylabel={RMSE},
      title={(a) Accuracy}, legend pos=north west]
    \addplot[sweepGSLR] table[col sep=comma, x=sigma, y=gslr_rmse, y error=gslr_rmse_std]{figures/results_sv_sweep_summary.csv};
    \addlegendentry{REVS + GSLR}
    \addplot[sweepFH] table[col sep=comma, x=sigma, y=fh_rmse, y error=fh_rmse_std]{figures/results_sv_sweep_summary.csv};
    \addlegendentry{REVS + FH}
  \end{axis}

  \begin{axis}[name=nllax, at={(rmseax.east)}, anchor=west, xshift=1.5cm,
      width=0.48\linewidth, height=4.5cm,
      xlabel={Parameter $\sigma$}, ylabel={NLPD},
      title={(b) Negative log score}, legend pos=north west]
    \addplot[sweepGSLR] table[col sep=comma, x=sigma, y=gslr_nll, y error=gslr_nll_std]{figures/results_sv_sweep_summary.csv};
    \addplot[sweepFH] table[col sep=comma, x=sigma, y=fh_nll, y error=fh_nll_std]{figures/results_sv_sweep_summary.csv};
  \end{axis}

  \begin{axis}[name=trajax, at={(rmseax.south west)}, anchor=north west, yshift=-1.5cm,
      width=0.95\linewidth, height=5cm,
      xmin=0, xmax=1000, enlarge x limits=false,
      xlabel={Time step $k$}, ylabel={Log-volatility $x_k$},
      title={(c) Inferred log-volatility},
      legend columns=3, legend style={font=\scriptsize, draw=none,
        fill=white, fill opacity=0.8, text opacity=1, at={(0.5,0.99)}, anchor=north}]
    \addplot[draw=none, forget plot, name path=glo]
      table[col sep=comma, x=k, y expr={\thisrow{gslr_mean}-2*\thisrow{gslr_std}}]{figures/results_sv_trajectory.csv};
    \addplot[draw=none, forget plot, name path=ghi]
      table[col sep=comma, x=k, y expr={\thisrow{gslr_mean}+2*\thisrow{gslr_std}}]{figures/results_sv_trajectory.csv};
    \addplot[black!8, forget plot] fill between[of=glo and ghi];
    \addplot[draw=none, forget plot, name path=flo]
      table[col sep=comma, x=k, y expr={\thisrow{fh_mean}-2*\thisrow{fh_std}}]{figures/results_sv_trajectory.csv};
    \addplot[draw=none, forget plot, name path=fhi]
      table[col sep=comma, x=k, y expr={\thisrow{fh_mean}+2*\thisrow{fh_std}}]{figures/results_sv_trajectory.csv};
    \addplot[black!20, forget plot] fill between[of=flo and fhi];
    \addplot[black, line width=0.8pt] table[col sep=comma, x=k, y=x_true]{figures/results_sv_trajectory.csv};
    \addlegendentry{true $x_k$}
    \addplot[black!75, densely dashed, line width=1.0pt]
      table[col sep=comma, x=k, y=fh_mean]{figures/results_sv_trajectory.csv};
    \addlegendentry{Mean via REVS + FH}
    \addplot[black!45, dotted, line width=1.0pt]
      table[col sep=comma, x=k, y=gslr_mean]{figures/results_sv_trajectory.csv};
    \addlegendentry{Mean via REVS + GSLR}
  \end{axis}
  \path ([xshift=-1.3cm, yshift=-2.5cm]trajax.south west) rectangle (nllax.north east);
\end{tikzpicture}
    \vspace{-0.35cm}
    \caption{Stochastic volatility problem. \textbf{(a)} Root mean squared error (RMSE) and \textbf{(b)} negative log posterior density (NLPD) of the true log-volatility as a function of $\sigma$. Curves show the mean $\pm 1$ standard deviation over $10$ trials. The variant with generalized statistical linear regression (GSLR) incurs larger errors that grow rapidly with $\sigma$, whereas the Fourier--Hermite (FH) variant degrades more gradually. \textbf{(c)} An example trajectory for $T=1000$. The GSLR variant follows the stationary prior with constant mean and variance, whereas the FH variant tracks the log-volatility while maintaining reasonable uncertainty bands.}
    \vspace{-0.25cm}
    \label{fig:stochastic-volatility}
\end{figure}

\paragraph{Cubic Sensor.} Next, we consider the cubic sensor problem~\citep{katayama2013equivalent, tronarp2025recursive}, in which a linear--Gaussian state is observed through a cubic measurement. For $k = 1, \dots, T$,
\begin{equation}
    x_0 \sim \mathcal{N}(\mu_0, \Sigma_0), \qquad
    x_k = \phi_0\, x_{k-1} + (1-\phi_0)\,\mu_0 + \omega_k, \qquad
    y_k = \rho\, x_k^3 + \delta_k,
\end{equation}
with $\omega_k \sim \mathcal{N}(0, \Omega)$ and $\delta_k \sim \mathcal{N}(0, \Delta)$, where $\Omega = (1-\phi_0^2)\,\Sigma_0$. We set $\mu_0 = 0.4$, $\Sigma_0 = 0.36$, $\phi_0 = 0.95$, $\rho = 1$, and $\Delta = 1$, and simulate trajectories of length $T = 4096$.

Although the dynamics are linear, the cubic sensor implies an observation log-likelihood with a degree-six polynomial in the state, $\log h(x_k) = \sum_{m=0}^{6} \ell^{(m)}_k\, x_k^m$, which presents a challenging nonlinearity. 

In this evaluation, we report results from the hybrid-Markov variational smoother (HEVS) with a cubature numerical integration rule. The forward- and reverse-Markov variants produce comparable performance. Figure~\ref{fig:cubic-sensor} shows the evidence lower bound (ELBO) and the negative log posterior density across iterations. Both HEVS variants, based on FH and GSLR approximations, exhibit stable ELBO convergence. However, the GSLR variant attains a worse NLPD than FH. In contrast, the undamped IPLS baseline~\citep{garcia2016iterated, tronarp2018iterative}, which lacks an explicit mechanism to control the step size, overshoots and oscillates between two solutions. The Levenberg--Marquardt regularized IPLS (LM-IPLS) baseline~\citep{lindqvist2021posterior} dampens the oscillation, but still produces a lower ELBO and worse NLPD than HEVS. The ELBO is estimated using a second-order Fourier--Hermite expansion together with a fifth-order Gauss--Hermite quadrature rule.

\begin{figure}[t]
  \centering
  \pgfplotsset{
    cubic curve/.style={mark size=1.1pt, mark repeat=5, line width=1.0pt},
    fh one/.style={cubic curve, solid, color=black!50, mark=*, mark size=1.5pt},
    fh two/.style={cubic curve, solid, color=black!90, mark=*, mark size=1.5pt},
    gslr one/.style={cubic curve, dashed, color=black!50, mark=triangle*, mark options={solid}, mark size=2pt},
    gslr two/.style={cubic curve, dashed, color=black!90, mark=triangle*, mark options={solid}, mark size=2pt},
    ipls/.style={cubic curve, solid, color=black!25, mark=x, mark size=1.7pt},
    lm ipls/.style={cubic curve, dotted, color=black!65},
}

\begin{tikzpicture}
  \begin{groupplot}[
      group style={group size=2 by 1, horizontal sep=1.7cm},
      width=8cm, height=5.5cm,
      xlabel={Iteration}, xmin=-3, xmax=83,
      tick label style={font=\small}, label style={font=\small}, title style={font=\small},
    ]
    \nextgroupplot[
        title={(a) Evidence lower bound},
        ylabel={ELBO},
        ymin=-12900, ymax=-5500,
        scaled y ticks=base 10:-3,          
        legend style={font=\footnotesize, draw=none, fill=none, at={(0.97,0.03)}, anchor=south east, legend columns=1, legend cell align=left},
      ]
      \addplot[ipls]      table [x index=0, y index=5, col sep=comma] {figures/results_cubic_elbo_hyb.csv}; \addlegendentry{IPLS}
      \addplot[lm ipls]   table [x index=0, y index=6, col sep=comma] {figures/results_cubic_elbo_hyb.csv}; \addlegendentry{LM-IPLS}
      \addplot[fh one]    table [x index=0, y index=1, col sep=comma] {figures/results_cubic_elbo_hyb.csv}; \addlegendentry{HEVS + FH ($\varepsilon{=}1$)}
      \addplot[fh two]    table [x index=0, y index=2, col sep=comma] {figures/results_cubic_elbo_hyb.csv}; \addlegendentry{HEVS + FH ($\varepsilon{=}5$)}
      \addplot[gslr one]  table [x index=0, y index=3, col sep=comma] {figures/results_cubic_elbo_hyb.csv}; \addlegendentry{HEVS + GSLR ($\varepsilon{=}1$)}
      \addplot[gslr two]  table [x index=0, y index=4, col sep=comma] {figures/results_cubic_elbo_hyb.csv}; \addlegendentry{HEVS + GSLR ($\varepsilon{=}5$)}

    \nextgroupplot[
        title={(b) Negative log score},
        ylabel={NLPD},
        ymin=0.15, ymax=1.0,
      ]
      \addplot[fh one]    table [x index=0, y index=1, col sep=comma] {figures/results_cubic_nlpd_hyb.csv};
      \addplot[fh two]    table [x index=0, y index=2, col sep=comma] {figures/results_cubic_nlpd_hyb.csv};
      \addplot[gslr one]  table [x index=0, y index=3, col sep=comma] {figures/results_cubic_nlpd_hyb.csv};
      \addplot[gslr two]  table [x index=0, y index=4, col sep=comma] {figures/results_cubic_nlpd_hyb.csv};
      \addplot[ipls]      table [x index=0, y index=5, col sep=comma] {figures/results_cubic_nlpd_hyb.csv};
      \addplot[lm ipls]   table [x index=0, y index=6, col sep=comma] {figures/results_cubic_nlpd_hyb.csv};
  \end{groupplot}
\end{tikzpicture}
  \vspace{-0.25cm}
  \caption{Cubic sensor problem. \textbf{(a)} Evidence lower bound (ELBO) and \textbf{(b)} negative log posterior density (NLPD) over iterations for the hybrid-Markov variational smoother (HEVS), using Fourier--Hermite (FH) and generalized statistical linear regression (GSLR) approximations, together with the undamped iterated posterior linearization smoother (IPLS) and its Levenberg--Marquardt regularized variant (LM-IPLS). Both HEVS variants improve the ELBO monotonically, while IPLS and LM-IPLS settle into limit cycles.}
  \vspace{-0.35cm}
  \label{fig:cubic-sensor}
\end{figure}

\paragraph{Differential-drive robot.} Finally, we consider the differential-drive model of \citet{courts2021gaussian}. The state $x=(q_1,q_2,q_3,p_1,p_2)^\top\in\mathbb{R}^5$ contains the planar position $(q_1,q_2)$, heading $q_3$, linear momentum $p_1$, and angular momentum $p_2$. For wheel forces $u=(u_1,u_2)^\top$, the continuous-time drift is
\begin{equation}
    f(x,u) =
    \begin{bmatrix}
        \cos(q_3)p_1/m \\[0.5ex]
        \sin(q_3)p_1/m \\[0.5ex]
        p_2/(J+ml^2) \\[0.5ex]
        -r_1p_1/m - mlp_2^2/(J+ml^2)^2 + u_1 + u_2 \\[0.5ex]
        (lp_1-r_2)p_2/(J+ml^2) + a(u_1-u_2)
    \end{bmatrix}.
\end{equation}
Here, $r_1=r_2=1$, $a=0.5$, $m=5$, $J=2$, and $l=0.15$. We use an Euler discretization scheme $x_k = x_{k-1} + \dif t\,f(x_{k-1},u_k) + \omega_k$, with $\dif t=0.1$, and noisy pose observations $y_k = Hx_k + \delta_k$, with $H=\begin{bmatrix}I_3 & 0_{3\times2}\end{bmatrix}$. The noise processes are Gaussian, so that $\omega_k\sim\mathcal{N}(0,\Omega)$ and $\delta_k\sim\mathcal{N}(0,\Delta)$, with $\Omega=10^{-3} \times \operatorname{diag}(1,1,1.75,1,1)$ and $\Delta=\operatorname{diag}(0.1^2,0.1^2,0.035^2)$. The prior is $\mathcal{N}(0,\Sigma_0)$ with $\Sigma_0 = \operatorname{diag}(0.1^2,0.1^2,0.1^2,0.5^2,0.2^2)$. We consider observation sequences of length $T=2000$.

In this example, all Gaussian expectations in the ELBO admit closed-form expressions, including expectations that determine the coefficients of our Fourier--Hermite approximation. We compare our reverse-Markov variational smoother (REVS) with the approaches from~\citet{barfoot2020exactly} and \citet{courts2021gaussian}. The sparse batch method of \citet{barfoot2020exactly} can be understood as a structured natural-gradient variant of our approach that exploits the block-tridiagonal Gaussian precision. The formulation of \citet{courts2021gaussian}, while also optimizing the ELBO, parameterizes the pairwise smoothing marginals and enforces marginal consistency with explicit constraints by using an off-the-shelf optimizer.

Figure~\ref{fig:differential-drive} compares the methods under two initialization scenarios. The first uses the posterior obtained from a linearized RTS smoothing pass, while the second uses the linearized prior process. We apply the same ELBO convergence tolerance to all methods. The trust-region radii used by REVS and the step sizes used by \citet{barfoot2020exactly} influence their convergence speed. Starting from the prior leads to slower progress in both methods, but they still approach the same solution. In contrast, the method of \citet{courts2021gaussian} makes little progress and fails to satisfy its marginal consistency constraints. One possible explanation is that the optimizer struggles to enforce all equality constraints for such a long sequence. This appears to make the optimization more sensitive to initialization. Our method and that of~\citet{barfoot2020exactly} do not require such constraints, they inherently enforce consistency.

\begin{figure}[t]
    \centering
    \pgfplotsset{
    drive curve/.style={line width=1.0pt, mark repeat=5, mark size=1.4pt},
    drive revs five/.style={drive curve, solid, color=black, mark=*},
    drive revs twentyfive/.style={drive curve, solid, color=black!50, mark=*},
    drive barfoot quarter/.style={drive curve, dashed, color=black!50, mark=square*, mark options={solid}},
    drive barfoot half/.style={drive curve, dashed, color=black!90, mark=square*, mark options={solid}},
    drive courts/.style={drive curve, dotted, color=black!65, mark=triangle*, mark options={solid}},
}

\begin{tikzpicture}
  \begin{groupplot}[
      group style={group size=2 by 2, horizontal sep=1.7cm, vertical sep=0.3cm},
      width=8cm,
      tick label style={font=\small},
      label style={font=\small},
      title style={font=\small},
      unbounded coords=jump,
    ]
    \nextgroupplot[
        title={(a) RTS Initialization},
        height=5.5cm,
        xmin=-0.5, xmax=15.5,
        restrict x to domain=0:15,
        ymin=5586, ymax=5608,
        ylabel={ELBO},
        xtick={0,3,...,15},
        xticklabels=\empty,
        legend columns=1,
        legend style={
          font=\scriptsize,
          draw=none,
          fill=none,
          at={(0.99,0.03)},
          anchor=south east,
        },
        legend cell align=left,
      ]
      \addplot[drive revs five, mark repeat=3] table[x=iteration, y=REVS_eps5, col sep=comma] {figures/results_drive_elbo_obs.csv};
      \addlegendentry{REVS + FH ($\varepsilon{=}5$)}
      \addplot[drive revs twentyfive, mark repeat=3] table[x=iteration, y=REVS_eps25, col sep=comma] {figures/results_drive_elbo_obs.csv};
      \addlegendentry{REVS + FH ($\varepsilon{=}25$)}
      \addplot[drive courts, mark repeat=3] table[x=iteration, y=Courts, col sep=comma] {figures/results_drive_elbo_obs.csv};
      \addlegendentry{Courts et al.\ (2021)}
      \addplot[drive barfoot quarter, mark repeat=3] table[x=iteration, y=Barfoot_alpha025, col sep=comma] {figures/results_drive_elbo_obs.csv};
      \addlegendentry{Barfoot et al.\ (2020) ($\tau{=}0.25$)}
      \addplot[drive barfoot half, mark repeat=3] table[x=iteration, y=Barfoot_alpha05, col sep=comma] {figures/results_drive_elbo_obs.csv};
      \addlegendentry{Barfoot et al.\ (2020) ($\tau{=}0.5$)}

    \nextgroupplot[
        title={(b) Prior Initialization},
        height=5.5cm,
        xmin=-2.5, xmax=77.5,
        restrict x to domain=0:75,
        ymin=-5.25, ymax=0.55,
        ylabel={ELBO},
        xtick={0,15,...,75},
        xticklabels=\empty,
        ytick={-5,-3,-1.0414,0},
        yticklabels={$-10^9$,$-10^7$,$-10^5$,$0$},
      ]
      \addplot[drive revs five] table[x=iteration, y expr={sign(\thisrow{REVS_eps5})*ln(1+abs(\thisrow{REVS_eps5})/10000)/ln(10)}, col sep=comma] {figures/results_drive_elbo_prior.csv};
      \addplot[drive revs twentyfive] table[x=iteration, y expr={sign(\thisrow{REVS_eps25})*ln(1+abs(\thisrow{REVS_eps25})/10000)/ln(10)}, col sep=comma] {figures/results_drive_elbo_prior.csv};
      \addplot[drive barfoot quarter] table[x=iteration, y expr={sign(\thisrow{Barfoot_alpha025})*ln(1+abs(\thisrow{Barfoot_alpha025})/10000)/ln(10)}, col sep=comma] {figures/results_drive_elbo_prior.csv};
      \addplot[drive barfoot half] table[x=iteration, y expr={sign(\thisrow{Barfoot_alpha05})*ln(1+abs(\thisrow{Barfoot_alpha05})/10000)/ln(10)}, col sep=comma] {figures/results_drive_elbo_prior.csv};
      \addplot[drive courts] table[x=iteration, y expr={sign(\thisrow{Courts})*ln(1+abs(\thisrow{Courts})/10000)/ln(10)}, col sep=comma] {figures/results_drive_elbo_prior.csv};

    \nextgroupplot[
        height=3.25cm,
        xmin=-0.5, xmax=15.5,
        restrict x to domain=0:15,
        ymin=1e-17, ymax=1e-1,
        ymode=log,
        xlabel={Iteration},
        ylabel={Residual},
        xtick={0,3,...,15},
      ]
      \addplot[drive courts, mark repeat=3] table[x=iteration, y=Courts_consistency, col sep=comma] {figures/results_drive_elbo_obs.csv};
      \addplot[black!45, dashed, domain=0:15] {1e-6};
      \node[font=\scriptsize, anchor=south east, fill=white, inner sep=1pt] at (axis cs:14.7,1e-6) {Threshold${=}10^{-6}$};

    \nextgroupplot[
        height=3.25cm,
        xmin=-2.5, xmax=77.5,
        restrict x to domain=0:75,
        ymin=1e-11, ymax=1e6,
        ymode=log,
        xlabel={Iteration},
        ylabel={Residual},
        xtick={0,15,...,75},
      ]
      \addplot[drive courts] table[x=iteration, y=Courts_consistency, col sep=comma] {figures/results_drive_elbo_prior.csv};
      \addplot[black!45, dashed, domain=0:75] {1e-6};
      \node[font=\scriptsize, anchor=south east, fill=white, inner sep=1pt] at (axis cs:73.5,1e-6) {Threshold${=}10^{-6}$};
  \end{groupplot}
\end{tikzpicture}
    \vspace{-0.3cm}
    \caption{Differential-drive problem. The top row shows ELBO progress starting from \textbf{(a)} an RTS-based initialization and \textbf{(b)} a prior-based initialization. The trust-region radii used by REVS and the step sizes used by~\citet{barfoot2020exactly} lead to different convergence speeds. The bottom row shows the consistency residual for~\citet{courts2021gaussian}. Under prior-based initialization, this method struggles to enforce marginal consistency and stalls, whereas REVS and the method of~\citet{barfoot2020exactly} converge.}
    \vspace{-0.40cm}
    \label{fig:differential-drive}
\end{figure}

\section{Discussion}
\label{sec:discussion}

We have presented a variational framework for approximate Bayesian inference in general state-space models. The main idea is to formulate inference as a sequence of entropic proximal updates over posterior distributions, subject to dynamic consistency constraints. This yields a dynamic optimization problem whose optimality conditions give rise to recursive forward-backward smoothing algorithms. The resulting KL-based trust regions regularize each update relative to the previous posterior approximation, leading to damped recursions and an adaptive step size mechanism. Our framework shows how different recursive smoothers arise from different factorizations of the variational posterior. Forward, reverse, and hybrid Gauss--Markov approximations each lead to distinct algorithms with linear complexity in the time horizon. For nonlinear and non-Gaussian models, the exact local potentials are generally intractable, so we enforce the recursive structure using generalized statistical linear regression and Fourier--Hermite moment matching. Overall, this proposed framework provides a principled bridge between variational inference, proximal optimization, and recursive Bayesian inference in state-space models.

Several limitations and extensions remain. The practical algorithms rely on local quadratic approximations, and their quality depends on the accuracy of the statistical approximation procedure. In nonlinear or multi-modal problems, a Gauss--Markov approximation may be insufficient to capture the posterior complexity. Future work could extend the framework to richer variational families. Furthermore, square-root implementations of the proposed recursions may improve numerical stability, especially for long time horizons~\citep{bierman2006factorization, yaghoobi2025parallel}. Another direction is to leverage these dynamic optimization ideas within probabilistic numerics~\citep{hennig2022probabilistic}, where differential equation solvers are formulated as Bayesian state estimation problems. This includes extensions to probabilistic ordinary differential equation solvers~\citep{tronarp2019probabilistic, bosch2024parallel} and to probabilistic solvers for time-dependent partial differential equations~\citep{kramer2022probabilistic, iqbal2024parallel}. 




\bibliography{main}
\bibliographystyle{tmlr}

\newpage

\appendix

\section{Proof of Proposition~\ref{prop:static-proximal}}\label{app:proof-static-proximal}
To find the critical point of \eqref{eq:static-proximal} with respect to distribution $q(x)$ given the constraints, we first construct the corresponding Lagrangian functional
\begin{equation}\label{app-eq:static-lagrangian}
    \mathcal{R}(q, \lambda, \alpha) = 
    \begin{aligned}[t]
        & \mathbb{E}_{q} \Big[ \log p(y, x) \Big] - \mathbb{E}_{q} \Big[ \log q(x) \Big] \\
        & + \lambda \left[ 1 - \int q(x) \dif x \right] + \alpha \left[ \varepsilon - \mathbb{D}_{\mathrm{KL}} \left[ q(x) \ggiven q^{[i]}(x) \right] \right],
    \end{aligned}
\end{equation}
where $\lambda$ and $\alpha \geq 0$ are the Lagrangian multipliers associated with the constraints. Next, we set the functional derivative of $\mathcal{R}(\cdot)$ with respect to $q(x)$ to zero
\begin{equation}
    \frac{\partial \mathcal{R}(q, \lambda, \alpha)}{\partial q(x)} = \log p(y, x) - \log q(x) - 1 - \lambda - \alpha \log q(x) + \alpha \log q^{[i]}(x) - \alpha \coloneqq 0,
\end{equation}
which in turn leads to the optimal iterate $q^{[i+1]}(x)$
\begin{equation}\label{app-eq:static-optimal}
    q^{[i+1]}(x) = 
    \begin{aligned}[t]
        & \Big[ \exp \left\{ \lambda + (1 + \alpha) \right\} \Big]^{-1 / (1 + \alpha)} \Big[ p(y \given x) \, p(x) \Big]^{1 / (1 + \alpha)} \Big[ q^{[i]}(x) \Big]^{\alpha / (1 + \alpha)} \\
        & \propto \Big[ p(y \given x) \, p(x) \Big]^{1 / (1 + \alpha)} \Big[ q^{[i]}(x) \Big]^{\alpha / (1 + \alpha)}.
    \end{aligned}
\end{equation}
Plugging the result from \eqref{app-eq:static-optimal} back into the Lagrangian \eqref{app-eq:static-lagrangian}, we get the dual problem
\begin{equation}\label{app-eq:static-dual}
    \minimize_{\lambda, \alpha} \quad \mathcal{G}(\lambda, \alpha) = \alpha \, \varepsilon + \lambda + (1 + \alpha) \int q^{[i+1]}(x) \dif x, \quad \subject \,\, \alpha \geq 0.
\end{equation}
If we compute and set the gradient of $\mathcal{G}(\cdot)$ with respect to $\lambda$ to zero, we get
\begin{align}\label{app-eq:static-normalizer}
    \lambda^{[i+1]} & = (1 + \alpha) \left[ -1 + \log \int \Big[ p(y \given x) \, p(x) \Big]^{1 / (1 + \alpha)} \Big[ q^{[i]}(x) \Big]^{\alpha / (1 + \alpha)} \dif x \right] \\
    & = ( 1 + \alpha) \left[ - 1 + \log \mathcal{Z}^{[i+1]}(\alpha) \right]
\end{align}
which when plugged back into \eqref{app-eq:static-optimal} leads to the normalized approximate Gibbs posterior
\begin{equation}\label{app-eq:static-optimal-normalized}
    q^{[i+1]}(x) = \left[ \mathcal{Z}^{[i+1]}(\beta) \right]^{-1} \Big[ p(y \given x) \, p(x) \Big]^{1 - \beta} \Big[ q^{[i]}(x) \Big]^{\beta},
\end{equation}
where we have defined $\beta \coloneqq \alpha / (1 + \alpha)$, so that $\beta = 0$ when $\alpha = 0$ and $\beta \rightarrow 1$ as $\alpha \rightarrow \infty$. By using the results from \eqref{app-eq:static-normalizer} and \eqref{app-eq:static-optimal-normalized}, we can rewrite the dual problem \eqref{app-eq:static-dual} as follows
\begin{equation}\label{app-eq:static-dual-divergence}
    \minimize_{\beta} \quad \mathcal{G}(\beta) = \frac{\beta \varepsilon}{1 - \beta} + \frac{1}{1 - \beta} \log \mathcal{Z}^{[i+1]}(\beta), \quad \subject \,\, 0 \leq \beta < 1.
\end{equation}

\section{Proof of Proposition~\ref{prop:forward-markov}}
\label{app:forward-markov-proof}
Given the forward-Markov decomposition from Assumption~\ref{asm:forward-markov}, we can factorize problem~\eqref{eq:smoothing-proximal} over time. Starting with the ELBO objective, we can write
\begin{align}
    \mathcal{L}(\rh{q}) & = \mathbb{E}_{\, \rh{q}} \Big[ \log p( x_{0:\T}, y_{1:\T}) \Big] - \mathbb{E}_{\, \rh{q}} \Big[ \log \rh{q}(x_{0:\T}) \Big] \\
    & = 
    \begin{aligned}[t]
        & \int \rhk{q}{0}(x_{0}) \Big[ \log p(x_{0}) - \log \rhk{q}{0}(x_{0}) \Big] \dif x_{0} + \sum_{k=1}^{T} \int \rhk{q}{k}(x_{k}) \log h_{k}(y_{k} \given x_{k}) \dif x_{k} \\
        & + \sum_{k=0}^{T-1} \int \rhk{q}{k}(x_{k}) \, \rhk{q}{k}(x_{k+1} \given x_{k}) \Big[ \log f_{k}(x_{k+1} \given x_{k}) - \log \rhk{q}{k}(x_{k+1} \given x_{k}) \Big] \dif x_{k} \dif x_{k+1}.
    \end{aligned}
\end{align}
Additionally, the Kullback--Leibler divergence factorizes forward as follows
\begin{align}
    \varepsilon & \geq \mathbb{D}_{\mathrm{KL}} \left[ \rh{q}(x_{0:\T}) \ggiven \rhi{q}{i}(x_{0:\T}) \right] \\
    \varepsilon & \geq \int \rhk{q}{0}(x_{0}) \prod_{k=0}^{T-1} \rhk{q}{k}(x_{k+1} \given x_{k}) \log \frac{\rhk{q}{0}(x_{0}) \prod_{k=0}^{\T-1} \rhk{q}{k}(x_{k+1} \given x_{k})}{\rhki{q}{0}{i}(x_{0}) \prod_{k=0}^{\T-1} \rhki{q}{k}{i}(x_{k+1} \given x_{k})} \dif x_{0:\T} \\
    \varepsilon & \geq
    \begin{aligned}[t]
        & \int \rhk{q}{0}(x_{0}) \log \frac{\rhk{q}{0}(x_{0})}{\rhki{q}{0}{i}(x_{0})} \dif x_{0} \\
        & + \sum_{k=0}^{T-1} \int \rhk{q}{k}(x_{k}) \int \rhk{q}{k}(x_{k+1} \given x_{k}) \log \frac{ \rhk{q}{k}(x_{k+1} \given x_{k})}{ \rhki{q}{k}{i}(x_{k+1} \given x_{k})} \dif x_{k+1} \dif x_{k}.
    \end{aligned}
\end{align}
Finally, the normalization constraint also factorizes to
\begin{equation}
    1 = \int \rhk{q}{0}(x_{0}) \dif x_{0}, \quad 1 = \int \rhk{q}{k}(x_{k+1} \given x_{k}) \dif x_{k+1}, \quad \forall x_{k}, \,\, 0 \leq k < T.
\end{equation}
The information-theoretic proximal optimization problem \eqref{eq:smoothing-proximal} now has the form
\begin{align}
    & \maximize_{\substack{\rhk{q}{k}(x_{k+1} \given x_{k}), \\ \rhk{q}{0}(x_{0})}} & &
    \begin{aligned}[t]
        & \int \rhk{q}{0}(x_{0}) \Big[ \log p(x_{0}) - \log \rhk{q}{0}(x_{0}) \Big] \dif x_{0} + \sum_{k=1}^{T} \int \rhk{q}{k}(x_{k}) \log h_{k}(y_{k} \given x_{k}) \dif x_{k} \\
        & + \sum_{k=0}^{T-1} \int \rhk{q}{k}(x_{k}) \, \rhk{q}{k}(x_{k+1} \given x_{k}) \Big[ \log f_{k}(x_{k+1} \given x_{k}) - \log \rhk{q}{k}(x_{k+1} \given x_{k}) \Big] \dif x_{k} \dif x_{k+1},
    \end{aligned} \\
    & \subject \qquad & & \rhk{q}{k+1}(x_{k+1}) = \int \rhk{q}{k}(x_{k}) \, \rhk{q}{k}(x_{k+1} \given x_{k}) \dif x_{k}, \quad \forall \, x_{k+1}, \,\, 0 \leq k < T, \\
    & & & 1 = \int \rhk{q}{0}(x_{0}) \dif x_{0}, \quad 1 = \int \rhk{q}{k}(x_{k+1} \given x_{k}) \dif x_{k+1}, \quad \forall x_{k}, \,\, 0 \leq k < T, \\
    & & & \varepsilon \geq
    \begin{aligned}[t]
        & \int \rhk{q}{0}(x_{0}) \log \frac{\rhk{q}{0}(x_{0})}{\rhki{q}{0}{i}(x_{0})} \dif x_{0} \\
        & + \sum_{k=0}^{T-1} \int \rhk{q}{k}(x_{k}) \int \rhk{q}{k}(x_{k+1} \given x_{k}) \log \frac{ \rhk{q}{k}(x_{k+1} \given x_{k})}{ \rhki{q}{k}{i}(x_{k+1} \given x_{k})} \dif x_{k+1} \dif x_{k}.
    \end{aligned}
\end{align}
Next, we construct the Lagrangian functional $\rh{\mathcal{R}}(\rhk{q}{0}, \rhk{q}{k}, \rhk{\gamma}{0}, \rhk{\lambda}{k}, \rhkxx{V}{k}, \alpha)$
\begin{equation}\label{app-eq:forward-markov-lagrangian}
    \rh{\mathcal{R}}(\cdot) = 
    \begin{aligned}[t]
        & \int \rhk{q}{0}(x_{0}) \Big[ \log p(x_{0}) - \log \rhk{q}{0}(x_{0}) \Big] \dif x_{0} + \sum_{k=1}^{T} \int \rhk{q}{k}(x_{k}) \log h_{k}(y_{k} \given x_{k}) \dif x_{k} \\
        & + \sum_{k=0}^{T-1} \int \rhk{q}{k}(x_{k}) \, \rhk{q}{k}(x_{k+1} \given x_{k}) \Big[ \log f_{k}(x_{k+1} \given x_{k}) - \log \rhk{q}{k}(x_{k+1} \given x_{k}) \Big] \dif x_{k} \dif x_{k+1} \\
        & + \sum_{k=0}^{T-1} \int \rhkxx{V}{k+1}(x_{k+1}) \left[ \int \rhk{q}{k}(x_{k}) \, \rhk{q}{k}(x_{k+1} \given x_{k}) \dif x_{k} - \rhk{q}{k+1}(x_{k+1}) \right] \dif x_{k+1} \\
        & + \sum_{k=0}^{T-1} \int \rhk{\lambda}{k}(x_{k}) \left[ 1 - \int \rhk{q}{k}(x_{k+1} \given x_{k}) \dif x_{k+1} \right] \dif x_{k} + \rhk{\gamma}{0} \left[ 1 - \int \rhk{q}{0}(x_{0}) \dif x_{0} \right] \\
        & + \alpha 
        \begin{aligned}[t]
            \Bigg[ & \varepsilon - \int \rhk{q}{0}(x_{0}) \log \frac{\rhk{q}{0}(x_{0})}{\rhki{q}{0}{i}(x_{0})} \dif x_{0} \\
            & - \sum_{k=0}^{T-1} \int \rhk{q}{k}(x_{k}) \int \rhk{q}{k}(x_{k+1} \given x_{k}) \log \frac{ \rhk{q}{k}(x_{k+1} \given x_{k})}{ \rhki{q}{k}{i}(x_{k+1} \given x_{k})} \dif x_{k+1} \dif x_{k} \Bigg],
        \end{aligned}
    \end{aligned}
\end{equation}
where $\rhk{V}{k}(x_{k})$, $\rhk{\lambda}{k}(x_{k})$, $\rhk{\gamma}{0}$, and $\alpha \geq 0$ are Lagrangian multipliers. To find the critical point, we set the functional derivative with respect to $\rhk{q}{k}(x_{k+1} \given x_{k})$ to zero
\begin{equation}\label{app-eq:forward-markov-optimal-conditionals}
    \rhki{q}{k}{i+1}(x_{k+1} \given x_{k}) = 
    \begin{aligned}[t]
        & \Big[ \exp \left\{ \rhk{\lambda}{k}(x_{k}) / \rhk{q}{k}(x_{k}) + (1 + \alpha) \right\} \Big]^{-1 / (1 + \alpha)} \\ 
        & \times \Big[ \rhki{q}{k}{i}(x_{k+1} \given x_{k}) \Big]^{\alpha / (1 + \alpha)} \Big[ f_{k}(x_{k+1} \given x_{k}) \exp \left\{ \rhkxx{V}{k+1}(x_{k+1}) \right\} \Big]^{1/(1 + \alpha)}.
    \end{aligned}
\end{equation}
Additionally, we set the functional derivative with respect to $\rhk{q}{0}(x_{0})$ to zero and get
\begin{equation}\label{app-eq:forward-markov-optimal-marginal}
    \rhki{q}{0}{i+1}(x_{0}) = \Big[ \exp \Big\{ \rhk{\gamma}{0} + (1 + \alpha) \Big\} \Big]^{- 1 / (1 + \alpha)} \Big[ \rhki{q}{0}{i}(x_{0}) \Big]^{\alpha / (1 + \alpha)} \Big[ \exp \left\{ \rhkxx{V}{0}(x_{0}) \right\} \Big]^{1 / (1 + \alpha)},
\end{equation}
where, for convenience of notation, we define
\begin{equation}\label{app-eq:forward-markov-first-potential}
    \rhkxx{V}{0}(x_{0}) = \log p_{0}(x_{0}) + (1 + \alpha) \log \int \left[ \rhki{q}{0}{i}(x_{1} \given x_{0}) \right]^{\alpha / (1 + \alpha)} \Big[ f_{0}(x_{1} \given x_{0}) \exp \left\{ \rhkxx{V}{1}(x_{1}) \right\} \Big]^{1 / (1 + \alpha)} \dif x_{1}.
\end{equation}
Notice that it is not clear at this point that the solutions \eqref{app-eq:forward-markov-optimal-conditionals} and \eqref{app-eq:forward-markov-optimal-marginal} are normalized densities. Establishing that requires solving for the (functional) multipliers $\rhk{\lambda}{k}(x_{k})$ and $\rhk{\gamma}{0}$, which are associated with the normalization constraints. Plugging the solution \eqref{app-eq:forward-markov-optimal-conditionals} into the Lagrangian \eqref{app-eq:forward-markov-lagrangian} results in the dual functional
\begin{equation}\label{app-eq:forward-markov-dual}
    \rh{\mathcal{G}}(\cdot) = 
    \begin{aligned}[t]
        & \alpha \, \varepsilon + \int \rhk{q}{0}(x_{0}) \Big[ \log p(x_{0}) - \log \rhk{q}{0}(x_{0}) \Big] \dif x_{0} + \sum_{k=1}^{T} \int \rhk{q}{k}(x_{k}) \log h_{k}(y_{k} \given x_{k}) \dif x_{k} \\
        & + \alpha \int \rhk{q}{0}(x_{0}) \left[ \log \rhki{q}{0}{i}(x_{0}) - \log \rhk{q}{0}(x_{0}) \right] \dif x_{0} + \rhk{\gamma}{0} \left[ 1 - \int \rhk{q}{0}(x_{0}) \dif x_{0} \right] \\
        & + \sum_{k=0}^{T-1} \int \rhk{\lambda}{k}(x_{k}) \dif x_{k} - \sum_{k=0}^{T-1} \int \rhkxx{V}{k+1}(x_{k+1}) \rhk{q}{k+1}(x_{k+1}) \dif x_{k+1} \\
        & + (1 + \alpha) \sum_{k=0}^{T-1} \int \rhk{q}{k}(x_{k}) \int \rhki{q}{k}{i+1}(x_{k+1} \given x_{k}) \dif x_{k+1} \dif x_{k}.
    \end{aligned}
\end{equation}
Now, we can solve for the multipliers $\rhk{\lambda}{k}(x_{k})$ by setting the associated derivatives to zero
\begin{align}\label{app-eq:forward-markov-conditionals-normalizer}
    \rhki{\lambda}{k}{i+1}(x_{k}) & = 
    \begin{aligned}[t]
        (1 + \alpha) \, \rhk{q}{k}(x_{k}) \Big[ -1 & + \log \int \Big[ \rhki{q}{k}{i}(x_{k+1} \given x_{k}) \Big]^{\alpha / (1 + \alpha)} \\ 
        & \times \Big[ f_{k}(x_{k+1} \given x_{k})  \exp \left\{ \rhkxx{V}{k+1}(x_{k+1}) \right\} \Big]^{1 / (1 + \alpha)} \dif x_{k+1} \Big]
    \end{aligned} \\
    & = (1 + \alpha) \, \rhk{q}{k}(x_{k}) \Big[ -1 + \log \rhki{\psi}{k}{i+1}(x_{k}) \Big].
\end{align}
Plugging \eqref{app-eq:forward-markov-conditionals-normalizer} into \eqref{app-eq:forward-markov-optimal-conditionals} returns the normalized tilted conditional
\begin{equation}\label{app-eq:forward-markov-normalized-conditionals}
    \rhki{q}{k}{i+1}(x_{k+1} \given x_{k}) = 
    \begin{aligned}[t]
        \Big[ \rhki{\psi}{k}{i+1}(x_{k}) \Big]^{-1} & \Big[ \rhki{q}{k}{i}(x_{k+1} \given x_{k}) \Big]^{\beta} \\
        & \times \Big[ f_{k}(x_{k+1} \given x_{k}) \exp \left\{ \rhkxx{V}{k+1}(x_{k+1}) \right\} \Big]^{1 - \beta},
    \end{aligned}
\end{equation}
where we have defined $\beta \coloneqq \alpha / (1 + \alpha)$, so that $\beta = 0$ when $\alpha = 0$, and $\beta \rightarrow 1$ as $\alpha \rightarrow \infty$. Next, we use the results from \eqref{app-eq:forward-markov-optimal-marginal}, \eqref{app-eq:forward-markov-conditionals-normalizer}, and \eqref{app-eq:forward-markov-normalized-conditionals} to simplify the functional $\rh{\mathcal{G}}(\cdot)$ further
\begin{equation}
    \rh{\mathcal{G}}(\cdot) = 
    \begin{aligned}[t]
        & \alpha \, \varepsilon + \sum_{k=1}^{T} \int \rhk{q}{k}(x_{k}) \log h_{k}(y_{k} \given x_{k}) \dif x_{k} - \sum_{k=1}^{T} \int \rhkxx{V}{k}(x_{k}) \rhk{q}{k}(x_{k}) \dif x_{k} \\
        & + (1 + \alpha) \sum_{k=1}^{T-1} \int \rhk{q}{k}(x_{k}) \log \rhki{\psi}{k}{i+1}(x_{k}) \dif x_{k} + \rhk{\gamma}{0} + (1 + \alpha) \int \rhki{q}{0}{i+1}(x_{0}) \dif x_{0},
    \end{aligned}
\end{equation}
and solve for the optimal multiplier $\rhk{\gamma}{0}$ by zeroing the associated derivative, leading to
\begin{align}\label{app-eq:forward-markov-marginal-normalizer}
    \rhki{\gamma}{0}{i+1} & = (1 + \alpha) \left[ -1 + \log \int \Big[ \rhki{q}{0}{i}(x_{0}) \Big]^{\alpha / (1 + \alpha)} \Big[ \exp \left\{ \rhkxx{V}{0}(x_{0}) \right\} \Big]^{1 / (1 + \alpha)} \dif x_{0} \right] \\
    & = (1 + \alpha) \left[ -1 + \log \rhki{\mathcal{Z}}{0}{i+1} \right],
\end{align}
which after plugging into \eqref{app-eq:forward-markov-optimal-marginal} delivers
\begin{equation}\label{app-eq:forward-markov-normalized-marginal}
    \rhki{q}{0}{i+1}(x_{0}) = \Big[ \rhki{\mathcal{Z}}{0}{i+1} \Big]^{-1} \Big[ \rhki{q}{0}{i}(x_{0}) \Big]^{\beta} \Big[ \exp \left\{ \rhkxx{V}{0}(x_{0}) \right\} \Big]^{1 - \beta}.
\end{equation}
Plugging \eqref{app-eq:forward-markov-normalized-marginal} and \eqref{app-eq:forward-markov-marginal-normalizer} back into $\rh{\mathcal{G}}(\cdot)$ leads to further simplification of the dual
\begin{equation}
    \rh{\mathcal{G}}(\cdot) = 
    \begin{aligned}[t]
        & \alpha \, \varepsilon + \sum_{k=1}^{T} \int \rhk{q}{k}(x_{k}) \log h_{k}(y_{k} \given x_{k}) \dif x_{k} - \sum_{k=1}^{T} \int \rhkxx{V}{k}(x_{k}) \rhk{q}{k}(x_{k}) \dif x_{k} \\
        & + (1 + \alpha) \sum_{k=1}^{T-1} \int \rhk{q}{k}(x_{k}) \log \rhki{\psi}{k}{i+1}(x_{k}) \dif x_{k} + (1 + \alpha) \log \rhki{\mathcal{Z}}{0}{i+1}.
    \end{aligned}
\end{equation}
Finally, to find the optimal potentials, we zero the derivative of $\rh{\mathcal{G}}(\cdot)$ with respect to $\rhk{q}{k}(x_{k}), \forall \, 0 < k \leq T$. Combined with the results from \eqref{app-eq:forward-markov-first-potential} and \eqref{app-eq:forward-markov-conditionals-normalizer}, we can write
\begin{equation}\label{app-eq:forward-markov-optimal-potentials}
    \rhkixx{V}{k}{i+1}(x_{k}) = 
    \begin{cases}
        \, \log h_{\T}(y_{\T} \given x_{\T}) & \mathrm{if} \,\, k = T, \\[0.5em]
        \, \log h_{k}(y_{k} \given x_{k}) + 1 / (1 - \beta) \log \rhki{\psi}{k}{i+1}(x_{k}) & \mathrm{if} \,\, 0 < k < T, \\[0.5em]
        \, \log p_{0}(x_{0}) + 1 / (1 - \beta) \log \rhki{\psi}{0}{i+1}(x_{0}) & \,\, \mathrm{if} \,\, k = 0,
    \end{cases}
\end{equation}
which, when plugged back into the dual $\rh{\mathcal{G}}(\cdot)$, leads to the final simplification
\begin{equation}
    \rh{\mathcal{G}}(\beta) = \frac{\beta \varepsilon}{1 - \beta} + \frac{1}{1 - \beta} \log \rhki{\mathcal{Z}}{0}{i+1}(\beta).
\end{equation}

\section{Proof of Proposition~\ref{prop:reverse-markov}}
\label{app:reverse-markov-proof}
This proof follows a similar scheme to that in Section~\ref{app:forward-markov-proof}. However, the reverse-Markov decomposition results in novel recursions. Starting from the reverse-Markov decomposition in Assumption~\ref{asm:reverse-markov}, we factorize problem~\eqref{eq:smoothing-proximal} over time. Thus, we can write the ELBO as
\begin{align}
    \mathcal{L}(\lh{q}) & = \mathbb{E}_{\, \lh{q}} \Big[ \log p( x_{0:\T}, y_{1:\T}) \Big] - \mathbb{E}_{\, \lh{q}} \Big[ \log \lh{q}(x_{0:\T}) \Big] \\
    & = 
    \begin{aligned}[t]
        & \int \lhk{q}{0}(x_{0}) \log p(x_{0}) \dif x_{0} - \int \lhk{q}{\T}(x_{\T}) \log \lhk{q}{\T}(x_{\T}) \dif x_{\T} + \sum_{k=1}^{T} \int \lhk{q}{k}(x_{k}) \log h_{k}(y_{k} \given x_{k}) \dif x_{k} \\
        & + \sum_{k=1}^{T} \int \lhk{q}{k}(x_{k}) \, \lhk{q}{k}(x_{k-1} \given x_{k}) \Big[ \log f_{k-1}(x_{k} \given x_{k-1}) - \log \lhk{q}{k}(x_{k-1} \given x_{k}) \Big] \dif x_{k-1} \dif x_{k}.
    \end{aligned}
\end{align}
Furthermore, the Kullback--Leibler divergence factorizes in reverse to
\begin{align}
    \varepsilon & \geq \mathbb{D}_{\mathrm{KL}} \left[ \lh{q}(x_{0:\T}) \ggiven \lhi{q}{i}(x_{0:\T}) \right] \\
    \varepsilon & \geq \int \lhk{q}{\T}(x_{\T}) \prod_{k=1}^{T} \lhk{q}{k}(x_{k-1} \given x_{k}) \log \frac{\lhk{q}{\T}(x_{\T}) \prod_{k=1}^{\T} \lhk{q}{k}(x_{k-1} \given x_{k})}{\lhki{q}{\T}{i}(x_{\T}) \prod_{k=1}^{\T} \lhki{q}{k}{i}(x_{k-1} \given x_{k})} \dif x_{0:\T} \\
    \varepsilon & \geq
    \begin{aligned}[t]
        & \int \lhk{q}{\T}(x_{\T}) \log \frac{\lhk{q}{\T}(x_{\T})}{\lhki{q}{\T}{i}(x_{\T})} \dif x_{\T} \\
        & + \sum_{k=1}^{T} \int \lhk{q}{k}(x_{k}) \int \lhk{q}{k}(x_{k-1} \given x_{k}) \log \frac{\lhk{q}{k}(x_{k-1} \given x_{k})}{\lhki{q}{k}{i}(x_{k-1} \given x_{k})} \dif x_{k-1} \dif x_{k}
    \end{aligned}
\end{align}
Finally, the normalization constraint also factorizes to
\begin{equation}
    1 = \int \lhk{q}{\T}(x_{\T}) \dif x_{\T}, \quad 1 = \int \lhk{q}{k}(x_{k-1} \given x_{k}) \dif x_{k-1}, \quad \forall x_{k}, \,\, 0 < k \leq T.
\end{equation}
Now we can rewrite the optimization problem \eqref{eq:smoothing-proximal} as follows
\begin{align}
    & \maximize_{\substack{\lhk{q}{k}(x_{k-1} \given x_{k}), \\ \lhk{q}{\sT}(x_{\sT})}} & &
    \begin{aligned}[t]
        & \int \lhk{q}{0}(x_{0}) \log p(x_{0}) \dif x_{0} + \sum_{k=1}^{T} \int \lhk{q}{k}(x_{k}) \log h_{k}(y_{k} \given x_{k}) \dif x_{k} \\
        & + \sum_{k=1}^{T} \int \lhk{q}{k}(x_{k}) \, \lhk{q}{k}(x_{k-1} \given x_{k}) \Big[ \log f_{k-1}(x_{k} \given x_{k-1}) - \log \lhk{q}{k}(x_{k-1} \given x_{k}) \Big] \dif x_{k-1} \dif x_{k} \\
        & - \int \lhk{q}{\T}(x_{\T}) \log \lhk{q}{\T}(x_{\T}) \dif x_{\T}
    \end{aligned} \\
    & \subject \qquad & & \lhk{q}{k-1}(x_{k-1}) = \int \lhk{q}{k}(x_{k}) \, \lhk{q}{k}(x_{k-1} \given x_{k}) \dif x_{k}, \quad \forall \, x_{k-1}, \, 0 < k \leq T, \\
    & & & 1 = \int \lhk{q}{\T}(x_{\T}) \dif x_{\T}, \quad 1 = \int \lhk{q}{k}(x_{k-1} \given x_{k}) \dif x_{k-1}, \quad \forall x_{k}, \,\, 0 < k \leq T, \\
    & & & \varepsilon \geq
    \begin{aligned}[t]
        & \int \lhk{q}{\T}(x_{\T}) \log \frac{\lhk{q}{\T}(x_{\T})}{\lhki{q}{\T}{i}(x_{\T})} \dif x_{\T} \\
        & + \sum_{k=1}^{T} \int \lhk{q}{k}(x_{k}) \int \lhk{q}{k}(x_{k-1} \given x_{k}) \log \frac{\lhk{q}{k}(x_{k-1} \given x_{k})}{\lhki{q}{k}{i}(x_{k-1} \given x_{k})} \dif x_{k-1} \dif x_{k},
    \end{aligned}
\end{align}
and the corresponding Lagrangian functional $\lh{\mathcal{R}}(\lhk{q}{\T}, \lhk{q}{k}, \lhk{\gamma}{\T}, \lhk{\lambda}{k}, \lhkxx{V}{k}, \alpha)$ is
\begin{equation}\label{app-eq:reverse-markov-lagrangian}
    \lh{\mathcal{R}}(\cdot) = 
    \begin{aligned}[t]
        & \int \lhk{q}{0}(x_{0}) \log p(x_{0}) \dif x_{0} + \sum_{k=1}^{T} \int \lhk{q}{k}(x_{k}) \log h_{k}(y_{k} \given x_{k}) \dif x_{k} - \int \lhk{q}{\T}(x_{\T}) \log \lhk{q}{\T}(x_{\T}) \dif x_{\T} \\
        & + \sum_{k=1}^{T} \int \lhk{q}{k}(x_{k}) \, \lhk{q}{k}(x_{k-1} \given x_{k}) \Big[ \log f_{k-1}(x_{k} \given x_{k-1}) - \log \lhk{q}{k}(x_{k-1} \given x_{k}) \Big] \dif x_{k-1} \dif x_{k} \\
        & + \sum_{k=1}^{T} \int \lhkxx{V}{k-1}(x_{k-1}) \left[ \int \lhk{q}{k}(x_{k}) \, \lhk{q}{k}(x_{k-1} \given x_{k}) \dif x_{k} - \lhk{q}{k-1}(x_{k-1}) \right] \dif x_{k-1} \\
        & + \sum_{k=1}^{T} \int \lhk{\lambda}{k}(x_{k}) \left[ 1 - \int \lhk{q}{k}(x_{k-1} \given x_{k}) \dif x_{k-1} \right] \dif x_{k} + \lhk{\gamma}{\T} \left[ 1 - \int \lhk{q}{\T}(x_{\T}) \dif x_{\T} \right] \\
        & + \alpha 
        \begin{aligned}[t]
            \Bigg[ & \varepsilon - \int \lhk{q}{\T}(x_{\T}) \log \frac{\lhk{q}{\T}(x_{\T})}{\lhki{q}{\T}{i}(x_{\T})} \dif x_{\T} \\ 
            & - \sum_{k=1}^{T} \int \lhk{q}{k}(x_{k}) \int \lhk{q}{k}(x_{k-1} \given x_{k}) \log \frac{\lhk{q}{k}(x_{k-1} \given x_{k})}{\lhki{q}{k}{i}(x_{k-1} \given x_{k})} \dif x_{k-1} \dif x_{k} \Bigg],
        \end{aligned}
    \end{aligned}
\end{equation}
where $\lhk{V}{k}(x_{k})$, $\lhk{\lambda}{k}(x_{k})$, $\lhk{\gamma}{\T}$, and $\alpha \geq 0$ are Lagrangian multipliers. To find the solution with respect to $\lhk{q}{k}(x_{k-1} \given x_{k})$, we zero the associated derivatives of $\lh{\mathcal{R}}(\cdot)$
\begin{align}\label{app-eq:reverse-markov-optimal-conditionals}
    \lhki{q}{k}{i+1}(x_{k-1} \given x_{k}) & = 
    \begin{aligned}[t]
        & \Big[ \exp \left\{ \lhk{\lambda}{k}(x_{k}) / \lhk{q}{k}(x_{k}) + (1 + \alpha) \right\} \Big]^{-1 / (1 + \alpha)} \\ 
        & \times \Big[ \lhki{q}{k}{i}(x_{k-1} \given x_{k}) \Big]^{\alpha / (1 + \alpha)} \Big[ f_{k-1}(x_{k} \given x_{k-1}) \exp \left\{ \lhkxx{V}{k-1}(x_{k-1}) \right\} \Big]^{1/(1 + \alpha)}
    \end{aligned}
\end{align}
Moreover, by setting the functional derivative with respect to $\lhk{q}{\T}(x_{\T})$ to zero, we get
\begin{equation}\label{app-eq:reverse-markov-optimal-marginal}
    \lhki{q}{\T}{i+1}(x_{\T}) = \Big[ \exp \Big\{ \lhk{\gamma}{\T} + (1 + \alpha) \Big\} \Big]^{- 1 / (1 + \alpha)} \Big[ \lhki{q}{\T}{i}(x_{\T}) \Big]^{\alpha / (1 + \alpha)} \Big[ \exp \left\{ \lhkxx{V}{\T}(x_{\T}) \right\} \Big]^{1 / (1 + \alpha)}
\end{equation}
where, for convenience of notation, we define
\begin{equation}\label{app-eq:reverse-markov-last-potential}
    \begin{aligned}[t]
        \lhkxx{V}{\T}(x_{\T}) = \log h_{\T}(y_{\T} \given x_{\T}) + (1 + \alpha) & \log \int \left[ \lhki{q}{\T}{i}(x_{\T-1} \given x_{\T}) \right]^{\alpha / (1 + \alpha)} \\
        & \times \Big[ f_{\T-1}(x_{\T} \given x_{\T-1}) \exp \left\{ \lhkxx{V}{\T-1}(x_{\T-1}) \right\} \Big]^{1 / (1 + \alpha)} \dif x_{\T-1}.
    \end{aligned}
\end{equation}
Again, the solutions \eqref{app-eq:reverse-markov-optimal-conditionals} and \eqref{app-eq:reverse-markov-optimal-marginal} are not \emph{yet} normalized densities. We need to solve for the (functional) multipliers associated with the normalization constraints, $\lhk{\lambda}{k}(x_{k})$ and $\lhk{\gamma}{\T}$. Plugging the solution \eqref{app-eq:reverse-markov-optimal-conditionals} into the Lagrangian \eqref{app-eq:reverse-markov-lagrangian} results in the dual functional
\begin{equation}\label{app-eq:reverse-markov-dual}
    \lh{\mathcal{G}}(\cdot) = 
    \begin{aligned}[t]
        & \alpha \, \varepsilon + \int \lhk{q}{0}(x_{0}) \log p(x_{0}) \dif x_{0} + \sum_{k=1}^{T} \int \lhk{q}{k}(x_{k}) \log h_{k}(y_{k} \given x_{k}) \dif x_{k} \\
        & + \int \lhk{q}{\T}(x_{\T}) \left[ \alpha \log \lhki{q}{\T}{i}(x_{\T}) - (1 + \alpha) \log \lhk{q}{\T}(x_{\T}) \right] \dif x_{\T} + \lhk{\gamma}{\T} \left[ 1 - \int \lhk{q}{\T}(x_{\T}) \dif x_{\T} \right] \\
        & + \sum_{k=1}^{T} \int \lhk{\lambda}{k}(x_{k}) \dif x_{k} - \sum_{k=1}^{T} \int \lhkxx{V}{k-1}(x_{k-1}) \lhk{q}{k-1}(x_{k-1}) \dif x_{k-1} \\
        & + (1 + \alpha) \sum_{k=1}^{T} \int \lhk{q}{k}(x_{k}) \int \lhki{q}{k}{i+1}(x_{k-1} \given x_{k}) \dif x_{k-1} \dif x_{k},
    \end{aligned}
\end{equation}
which we use to solve for the multipliers $\lhk{\lambda}{k}(x_{k})$ by zeroing the associated derivatives
\begin{align}\label{app-eq:reverse-markov-conditionals-normalizer}
    \lhki{\lambda}{k}{i+1}(x_{k}) & = 
    \begin{aligned}[t]
        (1 + \alpha) \, \lhk{q}{k}(x_{k}) \Big[ -1 & + \log \int \Big[ \lhki{q}{k}{i}(x_{k-1} \given x_{k}) \Big]^{\alpha / (1 + \alpha)} \\ 
        & \times \Big[ f_{k-1}(x_{k} \given x_{k-1}) \exp \left\{ \lhkxx{V}{k-1}(x_{k-1}) \right\} \Big]^{1 / (1 + \alpha)} \dif x_{k-1} \Big]
    \end{aligned} \\
    & = (1 + \alpha) \, \lhk{q}{k}(x_{k}) \Big[ -1 + \log \lhki{\psi}{k}{i+1}(x_{k}) \Big].
\end{align}
We retrieve the normalized tilted conditionals by plugging \eqref{app-eq:reverse-markov-conditionals-normalizer} into \eqref{app-eq:reverse-markov-optimal-conditionals}
\begin{equation}\label{app-eq:reverse-markov-normalized-conditionals}
    \lhki{q}{k}{i+1}(x_{k-1} \given x_{k}) =
    \begin{aligned}[t]
        \Big[ \lhki{\psi}{k}{i+1}(x_{k}) \Big]^{-1} & \Big[ \lhki{q}{k}{i}(x_{k-1} \given x_{k}) \Big]^{\beta} \\
        & \times \Big[ f_{k-1}(x_{k} \given x_{k-1}) \exp \left\{ \lhkxx{V}{k-1}(x_{k-1}) \right\} \Big]^{1 - \beta},
    \end{aligned}
\end{equation}
where we have defined $\beta \coloneqq \alpha / (1 + \alpha)$, so that $\beta = 0$ when $\alpha = 0$, and $\beta \rightarrow 1$ as $\alpha \rightarrow \infty$. Given the results \eqref{app-eq:reverse-markov-conditionals-normalizer} and \eqref{app-eq:reverse-markov-normalized-conditionals}, we simplify the functional $\lh{\mathcal{G}}(\cdot)$ further
\begin{equation}
    \lh{\mathcal{G}}(\cdot) = 
    \begin{aligned}[t]
        & \alpha \, \varepsilon + \int \lhk{q}{0}(x_{0}) \log p(x_{0}) \dif x_{0} + \sum_{k=1}^{T} \int \lhk{q}{k}(x_{k}) \log h_{k}(y_{k} \given x_{k}) \dif x_{k} \\
        & - \sum_{k=0}^{T} \int \lhkxx{V}{k}(x_{k}) \lhk{q}{k}(x_{k}) \dif x_{k} + (1 + \alpha) \sum_{k=1}^{T} \int \lhk{q}{k}(x_{k}) \log \lhki{\psi}{k}{i+1}(x_{k}) \dif x_{k} \\
        & + \lhk{\gamma}{\T} + (1 + \alpha) \int \lhki{q}{\T}{i+1}(x_{\T}) \dif x_{\T}.
    \end{aligned}
\end{equation}
Next, we solve for the optimal multiplier $\lhk{\gamma}{\T}$ by zeroing the associated derivative
\begin{align}\label{app-eq:reverse-markov-marginal-normalizer}
    \lhki{\gamma}{\T}{i+1} & = (1 + \alpha) \left[ -1 + \log \int \Big[ \lhki{q}{\T}{i}(x_{\T}) \Big]^{\alpha / (1 + \alpha)} \Big[ \exp \left\{ \lhkxx{V}{\T}(x_{\T}) \right\} \Big]^{1 / (1 + \alpha)} \dif x_{\T} \right] \\
    & = (1 + \alpha) \left[ -1 + \log \lhki{\mathcal{Z}}{\T}{i+1} \right],
\end{align}
leading to the normalization of the tilted distribution \eqref{app-eq:reverse-markov-optimal-marginal}
\begin{equation}\label{app-eq:reverse-markov-normalized-marginal}
    \lhki{q}{\T}{i+1}(x_{\T}) = \Big[ \lhki{\mathcal{Z}}{\T}{i+1} \Big]^{-1} \Big[ \lhki{q}{\T}{i}(x_{\T}) \Big]^{\beta} \Big[ \exp \left\{ \lhkxx{V}{\T}(x_{\T}) \right\} \Big]^{1 - \beta}.
\end{equation}
Using \eqref{app-eq:reverse-markov-normalized-marginal} and \eqref{app-eq:reverse-markov-marginal-normalizer} leads to further simplification of the dual $\lh{\mathcal{G}}(\cdot)$ 
\begin{equation}
    \lh{\mathcal{G}}(\cdot) = 
    \begin{aligned}[t]
        & \alpha \, \varepsilon + \int \lhk{q}{0}(x_{0}) \log p(x_{0}) \dif x_{0} + \sum_{k=1}^{T} \int \lhk{q}{k}(x_{k}) \log h_{k}(y_{k} \given x_{k}) \dif x_{k} \\
        & \hspace{-1.0cm} - \sum_{k=0}^{T} \int \lhkxx{V}{k}(x_{k}) \lhk{q}{k}(x_{k}) \dif x_{k} + (1 + \alpha) \sum_{k=1}^{T} \int \lhk{q}{k}(x_{k}) \log \lhki{\psi}{k}{i+1}(x_{k}) \dif x_{k} + (1 + \alpha) \log \lhki{\mathcal{Z}}{\T}{i+1}.
    \end{aligned}
\end{equation}
Finally, by taking \eqref{app-eq:reverse-markov-last-potential} into consideration and zeroing the derivative of $\lh{\mathcal{G}}(\cdot)$ with respect to $\lhk{q}{k}(x_{k}), \forall \, 0 \leq k < T$, we find the optimal potential functions
\begin{equation}\label{app-eq:reverse-markov-optimal-potentials}
    \lhkixx{V}{k}{i+1}(x_{k}) =
    \begin{cases}
        \, \log p_{0}(x_{0}) & \mathrm{if} \,\, k = 0, \\[0.5em]
        \, \log h_{k}(y_{k} \given x_{k}) + 1 / (1 - \beta) \log \lhki{\psi}{k}{i+1}(x_{k}) & \mathrm{if} \,\, 0 < k \leq T.
    \end{cases}
\end{equation}
If we plug this result back into the dual $\lh{\mathcal{G}}(\cdot)$, we retrieve the simplest form of the dual
\begin{equation}
    \lh{\mathcal{G}}(\beta) = \frac{\beta \varepsilon}{1 - \beta} + \frac{1}{1 - \beta} \log \lhki{\mathcal{Z}}{\T}{i+1}(\beta).
\end{equation}

\section{Proof of Proposition~\ref{prop:forward-gauss-markov-potentials}}
\label{app:forward-gauss-markov-potentials-proof}
At $k=T$, Definition~\ref{def:statistical-expansion-log-densities} readily delivers a quadratic form for the potential function
\begin{align}
    \rhkixx{V}{\T}{i+1}(x_{\T}) = \log h_{\T}(y_{\T} \given x_{\T}) & = - \frac{1}{2} x_{\T}^{\top} \, \rhki{R}{\T}{i+1} \, x_{\T} + x_{\T}^{\top} \, \rhki{r}{\T}{i+1} + \rhki{\rho}{\T}{i+1} \\
    & = - \frac{1}{2} x_{\T}^{\top} L_{\T}^{[i]} \, x_{\T} + x_{\T}^{\top} \, l_{\T}^{[i]} + \nu_{\T}^{\,[i]}.\\[-2.25em]
\end{align}
For all $0 \leq k < T$, the potential $\rhkixx{V}{k}{i+1}(x_k)$ satisfies the backward recursion
\begin{align}
    \rhkixx{V}{k}{i+1}(x_{k}) & = \log h_{k}(y_{k} \given x_{k}) + 1 / (1 - \beta) \log \rhki{\psi}{k}{i+1}(x_{k}) \\
    & = 
    \begin{aligned}[t]
        \log h_{k}(y_{k} \given x_{k}) + 1 / (1 - \beta) \log & \int \Big[ \rhki{q}{k}{i}(x_{k+1} \given x_{k}) \Big]^{\beta} \\
        & \times \Big[ f_{k}(x_{k+1} \given x_{k}) \exp \left\{ \rhkixx{V}{k+1}{i+1}(x_{k+1}) \right\} \Big]^{1 - \beta} \dif x_{k+1},
    \end{aligned}
\end{align}
with $\log p_0(x_0)$ replacing $\log h_k(y_k \given x_k)$ at $k=0$. We show that this recursion is a tractable reverse propagation of quadratic forms. We start by examining $\rhki{\psi}{k}{i+1}(x_{k})$, the integral over $x_{k+1}$. Let us drag all terms into the exponential, then we can treat the exponent as a quadratic function over $x_{k}$ and $x_{k+1}$
\begin{align}
    & - \frac{1}{2} 
    \begin{bmatrix} x_{k+1}^{\top} & x_{k}^{\top} \end{bmatrix}
    \begin{bmatrix} 
        \rhki{G}{\bar x \bar x, k}{i+1} & - \rhki{G}{\bar x x, k}{i+1} \\[0.5em]
        - \rhki{G}{x \bar x, k}{i+1} & \rhki{G}{xx, k}{i+1}
    \end{bmatrix}
    \begin{bmatrix} x_{k+1} \\[0.75em] x_{k} \end{bmatrix}
    + \begin{bmatrix} x_{k+1}^{\top} & x_{k}^{\top} \end{bmatrix}
    \begin{bmatrix} \rhki{g}{\bar x, k}{i+1} \\[0.5em] \rhki{g}{x, k}{i+1} \end{bmatrix}
    + \rhki{\theta}{k}{i+1} \\
    & = \beta \log \rhki{q}{k}{i}(x_{k+1} \given x_{k}) + (1 - \beta) \log f_{k}(x_{k+1} \given x_{k}) + (1 - \beta) \, \rhkixx{V}{k+1}{i+1}(x_{k+1}).
\end{align}
Now, given an affine-Gaussian $\rhki{q}{k}{i}(x_{k+1} \given x_{k})$ (Assumption~\ref{asm:forward-gauss-markov}), a quadratic $\log f_{k}(x_{k+1} \given x_{k})$ (Definition~\ref{def:statistical-expansion-log-densities}), and a quadratic potential function $\rhkixx{V}{k+1}{i+1}(x_{k+1})$ of the form~\eqref{eq:potentials-quadratic-form}, we can match the quadratic factors between the left- and right-hand sides, leading to
\begin{align}
    \rhki{G}{\bar x \bar x, k}{i+1} & = (1 - \beta) \left[ C_{\bar x \bar x, k}^{[i]} + \rhki{R}{k+1}{i+1} \right] + \beta \left[ \rhki{\Sigma}{k}{i} \right]^{-1}, \\
    \rhki{G}{xx, k}{i+1} & = (1 - \beta) \, C_{xx, k}^{[i]} + \beta \left[ \rhkix{F}{k}{i} \right]^{\top} \left[ \rhki{\Sigma}{k}{i} \right]^{-1} \rhkix{F}{k}{i}, \\
    \rhki{G}{\bar x x, k}{i+1} & = (1 - \beta) \, C_{\bar x x, k}^{[i]} + \beta \left[ \rhki{\Sigma}{k}{i} \right]^{-1} \rhkix{F}{k}{i}, \\
    \rhki{G}{x \bar x, k}{i+1} & = (1 - \beta) \, C_{x \bar x, k}^{[i]} + \beta \left[ \rhkix{F}{k}{i} \right]^{\top} \left[ \rhki{\Sigma}{k}{i} \right]^{-1}, \\
    \rhki{g}{\bar x, k}{i+1} & = (1 - \beta) \left[ c_{\bar x, k}^{[i]} + \rhki{r}{k+1}{i+1} \right] + \beta \left[ \rhki{\Sigma}{k}{i} \right]^{-1} \rhki{d}{k}{i}, \\
    \rhki{g}{x, k}{i+1} & = (1 - \beta) \, c_{x, k}^{[i]} - \beta \left[ \rhkix{F}{k}{i} \right]^{\top} \left[ \rhki{\Sigma}{k}{i} \right]^{-1} \rhki{d}{k}{i}, \\
    \rhki{\theta}{k}{i+1} & = (1 - \beta) \, \left[ \kappa_{k}^{[i]} + \rhki{\rho}{k+1}{i+1} \right] - \frac{\beta}{2} \log \Big| 2 \pi \rhki{\Sigma}{k}{i} \Big| - \frac{\beta}{2} \left[ \rhki{d}{k}{i} \right]^{\top} \left[ \rhki{\Sigma}{k}{i} \right]^{-1} \rhki{d}{k}{i}.
\end{align}
Next, we reformulate the quadratic function in the exponent explicitly as a function of $x_{k+1}$
\begin{equation}
    - \frac{1}{2} x_{k+1}^{\top} \, \rhki{G}{\bar x \bar x, k}{i+1} \, x_{k+1} + x_{k+1}^{\top} \left[ \rhki{G}{\bar x x, k}{i+1} \, x_{k} + \rhki{g}{\bar x, k}{i+1} \right] + \left[ - \frac{1}{2} x_{k}^{\top} \, \rhki{G}{xx, k}{i+1} \, x_{k} + x_{k}^{\top} \, \rhki{g}{x, k}{i+1} + \rhki{\theta}{k}{i+1} \right].
\end{equation}
Consequently, the exponential now resembles an unnormalized Gaussian distribution in \emph{information form} with a precision matrix $\rhki{G}{\bar x \bar x, k}{i+1}$. In this case, we can use the identity
\begin{equation}\label{app-eq:log-sum-exp-identity}
    \log \int \exp \left\{ - \frac{1}{2} x^{\top} U x + x^{\top} u + \sigma \right\} \dif x = \frac{1}{2} \log |2 \pi \, U^{-1}| + \frac{1}{2} u^{\top} U^{-1} u + \sigma,
\end{equation}
to express the log-normalizing function $\log \rhki{\psi}{k}{i+1}(x_{k})$ as a quadratic function
\begin{align}
    \log \rhki{\psi}{k}{i+1}(x_{k}) & = - \frac{1}{2} x_{k}^{\top} \, \rhkix{S}{k}{i+1} \, x_{k} + x_{k}^{\top} \rhki{s}{k}{i+1} + \rhki{\xi}{k}{i+1} \\
    & = 
    \begin{aligned}[t]
        & \frac{1}{2} \log \Big| 2 \pi \, \left[ \rhki{G}{\bar x \bar x, k}{i+1} \right]^{-1} \Big| + \left[ - \frac{1}{2} x_{k}^{\top} \, \rhki{G}{xx, k}{i+1} \, x_{k} + x_{k}^{\top} \, \rhki{g}{x, k}{i+1} + \rhki{\theta}{k}{i+1} \right] \\
        & + \frac{1}{2} \left[ \rhki{G}{\bar x x, k}{i+1} \, x_{k} + \rhki{g}{\bar x, k}{i+1} \right]^{\top} \left[ \rhki{G}{\bar x \bar x, k}{i+1} \right]^{-1} \left[ \rhki{G}{\bar x x, k}{i+1} \, x_{k} + \rhki{g}{\bar x, k}{i+1} \right] 
    \end{aligned}
\end{align}
which, after matching terms, leads to
\begin{align}
    \rhkix{S}{k}{i+1} & = \rhki{G}{xx, k}{i+1} - \left[ \rhki{G}{\bar x x, k}{i+1} \right]^{\top} \left[ \rhki{G}{\bar x \bar x, k}{i+1} \right]^{-1} \rhki{G}{\bar x x, k}{i+1}, \\
    \rhki{s}{k}{i+1} & = \rhki{g}{x, k}{i+1} + \left[ \rhki{G}{\bar x x, k}{i+1} \right]^{\top} \left[ \rhki{G}{\bar x \bar x, k}{i+1} \right]^{-1} \rhki{g}{\bar x, k}{i+1}, \\
    \rhki{\xi}{k}{i+1} & = \rhki{\theta}{k}{i+1} + \frac{1}{2} \log \Big| 2 \pi \, \left[ \rhki{G}{\bar x \bar x, k}{i+1} \right]^{-1} \Big| + \frac{1}{2} \, \left[ \rhki{g}{\bar x, k}{i+1} \right]^{\top} \, \left[ \rhki{G}{\bar x \bar x, k}{i+1} \right]^{-1} \rhki{g}{\bar x, k}{i+1}.
\end{align}
To get the final form of the \emph{quadratic} potential $\rhkixx{V}{k}{i+1}(x_{k})$, we add the contribution of the quadratic log-measurement and log-prior determined by Definition~\ref{def:statistical-expansion-log-densities}
\begin{equation}
    \rhki{R}{k}{i+1} = L_{k}^{[i]} + \frac{1}{1 - \beta} \, \rhkix{S}{k}{i+1}, \quad 
    \rhki{r}{k}{i+1} = l_{k}^{[i]} + \frac{1}{1 - \beta} \, \rhki{s}{k}{i+1}, \quad
    \rhki{\rho}{k}{i+1} = \nu_{k}^{[i]} + \frac{1}{1 - \beta} \, \rhki{\xi}{k}{i+1}.
\end{equation}
Finally, for a Gaussian $\rhki{q}{0}{i}(x_{0})$ and a quadratic $\rhkixx{V}{0}{i+1}(x_{0})$, we derive a quadratic $\log \rhki{\mathcal{Z}}{0}{i+1}$
\begin{equation}
    \log \rhki{\mathcal{Z}}{0}{i+1} = \log \int \Big[ \rhki{q}{0}{i}(x_{0}) \Big]^{\beta} \Big[  \exp \left\{ \rhkixx{V}{0}{i+1}(x_{0}) \right\} \Big]^{1 - \beta} \dif x_{0}.
\end{equation}
Again, we formulate a quadratic function over $x_{0}$ and $\rhki{m}{0}{i}$
\begin{align}
    & - \frac{1}{2} 
    \begin{bmatrix} x_{0}^{\top} & \left[ \rhki{m}{0}{i} \right]^{\top} \end{bmatrix}
    \begin{bmatrix} 
        \rhki{J}{xx}{i+1} & - \rhki{J}{xm}{i+1} \\[0.5em]
        - \rhki{J}{mx}{i+1} & \rhki{J}{mm}{i+1}
    \end{bmatrix}
    \begin{bmatrix} x_{0} \\[0.75em] \rhki{m}{0}{i} \end{bmatrix}
    + \begin{bmatrix} x_{0}^{\top} & \left[ \rhki{m}{0}{i} \right]^{\top} \end{bmatrix}
    \begin{bmatrix} \rhki{j}{x}{i+1} \\[0.5em] \rhki{j}{m}{i+1} \end{bmatrix}
    + \rhix{\tau}{i+1} \\
    & = \beta \log \rhki{q}{0}{i}(x_{0}) + (1 - \beta) \, \rhkixx{V}{0}{i+1}(x_{0}),
\end{align}
where by matching the quadratic factors, we get
\begin{align}
    \rhki{J}{xx}{i+1} & = (1 - \beta) \, \rhki{R}{0}{i+1} + \beta \left[ \rhkix{P}{0}{i} \right]^{-1}, & & \rhki{J}{xm}{i+1} = \beta \left[ \rhkix{P}{0}{i} \right]^{-1}, \\
    \rhki{J}{mm}{i+1} & = \beta \left[ \rhkix{P}{0}{i} \right]^{-1}, & & \rhki{J}{mx}{i+1} = \beta \left[ \rhkix{P}{0}{i} \right]^{-1}, \\
    \rhki{j}{x}{i+1} & = (1 - \beta) \, \rhki{r}{0}{i+1}, & & \rhki{j}{m}{i+1} = 0, \\
    \rhix{\tau}{i+1} & = (1 - \beta) \, \rhki{\rho}{0}{i+1} - \frac{\beta}{2} \log \Big| 2 \pi \, \rhkix{P}{0}{i} \Big|.
\end{align}
Using the identity~\eqref{app-eq:log-sum-exp-identity}, we get a log-normalizer as a quadratic function over $\rhki{m}{0}{i}$
\begin{equation}
    \log \rhki{\mathcal{Z}}{0}{i+1} = - \frac{1}{2} \left[ \rhki{m}{0}{i} \right]^{\top} \rhix{U}{i+1} \, \rhki{m}{0}{i} + \left[ \rhki{m}{0}{i} \right]^{\top} \rhix{u}{i+1} + \rhix{\eta}{i+1},
\end{equation}
where
\begin{equation}
    \begin{aligned}[t]
        \rhix{U}{i+1} & = \rhki{J}{mm}{i+1} - \left[ \rhki{J}{xm}{i+1} \right]^{\top} \left[ \rhki{J}{xx}{i+1} \right]^{-1} \rhki{J}{xm}{i+1}, \\
        \rhix{u}{i+1} & = \rhki{j}{m}{i+1} + \left[ \rhki{J}{xm}{i+1} \right]^{\top} \left[ \rhki{J}{xx}{i+1} \right]^{-1} \rhki{j}{x}{i+1}, \\
        \rhix{\eta}{i+1} & = \rhix{\tau}{i+1} + \frac{1}{2} \log \Big| 2 \pi \, \left[ \rhki{J}{xx}{i+1} \right]^{-1} \Big| + \frac{1}{2} \Big[ \rhki{j}{x}{i+1} \Big]^{\top} \Big[ \rhki{J}{xx}{i+1} \Big]^{-1} \Big[ \rhki{j}{x}{i+1} \Big].
    \end{aligned}
\end{equation}
Plugging in the corresponding terms leads to the final result.

\section{Proof of Proposition~\ref{prop:reverse-gauss-markov-potentials}}
\label{app:reverse-gauss-markov-potentials-proof}
For $k=0$, the log-prior is a quadratic function per Definition~\ref{def:statistical-expansion-log-densities}
\begin{align}
    \lhkixx{V}{0}{i+1}(x_{0}) = \log p_{0}(x_{0}) & = - \frac{1}{2} x_{0}^{\top} \, \lhki{R}{0}{i+1} \, x_{0} + x_{0}^{\top} \, \lhki{r}{0}{i+1} + \lhki{\rho}{0}{i+1} \\
    & = - \frac{1}{2} x_{0}^{\top} L_{0}^{[i]} \, x_{0} + x_{0}^{\top} \, l_{0}^{[i]} + \nu_{0}^{\,[i]}.
\end{align}
For $\lhkixx{V}{k}{i+1}(x_{k})$, for all $0 < k \leq T$, we have derived a forward recursion
\begin{align}
    \lhkixx{V}{k}{i+1}(x_{k}) & = \log h_{k}(y_{k} \given x_{k}) + 1 / (1 - \beta) \log \lhki{\psi}{k}{i+1}(x_{k}) \\
    & = 
    \begin{aligned}[t]
        \log h_{k}(y_{k} \given x_{k}) + 1 / (1 - \beta) \log & \int \Big[ \lhki{q}{k}{i}(x_{k-1} \given x_{k}) \Big]^{\beta} \\
        & \times \Big[ f_{k-1}(x_{k} \given x_{k-1}) \exp \left\{ \lhkixx{V}{k-1}{i+1}(x_{k-1}) \right\} \Big]^{1 - \beta} \dif x_{k-1}.
    \end{aligned}
\end{align}
We show that this recursion is a tractable forward propagation of quadratic forms. Let's move all terms within the integral into the exponential and treat the exponent as a quadratic function over $x_{k}$ and $x_{k-1}$
\begin{align}
    & - \frac{1}{2} 
    \begin{bmatrix} x_{k}^{\top} & x_{k-1}^{\top} \end{bmatrix}
    \begin{bmatrix} 
        \lhki{G}{\bar x \bar x, k}{i+1} & - \lhki{G}{\bar x x, k}{i+1} \\[0.5em]
        - \lhki{G}{x \bar x, k}{i+1} & \lhki{G}{xx, k}{i+1}
    \end{bmatrix}
    \begin{bmatrix} x_{k} \\[0.75em] x_{k-1} \end{bmatrix}
    + \begin{bmatrix} x_{k}^{\top} & x_{k-1}^{\top} \end{bmatrix}
    \begin{bmatrix} \lhki{g}{\bar x, k}{i+1} \\[0.5em] \lhki{g}{x, k}{i+1} \end{bmatrix}
    + \lhki{\theta}{k}{i+1} \\
    & = \beta \log \lhki{q}{k}{i}(x_{k-1} \given x_{k}) + (1 - \beta) \log f_{k-1}(x_{k} \given x_{k-1}) + (1 - \beta) \, \lhkixx{V}{k-1}{i+1}(x_{k-1}).
\end{align}
For an affine-Gaussian $\lhki{q}{k}{i}(x_{k-1} \given x_{k})$ (Assumption~\ref{asm:reverse-gauss-markov}), a quadratic $\log f_{k-1}(x_{k} \given x_{k-1})$ (Definition~\ref{def:statistical-expansion-log-densities}), and a quadratic potential function $\lhkixx{V}{k-1}{i+1}(x_{k-1})$ of the form~\eqref{eq:potentials-quadratic-form}, we get
\begin{align}
    \lhki{G}{\bar x \bar x, k}{i+1} & = (1 - \beta) \, C_{\bar x \bar x, k-1}^{[i]} + \beta \left[ \lhkix{F}{k}{i} \right]^{\top} \left[ \lhki{\Sigma}{k}{i} \right]^{-1} \lhkix{F}{k}{i}, \\
    \lhki{G}{xx, k}{i+1} & = (1 - \beta) \, \left[ C_{xx, k-1}^{[i]} + \lhki{R}{k-1}{i+1} \right] + \beta \left[ \lhki{\Sigma}{k}{i} \right]^{-1}, \\
    \lhki{G}{\bar x x, k}{i+1} & = (1 - \beta) \, C_{\bar x x, k-1}^{[i]} + \beta \left[ \lhkix{F}{k}{i} \right]^{\top} \left[ \lhki{\Sigma}{k}{i} \right]^{-1}, \\
    \lhki{G}{x \bar x, k}{i+1} & = (1 - \beta) \, C_{x \bar x, k-1}^{[i]} + \beta \left[ \lhki{\Sigma}{k}{i} \right]^{-1} \lhkix{F}{k}{i}, \\
    \lhki{g}{\bar x, k}{i+1} & = (1 - \beta) \, c_{\bar x, k-1}^{[i]} - \beta \left[ \lhkix{F}{k}{i} \right]^{\top} \left[ \lhki{\Sigma}{k}{i} \right]^{-1} \lhki{d}{k}{i}, \\
    \lhki{g}{x, k}{i+1} & = (1 - \beta) \, \left[ c_{x, k-1}^{[i]} + \lhki{r}{k-1}{i+1} \right] + \beta \left[ \lhki{\Sigma}{k}{i} \right]^{-1} \lhki{d}{k}{i}, \\
    \lhki{\theta}{k}{i+1} & = (1 - \beta) \, \left[ \kappa_{k-1}^{[i]} + \lhki{\rho}{k-1}{i+1} \right] - \frac{\beta}{2} \log \Big| 2 \pi \lhki{\Sigma}{k}{i} \Big| - \frac{\beta}{2} \left[ \lhki{d}{k}{i} \right]^{\top} \left[ \lhki{\Sigma}{k}{i} \right]^{-1} \lhki{d}{k}{i}.
\end{align}
By writing this quadratic function explicitly in terms of $x_{k-1}$
\begin{equation}
    - \frac{1}{2} x_{k-1}^{\top} \, \lhki{G}{xx, k}{i+1} \, x_{k-1} + x_{k-1}^{\top} \left[ \lhki{G}{x \bar x, k}{i+1} \, x_{k} + \lhki{g}{x, k}{i+1} \right] + \left[ - \frac{1}{2} x_{k}^{\top} \, \lhki{G}{\bar x \bar x, k}{i+1} \, x_{k} + x_{k}^{\top} \, \lhki{g}{\bar x, k}{i+1} + \lhki{\theta}{k}{i+1} \right].
\end{equation}
we can make use of identity~\eqref{app-eq:log-sum-exp-identity} to express the log-normalizing function $\log \lhki{\psi}{k}{i+1}(x_{k})$ itself as a quadratic function in $x_{k}$
\begin{align}
    \log \lhki{\psi}{k}{i+1}(x_{k}) & = - \frac{1}{2} x_{k}^{\top} \, \lhkix{S}{k}{i+1} \, x_{k} + x_{k}^{\top} \lhki{s}{k}{i+1} + \lhki{\xi}{k}{i+1} \\
    & = 
    \begin{aligned}[t]
        & \frac{1}{2} \log \Big| 2 \pi \, \left[ \lhki{G}{xx, k}{i+1} \right]^{-1} \Big| + \left[ - \frac{1}{2} x_{k}^{\top} \, \lhki{G}{\bar x \bar x, k}{i+1} \, x_{k} + x_{k}^{\top} \, \lhki{g}{\bar x, k}{i+1} + \lhki{\theta}{k}{i+1} \right] \\
        & + \frac{1}{2} \left[ \lhki{G}{x \bar x, k}{i+1} \, x_{k} + \lhki{g}{x, k}{i+1} \right]^{\top} \left[ \lhki{G}{xx, k}{i+1} \right]^{-1} \left[ \lhki{G}{x \bar x, k}{i+1} \, x_{k} + \lhki{g}{x, k}{i+1} \right]
    \end{aligned}
\end{align}
which, after matching terms, leads to
\begin{align}
    \lhkix{S}{k}{i+1} & = \lhki{G}{\bar x \bar x, k}{i+1} - \left[ \lhki{G}{x \bar x, k}{i+1} \right]^{\top} \left[ \lhki{G}{xx, k}{i+1} \right]^{-1} \lhki{G}{x \bar x, k}{i+1}, \\
    \lhki{s}{k}{i+1} & = \lhki{g}{\bar x, k}{i+1} + \left[ \lhki{G}{x \bar x, k}{i+1} \right]^{\top} \left[ \lhki{G}{xx, k}{i+1} \right]^{-1} \lhki{g}{x, k}{i+1}, \\
    \lhki{\xi}{k}{i+1} & = \lhki{\theta}{k}{i+1} + \frac{1}{2} \log \Big| 2 \pi \, \left[ \lhki{G}{xx, k}{i+1} \right]^{-1} \Big| + \frac{1}{2} \, \left[ \lhki{g}{x, k}{i+1} \right]^{\top} \, \left[ \lhki{G}{xx, k}{i+1} \right]^{-1} \lhki{g}{x, k}{i+1}.
\end{align}
Given $\log \lhki{\psi}{k}{i+1}(x_{k})$, we can now construct the \emph{quadratic} potential function $\lhkixx{V}{k}{i+1}(x_{k})$ that accounts for the log-measurement contribution according to Definition~\ref{def:statistical-expansion-log-densities}
\begin{equation}
    \lhki{R}{k}{i+1} = L_{k}^{[i]} + \frac{1}{1 - \beta} \, \lhkix{S}{k}{i+1}, \quad 
    \lhki{r}{k}{i+1} = l_{k}^{[i]} + \frac{1}{1 - \beta} \, \lhki{s}{k}{i+1}, \quad
    \lhki{\rho}{k}{i+1} = \nu_{k}^{[i]} + \frac{1}{1 - \beta} \, \lhki{\xi}{k}{i+1}.
\end{equation}
Finally, for a Gaussian $\lhki{q}{\T}{i}(x_{\T})$ and a quadratic $\lhkixx{V}{\T}{i+1}(x_{\T})$, we derive a quadratic $\log \lhki{\mathcal{Z}}{\T}{i+1}$
\begin{equation}
    \log \lhki{\mathcal{Z}}{\T}{i+1} = \log \int \Big[ \lhki{q}{\T}{i}(x_{\T}) \Big]^{\beta} \Big[ \exp \left\{ \lhkixx{V}{\T}{i+1}(x_{\T}) \right\} \Big]^{1 - \beta} \dif x_{\T}.
\end{equation}
Similar to the proof in Appendix~\ref{app:forward-gauss-markov-potentials-proof}, we formulate a quadratic over $x_{\T}$ and $\lhki{m}{\T}{i}$
\begin{align}
    & - \frac{1}{2} 
    \begin{bmatrix} \left[ \lhki{m}{\T}{i} \right]^{\top} & x_{\T}^{\top} \end{bmatrix}
    \begin{bmatrix} 
        \lhki{J}{xx}{i+1} & - \lhki{J}{xm}{i+1} \\[0.5em]
        - \lhki{J}{mx}{i+1} & \lhki{J}{mm}{i+1}
    \end{bmatrix}
    \begin{bmatrix} \lhki{m}{\T}{i} \\[0.75em] x_{\T} \end{bmatrix}
    + \begin{bmatrix} \left[ \lhki{m}{\T}{i} \right]^{\top} & x_{\T}^{\top} \end{bmatrix}
    \begin{bmatrix} \lhki{j}{x}{i+1} \\[0.5em] \lhki{j}{m}{i+1} \end{bmatrix}
    + \lhix{\tau}{i+1} \\
    & = \beta \log \lhki{q}{\T}{i}(x_{\T}) + (1 - \beta) \, \lhkixx{V}{\T}{i+1}(x_{\T}),
\end{align}
where by matching the quadratic factors, we get
\begin{align}
    \lhki{J}{xx}{i+1} & = \beta \left[ \lhkix{P}{\T}{i} \right]^{-1}, & & \lhki{J}{xm}{i+1} = \beta \left[ \lhkix{P}{\T}{i} \right]^{-1}, \\
    \lhki{J}{mm}{i+1} & = (1 - \beta) \, \lhki{R}{\T}{i+1} + \beta \left[ \lhkix{P}{\T}{i} \right]^{-1}, & & \lhki{J}{mx}{i+1} = \beta \left[ \lhkix{P}{\T}{i} \right]^{-1}, \\
    \lhki{j}{x}{i+1} & = 0, & & \lhki{j}{m}{i+1} = (1 - \beta) \, \lhki{r}{\T}{i+1}, \\
    \lhix{\tau}{i+1} & = (1 - \beta) \, \lhki{\rho}{\T}{i+1} - \frac{\beta}{2} \log \Big| 2 \pi \, \lhkix{P}{\T}{i} \Big|.
\end{align}
Using the identity~\eqref{app-eq:log-sum-exp-identity}, we get a log-normalizing constant as a quadratic function over $\lhki{m}{\T}{i}$
\begin{equation}
    \log \lhki{\mathcal{Z}}{\T}{i+1} = - \frac{1}{2} \left[ \lhki{m}{\T}{i} \right]^{\top} \lhix{U}{i+1} \, \lhki{m}{\T}{i} + \left[ \lhki{m}{\T}{i} \right]^{\top} \lhix{u}{i+1} + \lhix{\eta}{i+1},
\end{equation}
where
\begin{equation}
    \begin{aligned}[t]
        \lhix{U}{i+1} & = \lhki{J}{xx}{i+1} - \left[ \lhki{J}{mx}{i+1} \right]^{\top} \left[ \lhki{J}{mm}{i+1} \right]^{-1} \lhki{J}{mx}{i+1}, \\
        \lhix{u}{i+1} & = \lhki{j}{x}{i+1} + \left[ \lhki{J}{mx}{i+1} \right]^{\top} \left[ \lhki{J}{mm}{i+1} \right]^{-1} \lhki{j}{m}{i+1}, \\
        \lhix{\eta}{i+1} & = \lhix{\tau}{i+1} + \frac{1}{2} \log \Big| 2 \pi \, \left[ \lhki{J}{mm}{i+1} \right]^{-1} \Big| + \frac{1}{2} \Big[ \lhki{j}{m}{i+1} \Big]^{\top} \Big[ \lhki{J}{mm}{i+1} \Big]^{-1} \Big[ \lhki{j}{m}{i+1} \Big].
    \end{aligned}
\end{equation}
Plugging in the corresponding terms leads to the final result.

\section{Proof of Lemma~\ref{lem:gauss-markov-marginal-update}}
\label{app:gauss-markov-marginal-update-proof}
For a tilted distribution of the form
\begin{equation}
    q^{[i+1]}(x) = \left[ \mathcal{Z}^{[i+1]} \right]^{-1} \Big[ q^{[i]}(x) \Big]^{\beta} \Big[ \exp \left\{ V^{[i+1]}(x) \right\} \Big]^{1 - \beta}, 
\end{equation}
with a normalizing constant
\begin{equation}
    \mathcal{Z}^{[i+1]} = \int \Big[ q^{[i]}(x) \Big]^{\beta} \Big[ \exp \left\{ V^{[i+1]}(x) \right\} \Big]^{1 - \beta} \dif x,
\end{equation}
where $q^{[i]}(x) = \mathcal{N}(x \given m^{[i]}, P^{[i]})$, we compute the moments of $q^{[i+1]}(x)$ by considering the derivatives of $\mathcal{Z}^{[i+1]}$ with respect to $m^{[i]}$. Starting with the first-order derivative
\begin{align}
    \frac{\partial \mathcal{Z}^{[i+1]}}{\partial \, m^{[i]}} & = \beta \, \left[ P^{[i]} \right]^{-1} \int (x - m^{[i]}) \Big[ q^{[i]}(x) \Big]^{\beta} \Big[ \exp \left\{ V^{[i+1]}(x) \right\} \Big]^{1 - \beta} \dif x \\
    & = 
    \begin{aligned}[t]
        & \beta \, \left[ P^{[i]} \right]^{-1} \int x \Big[ q^{[i]}(x) \Big]^{\beta} \Big[ \exp \left\{ V^{[i+1]}(x) \right\} \Big]^{1 - \beta} \dif x \\
        & - \beta \, \left[ P^{[i]} \right]^{-1} m^{[i]} \int \Big[ q^{[i]}(x) \Big]^{\beta} \Big[ \exp \left\{ V^{[i+1]}(x) \right\} \Big]^{1 - \beta} \dif x
    \end{aligned} \\
    & = \beta \, \mathcal{Z}^{[i+1]} \left[ P^{[i]} \right]^{-1} \left[ \mathbb{E}^{[i+1]} \left[ x \right] - m^{[i]} \right].
\end{align}
After rearranging the terms, we get
\begin{align}
    \mathbb{E}^{[i+1]} \left[ x \right] & = m^{[i]} + \frac{1}{\beta} P^{[i]} \Big[ \mathcal{Z}^{[i+1]} \Big]^{-1} \frac{\partial \mathcal{Z}^{[i+1]}}{\partial \, m^{[i]}} \\
    & = m^{[i]} + \frac{1}{\beta} \, P^{[i]} \, \frac{\partial \log \mathcal{Z}^{[i+1]}}{\partial \, m^{[i]}},
\end{align}
which is an expression for the first moment of $q^{[i+1]}(x)$.

Next, we consider the second-order derivative $\mathcal{Z}^{[i+1]}$ with respect to $m^{[i]}$ 
\begin{align}
    \frac{\partial^{2} \mathcal{Z}^{[i+1]}}{\partial \, m^{[i]} \, \partial \left[ m^{[i]} \right]^{\top}} & = 
    \begin{aligned}[t]
        & \beta^{2} \, \left[ P^{[i]} \right]^{-1} \left[ \int (x - m^{[i]}) (x - m^{[i]})^{\top} \Big[ q^{[i]}(x) \Big]^{\beta} \Big[ \exp \left\{ V^{[i+1]}(x) \right\} \Big]^{1 - \beta} \dif x \right] \left[ P^{[i]} \right]^{-1} \\
        & - \beta \, \left[ P^{[i]} \right]^{-1} \int \Big[ q^{[i]}(x) \Big]^{\beta} \Big[ \exp \left\{ V^{[i+1]}(x) \right\} \Big]^{1 - \beta} \dif x
    \end{aligned} \\ 
    & = 
    \begin{aligned}[t]
        & \beta^{2} \, \mathcal{Z}^{[i+1]} \left[ P^{[i]} \right]^{-1} \mathbb{E}^{[i+1]} \left[ x \, x^{\top} \right] \left[ P^{[i]} \right]^{-1} + \beta^{2} \, \mathcal{Z}^{[i+1]} \left[ P^{[i]} \right]^{-1} m^{[i]} \, \Big[ m^{[i]} \Big]^{\top} \left[ P^{[i]} \right]^{-1} \\
        & - \beta^{2} \, \mathcal{Z}^{[i+1]} \left[ P^{[i]} \right]^{-1} \left[ \mathbb{E}^{[i+1]} \left[ x \right] \Big[ m^{[i]} \Big]^{\top} + m^{[i]} \mathbb{E}^{[i+1]} \left[ x \right]^{\top} \right] \left[ P^{[i]} \right]^{-1} - \beta \, \mathcal{Z}^{[i+1]} \left[ P^{[i]} \right]^{-1},\\[-0.5em]
    \end{aligned}
\end{align}
which leads to the second \emph{raw} moment of $q^{[i+1]}(x_{k})$
\begin{align}
    \mathbb{E}^{[i+1]} \left[ x \, x^{\top} \right] & = 
    \begin{aligned}[t]
        & - m^{[i]} \, \Big[ m^{[i]} \Big]^{\top} + \mathbb{E}^{[i+1]} \left[ x \right] \Big[ m^{[i]} \Big]^{\top} + m^{[i]} \mathbb{E}^{[i+1]} \left[ x \right]^{\top} \\
        & + \frac{1}{\beta} \, P^{[i]} + \frac{1}{\beta^{2}} \Big[ \mathcal{Z}^{[i+1]} \Big]^{-1} P^{[i]} \, \frac{\partial^{2} \mathcal{Z}^{[i+1]}}{\partial \, m^{[i]} \, \partial \left[ m^{[i]} \right]^{\top}} \, P^{[i]}.
    \end{aligned}
\end{align}
We can now derive the second \emph{central} moment as follows
\begin{align}
    \mathbb{V}^{[i+1]} \left[ x \right] & = \mathbb{E}^{[i+1]} \left[ x \, x^{\top} \right] - \mathbb{E}^{[i+1]} \left[ x \right] \mathbb{E}^{[i+1]} \left[ x \right]^{\top} \\ 
    & = 
    \begin{aligned}[t]
        & m^{[i]} \, \Big[ m^{[i]} \Big]^{\top} + \frac{1}{\beta} \, \Big[ \mathcal{Z}^{[i+1]} \Big]^{-1} \left[ P^{[i]} \, \frac{\partial \mathcal{Z}^{[i+1]}}{\partial \, m^{[i]}} \, \Big[ m^{[i]} \Big]^{\top} + m^{[i]} \, \left[ \frac{\partial \mathcal{Z}^{[i+1]}}{\partial \, m^{[i]}} \right]^{\top} P^{[i]} \right] + \frac{1}{\beta} \, P^{[i]} \\
        & + \frac{1}{\beta^{2}} \Big[ \mathcal{Z}^{[i+1]} \Big]^{-1} P^{[i]} \, \frac{\partial^{2} \mathcal{Z}^{[i+1]}}{\partial \, m^{[i]} \, \partial \left[ m^{[i]} \right]^{\top}} \, P^{[i]} - m^{[i]} \, \Big[ m^{[i]} \Big]^{\top} \\
        & - \frac{1}{\beta} \, \Big[ \mathcal{Z}^{[i+1]} \Big]^{-1} \left[ P^{[i]} \, \frac{\partial \mathcal{Z}^{[i+1]}}{\partial \, m^{[i]}} \, \Big[ m^{[i]} \Big]^{\top} + m^{[i]} \, \left[ \frac{\partial \mathcal{Z}^{[i+1]}}{\partial \, m^{[i]}} \right]^{\top} P^{[i]} \right] - \frac{1}{\beta^{2}} \, \Big[ \mathcal{Z}^{[i+1]} \Big]^{-2} P^{[i]} \, \frac{\partial \mathcal{Z}^{[i+1]}}{\partial\, m^{[i]}} \, \left[ \frac{\partial \mathcal{Z}^{[i+1]}}{\partial \, m^{[i]}} \right]^{\top} P^{[i]}
    \end{aligned} \\[-1em]
    & = \frac{1}{\beta} \, P^{[i]} + \frac{1}{\beta^{2}} \Big[ \mathcal{Z}^{[i+1]} \Big]^{-1} P^{[i]} \, \frac{\partial^{2} \mathcal{Z}^{[i+1]}}{\partial \, m^{[i]} \partial \left[ m^{[i]} \right]^{\top}} \, P^{[i]} - \frac{1}{\beta^{2}} \, \Big[ \mathcal{Z}^{[i+1]} \Big]^{-2} P^{[i]} \, \frac{\partial \mathcal{Z}^{[i+1]}}{\partial \, m^{[i]}} \left[ \frac{\partial \mathcal{Z}^{[i+1]}}{\partial \, m^{[i]}} \right]^{\top} P^{[i]} \\[-0.5em]
    & = \frac{1}{\beta} \, P^{[i]} + \frac{1}{\beta^{2}} \, P^{[i]} \frac{\partial^{2} \log \mathcal{Z}^{[i+1]}}{\partial \, m^{[i]} \, \partial \left[ m^{[i]} \right]^{\top}} \, P^{[i]}.
\end{align}

\section{Proof of Proposition~\ref{prop:forward-gauss-markov-conditionals-update}}
\label{app:forward-gauss-markov-conditionals-update-proof}
For a tilted forward-Markov conditional distribution of the form
\begin{equation}
    \rhki{q}{k}{i+1}(x_{k+1} \given x_{k}) = 
    \begin{aligned}[t]
        \left[ \rhki{\psi}{k}{i+1}(x_{k}) \right]^{-1} & \Big[ \rhki{q}{k}{i}(x_{k+1} \given x_{k}) \Big]^{\beta} \Big[ f_{k}(x_{k+1} \given x_{k}) \exp \left\{ \rhkixx{V}{k+1}{i+1}(x_{k+1}) \right\} \Big]^{1 - \beta}, 
    \end{aligned}
\end{equation}
with a normalizing function 
\begin{equation}
    \rhki{\psi}{k}{i+1}(x_{k}) = \int \Big[ \rhki{q}{k}{i}(x_{k+1} \given x_{k}) \Big]^{\beta} \Big[ f_{k}(x_{k+1} \given x_{k}) \exp \left\{ \rhkixx{V}{k+1}{i+1}(x_{k+1}) \right\} \Big]^{1 - \beta} \dif x_{k+1},
\end{equation}
where $\rhki{q}{k}{i}(x_{k+1} \given x_{k}) = \mathcal{N}(x_{k+1} \given \rhkix{F}{k}{i} \, x_{k} + \rhki{d}{k}{i}, \rhki{\Sigma}{k}{i})$ and $f_{k}(x_{k+1} \given x_{k}) \approx \exp \big\{ \ell_{f}^{[i]}(x_{k+1}, x_{k}) \big\}$
\begin{align}
    \ell_{f}^{[i]}(x_{k+1}, x_{k}) & \approx
    \begin{aligned}[t]
        - \frac{1}{2} 
        \begin{bmatrix} x_{k+1}^{\top} & x_{k}^{\top} \end{bmatrix}
        \begin{bmatrix} 
            C_{\bar x \bar x, k}^{[i]} & - C_{\bar x x, k}^{[i]} \\[0.5em]
            - C_{x \bar x, k}^{[i]} & C_{xx, k}^{[i]}
        \end{bmatrix}
        \begin{bmatrix} x_{k+1} \\[0.75em] x_{k} \end{bmatrix} 
        + \begin{bmatrix} x_{k+1}^{\top} & x_{k}^{\top} \end{bmatrix}
        \begin{bmatrix} c_{\bar x, k}^{[i]} \\[0.5em] c_{x, k}^{[i]} \end{bmatrix} + \kappa_{k}^{[i]}.
    \end{aligned}
\end{align}
We compute the \emph{conditional} moments of $\rhki{q}{k}{i+1}(x_{k+1} \given x_{k})$ from the derivatives of $\rhki{\psi}{k}{i+1}(x_{k})$ with respect to $x_{k}$. Starting with the first-order derivative
\begin{align}
    \frac{\partial \rhki{\psi}{k}{i+1}(x_{k})}{\partial x_{k}} & = 
    \begin{aligned}[t]
        & \beta \, \int  
        \begin{aligned}[t]
            & \frac{\partial \log \rhki{q}{k}{i}(x_{k+1} \given x_{k})}{\partial x_{k}} \, \Big[ \rhki{q}{k}{i}(x_{k+1} \given x_{k}) \Big]^{\beta} \\
            & \times \Big[ f_{k}(x_{k+1} \given x_{k}) \exp \left\{ \rhkixx{V}{k+1}{i+1}(x_{k+1}) \right\} \Big]^{1 - \beta} \dif x_{k+1}
        \end{aligned} \\
        & + (1 - \beta) \, \int 
        \begin{aligned}[t]
            & \frac{\partial \, \log f_{k}(x_{k+1} \given x_{k})}{\partial x_{k}} \, \Big[ \rhki{q}{k}{i}(x_{k+1} \given x_{k}) \Big]^{\beta} \\
            & \times \Big[ f_{k}(x_{k+1} \given x_{k}) \exp \left\{ \rhkixx{V}{k+1}{i+1}(x_{k+1}) \right\} \Big]^{1 - \beta} \dif x_{k+1},
        \end{aligned}
    \end{aligned} \\
    & = 
    \begin{aligned}[t]
        & \beta \, \int  
        \begin{aligned}[t]
            & \left[ \rhkix{F}{k}{i} \right]^{\top} \left[ \rhki{\Sigma}{k}{i} \right]^{-1} \left[x_{k+1} - \rhkix{F}{k}{i} \, x_{k} - \rhki{d}{k}{i} \right] \, \Big[ \rhki{q}{k}{i}(x_{k+1} \given x_{k}) \Big]^{\beta} \\
            & \times \Big[ f_{k}(x_{k+1} \given x_{k}) \exp \left\{ \rhkixx{V}{k+1}{i+1}(x_{k+1}) \right\} \Big]^{1 - \beta} \dif x_{k+1}
        \end{aligned} \\
        & + (1 - \beta) \, \int 
        \begin{aligned}[t]
            & \left[ - C_{xx, k}^{[i]} \, x_{k} + C_{x \bar x, k}^{[i]} \, x_{k+1} + c_{x, k}^{[i]} \right] \, \Big[ \rhki{q}{k}{i}(x_{k+1} \given x_{k}) \Big]^{\beta} \\
            & \times \Big[ f_{k}(x_{k+1} \given x_{k}) \exp \left\{ \rhkixx{V}{k+1}{i+1}(x_{k+1}) \right\} \Big]^{1 - \beta} \dif x_{k+1},
        \end{aligned}
    \end{aligned} \\
    & =
    \begin{aligned}[t]
        & \rhki{\psi}{k}{i+1}(x_{k}) \left[(1 - \beta) \, C_{x \bar x, k}^{[i]} + \beta \, \left[ \rhkix{F}{k}{i} \right]^{\top} \left[ \rhki{\Sigma}{k}{i} \right]^{-1} \right] \, \mathbb{E}^{[i+1]} \left[ x_{k+1} \given x_{k} \right] \\
        & - \rhki{\psi}{k}{i+1}(x_{k}) \left[ (1 - \beta) \, C_{xx, k}^{[i]} + \beta \left[ \rhkix{F}{k}{i} \right]^{\top} \left[ \rhki{\Sigma}{k}{i} \right]^{-1} \rhkix{F}{k}{i} \right] x_{k} \\
        & + \rhki{\psi}{k}{i+1}(x_{k}) \left[ (1 - \beta) \, c_{x, k}^{[i]} - \beta \, \left[ \rhkix{F}{k}{i} \right]^{\top} \left[ \rhki{\Sigma}{k}{i} \right]^{-1} \, \rhki{d}{k}{i} \right]
    \end{aligned} \\
    & = \rhki{\psi}{k}{i+1}(x_{k}) \, \rhki{G}{x \bar x, k}{i+1} \, \mathbb{E}^{[i+1]} \left[ x_{k+1} \given x_{k} \right] - \rhki{\psi}{k}{i+1}(x_{k}) \, \rhki{G}{xx, k}{i+1} \, x_{k} + \rhki{\psi}{k}{i+1}(x_{k}) \, \rhki{g}{x, k}{i+1},
\end{align}
where we use the definition of $\rhki{G}{x \bar x, k}{i+1}$, $\rhki{G}{xx, k}{i+1}$, and $\rhki{g}{x, k}{i+1}$ from Proposition~\ref{prop:forward-gauss-markov-potentials}.

\noindent Consequently, the first conditional moment of $\rhki{q}{k}{i+1}(x_{k+1} \given x_{k})$ is
\begin{equation}
    \mathbb{E}^{[i+1]} \left[ x_{k+1} \given x_{k} \right] = \Big[ \rhki{G}{x \bar x, k}{i+1} \Big]^{-1} \rhki{G}{xx, k}{i+1} \, x_{k} - \Big[ \rhki{G}{x \bar x, k}{i+1} \Big]^{-1} \rhki{g}{x, k}{i+1} + \Big[ \rhki{G}{x \bar x, k}{i+1} \Big]^{-1} \frac{\partial \log \rhki{\psi}{k}{i+1}(x_{k})}{\partial x_{k}}.
\end{equation}
Now we consider the second-order derivatives of $\rhki{\psi}{k}{i+1}(x_{k})$ with respect to $x_{k}$
\begin{align}
    \frac{\partial^{2} \rhki{\psi}{k}{i+1}(x_{k})}{\partial x_{k} \, \partial x_{k}^{\top}} & =  
    \begin{aligned}[t]
        & \beta^{2} \, \int 
        \begin{aligned}[t]
            & \left[ \rhkix{F}{k}{i} \right]^{\top} \left[ \rhki{\Sigma}{k}{i} \right]^{-1} \left[x_{k+1} - \rhkix{F}{k}{i} \, x_{k} - \rhki{d}{k}{i} \right] \Big[ \rhki{q}{k}{i}(x_{k+1} \given x_{k}) \Big]^{\beta} \\
            & \times \Big[ f_{k}(x_{k+1} \given x_{k}) \exp \left\{ \rhkixx{V}{k+1}{i+1}(x_{k+1}) \right\} \Big]^{1 - \beta} \\
            & \times \left[ x_{k+1} - \rhkix{F}{k}{i} \, x_{k} - \rhki{d}{k}{i} \right]^{\top} \left[ \rhki{\Sigma}{k}{i} \right]^{-1} \rhkix{F}{k}{i} \dif x_{k+1}
        \end{aligned} \\
        & + \beta \, (1 - \beta) \, \int 
        \begin{aligned}[t]
            & \left[ \rhkix{F}{k}{i} \right]^{\top} \left[ \rhki{\Sigma}{k}{i} \right]^{-1} \left[x_{k+1} - \rhkix{F}{k}{i} \, x_{k} - \rhki{d}{k}{i} \right] \Big[ \rhki{q}{k}{i}(x_{k+1} \given x_{k}) \Big]^{\beta} \\
            & \times \Big[ f_{k}(x_{k+1} \given x_{k}) \exp \left\{ \rhkixx{V}{k+1}{i+1}(x_{k+1}) \right\} \Big]^{1 - \beta} \\
            & \times \left[ - C_{xx, k}^{[i]} \, x_{k} + C_{x \bar x, k}^{[i]} \, x_{k+1} + c_{x, k}^{[i]} \right]^{\top} \dif x_{k+1}
        \end{aligned} \\
        & + (1 - \beta)^{2} \, \int 
        \begin{aligned}[t]
            & \left[ - C_{xx, k}^{[i]} \, x_{k} + C_{x \bar x, k}^{[i]} \, x_{k+1} + c_{x, k}^{[i]} \right] \, \Big[ \rhki{q}{k}{i}(x_{k+1} \given x_{k}) \Big]^{\beta} \\
            & \times \Big[ f_{k}(x_{k+1} \given x_{k}) \exp \left\{ \rhkixx{V}{k+1}{i+1}(x_{k+1}) \right\} \Big]^{1 - \beta} \\
            & \times \left[ - C_{xx, k}^{[i]} \, x_{k} + C_{x \bar x, k}^{[i]} \, x_{k+1} + c_{x, k}^{[i]} \right]^{\top} \dif x_{k+1}
        \end{aligned} \\
        & + \beta \, (1 - \beta) \, \int 
        \begin{aligned}[t]
            & \left[ - C_{xx, k}^{[i]} \, x_{k} + C_{x \bar x, k}^{[i]} \, x_{k+1} + c_{x, k}^{[i]} \right] \, \Big[ \rhki{q}{k}{i}(x_{k+1} \given x_{k}) \Big]^{\beta} \\
            & \times \Big[ f_{k}(x_{k+1} \given x_{k}) \exp \left\{ \rhkixx{V}{k+1}{i+1}(x_{k+1}) \right\} \Big]^{1 - \beta} \\
            & \times \left[ x_{k+1} - \rhkix{F}{k}{i} \, x_{k} - \rhki{d}{k}{i} \right]^{\top} \left[ \rhki{\Sigma}{k}{i} \right]^{-1} \rhkix{F}{k}{i} \dif x_{k+1}
        \end{aligned} \\
        & - \beta \, \rhki{\psi}{k}{i+1}(x_{k}) \, \left[ \rhkix{F}{k}{i} \right]^{\top} \left[ \rhki{\Sigma}{k}{i} \right]^{-1} \rhkix{F}{k}{i} - (1 - \beta) \, \rhki{\psi}{k}{i+1}(x_{k}) \, C_{xx, k}^{[i]}
    \end{aligned}
\end{align}

\begin{align}
    & = 
    \begin{aligned}[t]
        & \rhki{\psi}{k}{i+1}(x_{k}) \, \left[ \rhki{G}{x \bar x, k}{i+1} \right] \mathbb{E}^{[i+1]} \left[ x_{k+1} \, x_{k+1}^{\top} \given x_{k} \right] \left[ \rhki{G}{\bar x x, k}{i+1} \right] \\
        & + \rhki{\psi}{k}{i+1}(x_{k}) \, \left[ \rhki{G}{xx, k}{i+1} \right] \left[ x_{k} \, x_{k}^{\top} \right] \left[ \rhki{G}{xx, k}{i+1} \right]^{\top} \\
        & + \rhki{\psi}{k}{i+1}(x_{k}) \, \left[ \rhki{g}{x, k}{i+1} \right] \, \left[ \rhki{g}{x, k}{i+1} \right]^{\top} \\
        & - 2 \, \rhki{\psi}{k}{i+1}(x_{k}) \, \left[ \rhki{G}{x \bar x, k}{i+1} \right] \mathbb{E}^{[i+1]} \left[ x_{k+1} \given x_{k} \right] x_{k}^{\top} \left[ \rhki{G}{xx, k}{i+1} \right]^{\top} \\
        & + 2 \, \rhki{\psi}{k}{i+1}(x_{k}) \, \left[ \rhki{G}{x \bar x, k}{i+1} \right] \mathbb{E}^{[i+1]} \left[ x_{k+1} \given x_{k} \right] \left[ \rhki{g}{x, k}{i+1} \right]^{\top} \\
        & - 2 \, \rhki{\psi}{k}{i+1}(x_{k}) \, \left[ \rhki{G}{xx, k}{i+1} \right] x_{k} \left[ \rhki{g}{x, k}{i+1} \right]^{\top} \\
        & - \rhki{\psi}{k}{i+1}(x_{k}) \, \left[ \rhki{G}{xx, k}{i+1} \right].
    \end{aligned}
\end{align}

\noindent This gives us an expression for the second \emph{raw} conditional moment
\begin{equation}
    \mathbb{E}^{[i+1]} \left[ x_{k+1} \, x_{k+1}^{\top} \given x_{k} \right] = 
    \begin{aligned}[t]
        & \left[ \rhki{G}{x \bar x, k}{i+1} \right]^{-1} \left[ \rhki{G}{xx, k}{i+1} \right] \left[ \rhki{G}{\bar x x, k}{i+1} \right]^{-1} \\
        & + \left[ \rhki{\psi}{k}{i+1}(x_{k}) \right]^{-1} \, \left[ \rhki{G}{x \bar x, k}{i+1} \right]^{-1} \frac{\partial^{2} \rhki{\psi}{k}{i+1}(x_{k})}{\partial x_{k} \, \partial x_{k}^{\top}} \left[ \rhki{G}{\bar x x, k}{i+1} \right]^{-1} \\
        & - \left[ \rhki{G}{x \bar x, k}{i+1} \right]^{-1} \left[ \rhki{G}{xx, k}{i+1} \right] \left[ x_{k} \, x_{k}^{\top} \right] \left[ \rhki{G}{xx, k}{i+1} \right]^{\top} \left[ \rhki{G}{\bar x x, k}{i+1} \right]^{-1} \\
        & - \left[ \rhki{G}{x \bar x, k}{i+1} \right]^{-1} \left[ \rhki{g}{x, k}{i+1} \right] \, \left[ \rhki{g}{x, k}{i+1} \right]^{\top} \left[ \rhki{G}{\bar x x, k}{i+1} \right]^{-1} \\
        & + 2 \, \left[ \rhki{G}{x \bar x, k}{i+1} \right]^{-1} \left[ \rhki{G}{x \bar x, k}{i+1} \right] \mathbb{E}^{[i+1]} \left[ x_{k+1} \given x_{k} \right] x_{k}^{\top} \left[ \rhki{G}{xx, k}{i+1} \right]^{\top} \left[ \rhki{G}{\bar x x, k}{i+1} \right]^{-1} \\
        & - 2 \, \left[ \rhki{G}{x \bar x, k}{i+1} \right]^{-1} \left[ \rhki{G}{x \bar x, k}{i+1} \right] \mathbb{E}^{[i+1]} \left[ x_{k+1} \given x_{k} \right] \left[ \rhki{g}{x, k}{i+1} \right]^{\top} \left[ \rhki{G}{\bar x x, k}{i+1} \right]^{-1} \\
        & + 2 \, \left[ \rhki{G}{x \bar x, k}{i+1} \right]^{-1} \left[ \rhki{G}{xx, k}{i+1} \right] x_{k} \left[ \rhki{g}{x, k}{i+1} \right]^{\top} \left[ \rhki{G}{\bar x x, k}{i+1} \right]^{-1}.
    \end{aligned}
\end{equation}
Finally, we can derive the second \emph{central} moment according to
\begin{align}
    \mathbb{V}^{[i+1]} \left[ x_{k+1} \given x_{k} \right] & = \mathbb{E}^{[i+1]} \left[ x_{k+1} \, x_{k+1}^{\top} \given x_{k} \right] - \mathbb{E}^{[i+1]} \left[ x_{k+1} \given x_{k} \right] \mathbb{E}^{[i+1]} \left[ x_{k+1} \given x_{k} \right]^{\top} \\
    & = 
    \begin{aligned}[t]
        & \left[ \rhki{G}{x \bar x, k}{i+1} \right]^{-1} \left[ \rhki{G}{xx, k}{i+1} \right] \left[ \rhki{G}{\bar x x, k}{i+1} \right]^{-1} \\
        & + \left[ \rhki{\psi}{k}{i+1}(x_{k}) \right]^{-1} \, \left[ \rhki{G}{x \bar x, k}{i+1} \right]^{-1} \frac{\partial^{2} \rhki{\psi}{k}{i+1}(x_{k})}{\partial x_{k} \, \partial x_{k}^{\top}} \left[ \rhki{G}{\bar x x, k}{i+1} \right]^{-1} \\
        & - \left[ \rhki{\psi}{k}{i+1}(x_{k}) \right]^{-2} \Big[ \rhki{G}{x \bar x, k}{i+1} \Big]^{-1} \frac{\partial \rhki{\psi}{k}{i+1}(x_{k})}{\partial x_{k}} \Bigg[ \frac{\partial \rhki{\psi}{k}{i+1}(x_{k})}{\partial x_{k}} \Bigg]^{\top} \left[ \rhki{G}{\bar x x, k}{i+1} \right]^{-1}
    \end{aligned} \\
    & = \left[ \rhki{G}{x \bar x, k}{i+1} \right]^{-1} \left[ \rhki{G}{xx, k}{i+1} \right] \left[ \rhki{G}{\bar x x, k}{i+1} \right]^{-1} + \left[ \rhki{G}{x \bar x, k}{i+1} \right]^{-1} \frac{\partial^{2} \log \rhki{\psi}{k}{i+1}(x_{k})}{\partial x_{k} \, \partial x_{k}^{\top}} \left[ \rhki{G}{\bar x x, k}{i+1} \right]^{-1}.
\end{align}

\noindent When $\log \rhki{\psi}{k}{i+1}(x_{k})$ is evaluated under the quadratic approximation from Proposition~\ref{prop:forward-gauss-markov-potentials}, its derivatives are
\begin{equation}
    \frac{\partial \log \rhki{\psi}{k}{i+1}(x_{k})}{\partial x_{k}} = - \rhki{G}{xx, k}{i+1} \, x_{k} + \rhki{g}{x, k}{i+1} + \rhki{G}{x \bar x, k}{i+1} \Big[ \rhki{G}{\bar x \bar x, k}{i+1} \Big]^{-1} \left[ \rhki{G}{\bar x x, k}{i+1} \, x_{k} + \rhki{g}{\bar x, k}{i+1} \right],
\end{equation}
and
\begin{equation}
    \frac{\partial^{2} \log \rhki{\psi}{k}{i+1}(x_{k})}{\partial x_{k} \, \partial x_{k}^{\top}} = - \rhki{G}{xx, k}{i+1} + \rhki{G}{x \bar x, k}{i+1} \Big[ \rhki{G}{\bar x \bar x, k}{i+1} \Big]^{-1} \rhki{G}{\bar x x, k}{i+1}.
\end{equation}
Substituting these expressions into the derivative identities above gives
\begin{equation}
    \mathbb{E}^{[i+1]} \left[ x_{k+1} \given x_{k} \right] = \Big[ \rhki{G}{\bar x \bar x, k}{i+1} \Big]^{-1} \left[ \rhki{G}{\bar x x, k}{i+1} \, x_{k} + \rhki{g}{\bar x, k}{i+1} \right], \qquad     \mathbb{V}^{[i+1]} \left[ x_{k+1} \given x_{k} \right] = \Big[ \rhki{G}{\bar x \bar x, k}{i+1} \Big]^{-1}.
\end{equation}
Therefore, matching these moments to the affine-Gaussian conditional yields
\begin{equation}
    \rhki{\Sigma}{k}{i+1} = \Big[ \rhki{G}{\bar x \bar x, k}{i+1} \Big]^{-1}, \qquad
    \rhkix{F}{k}{i+1} = \Big[ \rhki{G}{\bar x \bar x, k}{i+1} \Big]^{-1} \, \rhki{G}{\bar x x, k}{i+1}, \qquad
    \rhki{d}{k}{i+1} = \Big[ \rhki{G}{\bar x \bar x, k}{i+1} \Big]^{-1} \, \rhki{g}{\bar x, k}{i+1}.
\end{equation}
Substituting the definitions of $\rhki{G}{\bar x \bar x, k}{i+1}$, $\rhki{G}{\bar x x, k}{i+1}$, and $\rhki{g}{\bar x, k}{i+1}$ from Proposition~\ref{prop:forward-gauss-markov-potentials}, we obtain
\begin{equation}
    \begin{aligned}
        \rhki{\Sigma}{k}{i+1} & = \left[ (1 - \beta) \left[ C_{\bar x \bar x, k}^{[i]} + \rhki{R}{k+1}{i+1} \right] + \beta \left[ \rhki{\Sigma}{k}{i} \right]^{-1} \right]^{-1}, \\
        \rhkix{F}{k}{i+1} & = \rhki{\Sigma}{k}{i+1} \left[ (1 - \beta) \, C_{\bar x x, k}^{[i]} + \beta \left[ \rhki{\Sigma}{k}{i} \right]^{-1} \rhkix{F}{k}{i} \right], \\
        \rhki{d}{k}{i+1} & = \rhki{\Sigma}{k}{i+1} \left[ (1 - \beta) \left[ c_{\bar x, k}^{[i]} + \rhki{r}{k+1}{i+1} \right] + \beta \left[ \rhki{\Sigma}{k}{i} \right]^{-1} \rhki{d}{k}{i} \right].
    \end{aligned}
\end{equation}

\section{Proof of Proposition~\ref{prop:reverse-gauss-markov-conditionals-update}}
\label{app:reverse-gauss-markov-conditionals-update-proof}
For a tilted reverse-Markov conditional distribution of the form
\begin{equation}
    \lhki{q}{k}{i+1}(x_{k-1} \given x_{k}) = 
    \begin{aligned}[t]
        \left[ \lhki{\psi}{k}{i+1}(x_{k}) \right]^{-1} & \Big[ \lhki{q}{k}{i}(x_{k-1} \given x_{k}) \Big]^{\beta} \Big[ f_{k-1}(x_{k} \given x_{k-1}) \exp \left\{ \lhkixx{V}{k-1}{i+1}(x_{k-1}) \right\} \Big]^{1 - \beta}, 
    \end{aligned}
\end{equation}
with a normalizing function 
\begin{equation}
    \lhki{\psi}{k}{i+1}(x_{k}) = \int \Big[ \lhki{q}{k}{i}(x_{k-1} \given x_{k}) \Big]^{\beta} \Big[ f_{k-1}(x_{k} \given x_{k-1}) \exp \left\{ \lhkixx{V}{k-1}{i+1}(x_{k-1}) \right\} \Big]^{1 - \beta} \dif x_{k-1},
\end{equation}
where $\lhki{q}{k}{i}(x_{k-1} \given x_{k}) = \mathcal{N}(x_{k-1} \given \lhkix{F}{k}{i} \, x_{k} + \lhki{d}{k}{i}, \lhki{\Sigma}{k}{i})$ and $f_{k-1}(x_{k} \given x_{k-1}) \approx \exp \big\{ \ell_{f}^{[i]}(x_{k}, x_{k-1}) \big\}$
\begin{align}
    \ell_{f}^{[i]}(x_{k}, x_{k-1}) & \approx
    \begin{aligned}[t]
        - \frac{1}{2} 
        \begin{bmatrix} x_{k}^{\top} & x_{k-1}^{\top} \end{bmatrix}
        \begin{bmatrix}
            C_{\bar x \bar x, k-1}^{[i]} & - C_{\bar x x, k-1}^{[i]} \\[0.5em]
            - C_{x \bar x, k-1}^{[i]} & C_{xx, k-1}^{[i]}
        \end{bmatrix}
        \begin{bmatrix} x_{k} \\[0.75em] x_{k-1} \end{bmatrix} 
        + \begin{bmatrix} x_{k}^{\top} & x_{k-1}^{\top} \end{bmatrix}
        \begin{bmatrix} c_{\bar x, k-1}^{[i]} \\[0.5em] c_{x, k-1}^{[i]} \end{bmatrix} + \kappa_{k-1}^{[i]}.
    \end{aligned}
\end{align}
The following steps are analogous to those described in Appendix~\ref{app:forward-gauss-markov-conditionals-update-proof}, thus we will omit redundant steps to avoid repetition. Starting with the first-order derivative of $\lhki{\psi}{k}{i+1}(x_{k})$
\begin{align}
    \frac{\partial \lhki{\psi}{k}{i+1}(x_{k})}{\partial x_{k}} & = 
    \begin{aligned}[t]
        & \lhki{\psi}{k}{i+1}(x_{k}) \left[(1 - \beta) \, C_{\bar x x, k-1}^{[i]} + \beta \, \left[ \lhkix{F}{k}{i} \right]^{\top} \left[ \lhki{\Sigma}{k}{i} \right]^{-1} \right] \, \mathbb{E}^{[i+1]} \left[ x_{k-1} \given x_{k} \right] \\
        & - \lhki{\psi}{k}{i+1}(x_{k}) \left[ (1 - \beta) \, C_{\bar x \bar x, k-1}^{[i]} + \beta \left[ \lhkix{F}{k}{i} \right]^{\top} \left[ \lhki{\Sigma}{k}{i} \right]^{-1} \lhkix{F}{k}{i} \right] x_{k} \\
        & + \lhki{\psi}{k}{i+1}(x_{k}) \left[ (1 - \beta) \, c_{\bar x, k-1}^{[i]} - \beta \, \left[ \lhkix{F}{k}{i} \right]^{\top} \left[ \lhki{\Sigma}{k}{i} \right]^{-1} \, \lhki{d}{k}{i} \right]
    \end{aligned} \\
    & = \lhki{\psi}{k}{i+1}(x_{k}) \, \lhki{G}{\bar x x, k}{i+1} \, \mathbb{E}^{[i+1]} \left[ x_{k-1} \given x_{k} \right] - \lhki{\psi}{k}{i+1}(x_{k}) \, \lhki{G}{\bar x \bar x, k}{i+1} \, x_{k} + \lhki{\psi}{k}{i+1}(x_{k}) \, \lhki{g}{\bar x, k}{i+1},
\end{align}
where we use the definition of $\lhki{G}{\bar x x, k}{i+1}$, $\lhki{G}{\bar x \bar x, k}{i+1}$, and $\lhki{g}{\bar x, k}{i+1}$ from Proposition~\ref{prop:reverse-gauss-markov-potentials}. Consequently, the first conditional moment of $\lhki{q}{k}{i+1}(x_{k-1} \given x_{k})$ is
\begin{equation}
    \mathbb{E}^{[i+1]} \left[ x_{k-1} \given x_{k} \right] = \Big[ \lhki{G}{\bar x x, k}{i+1} \Big]^{-1} \lhki{G}{\bar x \bar x, k}{i+1} \, x_{k} - \Big[ \lhki{G}{\bar x x, k}{i+1} \Big]^{-1} \lhki{g}{\bar x, k}{i+1} + \Big[ \lhki{G}{\bar x x, k}{i+1} \Big]^{-1} \frac{\partial \log \lhki{\psi}{k}{i+1}(x_{k})}{\partial x_{k}}.
\end{equation}
Now we consider the second-order derivatives of $\lhki{\psi}{k}{i+1}(x_{k})$ with respect to $x_{k}$
\begin{align}
    \frac{\partial^{2} \lhki{\psi}{k}{i+1}(x_{k})}{\partial x_{k} \, \partial x_{k}^{\top}} & =  
    \begin{aligned}[t]
        & \lhki{\psi}{k}{i+1}(x_{k}) \, \left[ \lhki{G}{\bar x x, k}{i+1} \right] \mathbb{E}^{[i+1]} \left[ x_{k-1} \, x_{k-1}^{\top} \given x_{k} \right] \left[ \lhki{G}{x \bar x, k}{i+1} \right] \\
        & + \lhki{\psi}{k}{i+1}(x_{k}) \, \left[ \lhki{G}{\bar x \bar x, k}{i+1} \right] \left[ x_{k} \, x_{k}^{\top} \right] \left[ \lhki{G}{\bar x \bar x, k}{i+1} \right]^{\top} \\
        & + \lhki{\psi}{k}{i+1}(x_{k}) \, \left[ \lhki{g}{\bar x, k}{i+1} \right] \, \left[ \lhki{g}{\bar x, k}{i+1} \right]^{\top} \\
        & - 2 \, \lhki{\psi}{k}{i+1}(x_{k}) \, \left[ \lhki{G}{\bar x x, k}{i+1} \right] \mathbb{E}^{[i+1]} \left[ x_{k-1} \given x_{k} \right] x_{k}^{\top} \left[ \lhki{G}{\bar x \bar x, k}{i+1} \right]^{\top} \\
        & + 2 \, \lhki{\psi}{k}{i+1}(x_{k}) \, \left[ \lhki{G}{\bar x x, k}{i+1} \right] \mathbb{E}^{[i+1]} \left[ x_{k-1} \given x_{k} \right] \left[ \lhki{g}{\bar x, k}{i+1} \right]^{\top} \\
        & - 2 \, \lhki{\psi}{k}{i+1}(x_{k}) \, \left[ \lhki{G}{\bar x \bar x, k}{i+1} \right] x_{k} \left[ \lhki{g}{\bar x, k}{i+1} \right]^{\top} \\
        & - \lhki{\psi}{k}{i+1}(x_{k}) \, \left[ \lhki{G}{\bar x \bar x, k}{i+1} \right].
    \end{aligned}
\end{align}

\noindent This gives us an expression for the second \emph{raw} conditional moment
\begin{equation}
    \mathbb{E}^{[i+1]} \left[ x_{k-1} \, x_{k-1}^{\top} \given x_{k} \right] = 
    \begin{aligned}[t]
        & \left[ \lhki{G}{\bar x x, k}{i+1} \right]^{-1} \left[ \lhki{G}{\bar x \bar x, k}{i+1} \right] \left[ \lhki{G}{x \bar x, k}{i+1} \right]^{-1} \\
        & + \left[ \lhki{\psi}{k}{i+1}(x_{k}) \right]^{-1} \, \left[ \lhki{G}{\bar x x, k}{i+1} \right]^{-1} \frac{\partial^{2} \lhki{\psi}{k}{i+1}(x_{k})}{\partial x_{k} \, \partial x_{k}^{\top}} \left[ \lhki{G}{x \bar x, k}{i+1} \right]^{-1} \\
        & - \left[ \lhki{G}{\bar x x, k}{i+1} \right]^{-1} \left[ \lhki{G}{\bar x \bar x, k}{i+1} \right] \left[ x_{k} \, x_{k}^{\top} \right] \left[ \lhki{G}{\bar x \bar x, k}{i+1} \right]^{\top} \left[ \lhki{G}{x \bar x, k}{i+1} \right]^{-1} \\
        & - \left[ \lhki{G}{\bar x x, k}{i+1} \right]^{-1} \left[ \lhki{g}{\bar x, k}{i+1} \right] \, \left[ \lhki{g}{\bar x, k}{i+1} \right]^{\top} \left[ \lhki{G}{x \bar x, k}{i+1} \right]^{-1} \\
        & + 2 \, \left[ \lhki{G}{\bar x x, k}{i+1} \right]^{-1} \left[ \lhki{G}{\bar x x, k}{i+1} \right] \mathbb{E}^{[i+1]} \left[ x_{k-1} \given x_{k} \right] x_{k}^{\top} \left[ \lhki{G}{\bar x \bar x, k}{i+1} \right]^{\top} \left[ \lhki{G}{x \bar x, k}{i+1} \right]^{-1} \\
        & - 2 \, \left[ \lhki{G}{\bar x x, k}{i+1} \right]^{-1} \left[ \lhki{G}{\bar x x, k}{i+1} \right] \mathbb{E}^{[i+1]} \left[ x_{k-1} \given x_{k} \right] \left[ \lhki{g}{\bar x, k}{i+1} \right]^{\top} \left[ \lhki{G}{x \bar x, k}{i+1} \right]^{-1} \\
        & + 2 \, \left[ \lhki{G}{\bar x x, k}{i+1} \right]^{-1} \left[ \lhki{G}{\bar x \bar x, k}{i+1} \right] x_{k} \left[ \lhki{g}{\bar x, k}{i+1} \right]^{\top} \left[ \lhki{G}{x \bar x, k}{i+1} \right]^{-1}.
    \end{aligned}
\end{equation}
Finally, we can derive the second \emph{central} moment according to
\begin{align}
    \mathbb{V}^{[i+1]} \left[ x_{k-1} \given x_{k} \right] & = \mathbb{E}^{[i+1]} \left[ x_{k-1} \, x_{k-1}^{\top} \given x_{k} \right] - \mathbb{E}^{[i+1]} \left[ x_{k-1} \given x_{k} \right] \mathbb{E}^{[i+1]} \left[ x_{k-1} \given x_{k} \right]^{\top} \\
    & = 
    \begin{aligned}[t]
        & \left[ \lhki{G}{\bar x x, k}{i+1} \right]^{-1} \left[ \lhki{G}{\bar x \bar x, k}{i+1} \right] \left[ \lhki{G}{x \bar x, k}{i+1} \right]^{-1} \\
        & + \left[ \lhki{\psi}{k}{i+1}(x_{k}) \right]^{-1} \, \left[ \lhki{G}{\bar x x, k}{i+1} \right]^{-1} \frac{\partial^{2} \lhki{\psi}{k}{i+1}(x_{k})}{\partial x_{k} \, \partial x_{k}^{\top}} \left[ \lhki{G}{x \bar x, k}{i+1} \right]^{-1} \\
        & - \left[ \lhki{\psi}{k}{i+1}(x_{k}) \right]^{-2} \Big[ \lhki{G}{\bar x x, k}{i+1} \Big]^{-1} \frac{\partial \lhki{\psi}{k}{i+1}(x_{k})}{\partial x_{k}} \Bigg[ \frac{\partial \lhki{\psi}{k}{i+1}(x_{k})}{\partial x_{k}} \Bigg]^{\top} \left[ \lhki{G}{x \bar x, k}{i+1} \right]^{-1}
    \end{aligned} \\
    & = \left[ \lhki{G}{\bar x x, k}{i+1} \right]^{-1} \left[ \lhki{G}{\bar x \bar x, k}{i+1} \right] \left[ \lhki{G}{x \bar x, k}{i+1} \right]^{-1} + \left[ \lhki{G}{\bar x x, k}{i+1} \right]^{-1} \frac{\partial^{2} \log \lhki{\psi}{k}{i+1}(x_{k})}{\partial x_{k} \, \partial x_{k}^{\top}} \left[ \lhki{G}{x \bar x, k}{i+1} \right]^{-1}.
\end{align}

\noindent When $\log \lhki{\psi}{k}{i+1}(x_{k})$ is evaluated under the quadratic approximation from Proposition~\ref{prop:reverse-gauss-markov-potentials}, its derivatives are
\begin{equation}
    \frac{\partial \log \lhki{\psi}{k}{i+1}(x_{k})}{\partial x_{k}} = - \lhki{G}{\bar x \bar x, k}{i+1} \, x_{k} + \lhki{g}{\bar x, k}{i+1} + \lhki{G}{\bar x x, k}{i+1} \Big[ \lhki{G}{xx, k}{i+1} \Big]^{-1} \left[ \lhki{G}{x \bar x, k}{i+1} \, x_{k} + \lhki{g}{x, k}{i+1} \right],
\end{equation}
and
\begin{equation}
    \frac{\partial^{2} \log \lhki{\psi}{k}{i+1}(x_{k})}{\partial x_{k} \, \partial x_{k}^{\top}} = - \lhki{G}{\bar x \bar x, k}{i+1} + \lhki{G}{\bar x x, k}{i+1} \Big[ \lhki{G}{xx, k}{i+1} \Big]^{-1} \lhki{G}{x \bar x, k}{i+1}.
\end{equation}
Substituting these expressions into the derivative identities above gives
\begin{equation}
    \mathbb{E}^{[i+1]} \left[ x_{k-1} \given x_{k} \right] = \Big[ \lhki{G}{xx, k}{i+1} \Big]^{-1} \left[ \lhki{G}{x \bar x, k}{i+1} \, x_{k} + \lhki{g}{x, k}{i+1} \right], \qquad     \mathbb{V}^{[i+1]} \left[ x_{k-1} \given x_{k} \right] = \Big[ \lhki{G}{xx, k}{i+1} \Big]^{-1}.
\end{equation}
Therefore, matching these moments to the affine-Gaussian conditional yields
\begin{equation}
    \lhki{\Sigma}{k}{i+1} = \Big[ \lhki{G}{xx, k}{i+1} \Big]^{-1}, \qquad
    \lhkix{F}{k}{i+1} = \Big[ \lhki{G}{xx, k}{i+1} \Big]^{-1} \, \lhki{G}{x \bar x, k}{i+1}, \qquad
    \lhki{d}{k}{i+1} = \Big[ \lhki{G}{xx, k}{i+1} \Big]^{-1} \, \lhki{g}{x, k}{i+1}.
\end{equation}
Substituting the definitions of $\lhki{G}{xx, k}{i+1}$, $\lhki{G}{x \bar x, k}{i+1}$, and $\lhki{g}{x, k}{i+1}$ from Proposition~\ref{prop:reverse-gauss-markov-potentials}, we obtain
\begin{equation}
    \begin{aligned}
        \lhki{\Sigma}{k}{i+1} & =  \left[ (1 - \beta) \, \left[ C_{xx, k-1}^{[i]} + \lhki{R}{k-1}{i+1} \right] + \beta \left[ \lhki{\Sigma}{k}{i} \right]^{-1} \right]^{-1}, \\
        \lhkix{F}{k}{i+1} & = \lhki{\Sigma}{k}{i+1} \left[ (1 - \beta) \, C_{x \bar x, k-1}^{[i]} + \beta \left[ \lhki{\Sigma}{k}{i} \right]^{-1} \lhkix{F}{k}{i} \right], \\
        \lhki{d}{k}{i+1} & = \lhki{\Sigma}{k}{i+1} \left[ (1 - \beta) \, \left[ c_{x, k-1}^{[i]} + \lhki{r}{k-1}{i+1} \right] + \beta \left[ \lhki{\Sigma}{k}{i} \right]^{-1} \lhki{d}{k}{i} \right].
    \end{aligned}
\end{equation}

\section{Natural Gradient Interpretation}
\label{app:natural-gradient}

We establish the correspondence between Gaussian natural gradient updates and the Gauss--Markov posterior updates with Fourier--Hermite coefficients. We present the derivations for the results in Corollary~\ref{cor:forward-gauss-markov-marginal-update} and Proposition~\ref{prop:forward-gauss-markov-conditionals-update}, but the same reasoning applies to the results in Corollary~\ref{cor:reverse-gauss-markov-marginal-update} and Proposition~\ref{prop:reverse-gauss-markov-conditionals-update}. The natural gradient identities used below are standard results by~\citet{khan2015proximal} and \citet{khan2017conjugate}.

\paragraph{Gaussian natural gradient.}

Let $q(z)=\mathcal N(z\given m,P)$ be a Gaussian variational distribution and consider the following variational objective
\begin{equation}
    \mathcal{L}_g(q)=\mathbb{E}_q\left[g(z)\right] - \mathbb{E}_q\left[\log q(z)\right],  
\end{equation}
where $g$ is a placeholder log-potential function. Furthermore, let us define the natural and mean parameters of a Gaussian, respectively:
\begin{equation}
    \lambda = 
    \begin{bmatrix}
        P^{-1}m\\[2mm]
        -\tfrac12 P^{-1}
    \end{bmatrix},
    \qquad
    \mu =
    \begin{bmatrix}
        m\\[2mm]
        M
    \end{bmatrix},
\end{equation}
where $M = P + mm^\top$. The natural gradient of the objective with respect to $\lambda$ can be written in terms of the gradient with respect to $\mu$
\begin{equation}
    F(\lambda)^{-1}\nabla_{\!\lambda}\mathcal{L}_g =\nabla_{\!\mu}\mathcal{L}_g,
\end{equation}
where $F$ denotes the Fisher information matrix \citep{khan2015proximal,khan2017conjugate}. 

To evaluate the right-hand side, first consider the expected log-potential
$\mathbb{E}_q[g(z)]$. Since $\mu^{(1)}=m$ and $\mu^{(2)}=M=P+mm^\top$, the chain rule gives
\begin{equation}
    \nabla_{\!\mu^{(1)}} f = \nabla_{\!m} f - 2(\nabla_{\!P} f)m, 
    \qquad
    \nabla_{\!\mu^{(2)}} f = \nabla_{\!P}\,f.
\end{equation}
Using the same Gaussian integration-by-parts identities from Definition~\ref{def:fourier-hermite}, we have
\begin{equation}
    \nabla_{\!m}\,\mathbb{E}_q\left[g(z)\right]
    = \mathbb{E}_q\left[\nabla_{\!z} g(z)\right], \qquad 
    \nabla_{\!P}\,\mathbb{E}_q\left[g(z)\right]
    = \tfrac{1}{2} \mathbb{E}_q\left[\nabla_{\!z}^2 g(z)\right].
\end{equation}
It follows that
\begin{equation}
    \nabla_{\!\mu}\,\mathbb{E}_q\left[g(z)\right]
    =
    \begin{bmatrix}
        \mathbb{E}_q\left[\nabla_{\!z} g(z)\right]
        - \mathbb{E}_q\left[\nabla_{\!z}^2 g(z)\right]m \\[2mm]
        \tfrac12\mathbb{E}_q\left[\nabla_{\!z}^2 g(z)\right]
    \end{bmatrix}.
\end{equation}

Let us recall that the Fourier--Hermite coefficients in Definition~\ref{def:fourier-hermite} are given by
\begin{equation}
    U = -\mathbb{E}_{q}\left[\nabla_{\!z}^2 g(z)\right],
    \qquad
    u = \mathbb{E}_{q}\left[\nabla_{\!z} g(z)\right] + U m.
\end{equation}
Hence, we can now write
\begin{equation}
    \nabla_{\!\mu}\,\mathbb{E}_{q}\left[g(z)\right]
    =
    \begin{bmatrix}
        u\\[1mm]
        -\tfrac12 U
    \end{bmatrix}.
\end{equation}
The linear and quadratic Fourier--Hermite coefficients are the components of the gradient of the expected log-potential with respect to the Gaussian mean parameters. For the Gaussian entropy term,
\begin{equation}
    -\nabla_{\!\mu}\,\mathbb{E}_q\left[\log q(z)\right] = -\lambda,
\end{equation}
which in turn gives
\begin{equation}
    F(\lambda)^{-1}\nabla_{\!\lambda}\mathcal{L}_g
    =
    \begin{bmatrix}
        u\\[1mm]
        -\tfrac12 U
    \end{bmatrix}
    - \lambda.
\end{equation}
A natural gradient ascent step with step size $1-\beta$ consequently takes the form
\begin{equation}
    \begin{aligned}
        \lambda^{[i+1]}
        &= \lambda^{[i]} + (1-\beta) F(\lambda^{[i]})^{-1} \nabla_{\!\lambda} \mathcal{L}_g(\lambda^{[i]}) \\
        &= \beta\lambda^{[i]} + (1-\beta)
        \begin{bmatrix}
            u^{[i]}\\[1mm]
            -\tfrac12 U^{[i]}
        \end{bmatrix}.
    \end{aligned}
    \label{eq:natural-gradient-update}
\end{equation}
Substituting the Gaussian natural parameters and matching the components gives
\begin{equation}
\begin{aligned}
    P^{[i+1]} &= \left[ \beta \left[ P^{[i]} \right]^{-1} + (1-\beta) U^{[i]} \right]^{-1}, \\
    &= \left[ \beta \left[ P^{[i]} \right]^{-1} - (1-\beta)\,\mathbb{E}_{q^{[i]}}\left[\nabla_{\!z}^2 g(z)\right] \right]^{-1}, \\
    m^{[i+1]} &= P^{[i+1]} \left[ \beta \left[ P^{[i]} \right]^{-1} m^{[i]} + (1-\beta) u^{[i]} \right] \\
    &= m^{[i]} + (1-\beta) P^{[i+1]}\,\mathbb{E}_{q^{[i]}}\left[\nabla_{\!z} g(z)\right].
\end{aligned}
\end{equation}
Finally, let $\widehat{g}^{[i+1]}$ denote the Fourier--Hermite quadratic surrogate of $g$ computed under $q^{[i]}$
\begin{equation}
    \widehat{g}^{[i+1]}(z) = - \frac{1}{2} z^{\top} U^{[i]} \, z + z^{\top} u^{[i]} + \eta^{[i]}.
\end{equation}
The natural parameter update in~\eqref{eq:natural-gradient-update} is equivalent to geometric interpolation
\begin{equation}
    q^{[i+1]}(z) \propto \left[q^{[i]}(z)\right]^{\beta} \left[\exp\big\{\widehat{g}^{[i+1]}(z)\big\}\right]^{1-\beta}.
\end{equation}

\paragraph{Gaussian marginal.}

For the forward boundary marginal, let $z = x_{0}$. The Gaussian variational distribution is
\begin{equation}
    q^{[i]}(z) = \rhki{q}{0}{i}(z) = \mathcal{N}(z \given \rhki{m}{0}{i}, \rhkix{P}{0}{i}),
\end{equation}
and the corresponding quadratic log-potential is
\begin{equation}
    \widehat{g}^{[i+1]}(z) = \rhkixx{V}{0}{i+1}(z) = - \frac{1}{2} z^{\top} \rhki{R}{0}{i+1} \, z + z^{\top} \rhki{r}{0}{i+1} + \rhki{\rho}{0}{i+1}.
\end{equation}
Since this potential is quadratic, its Fourier--Hermite coefficients are
\begin{equation}
    U^{[i]} = \rhki{R}{0}{i+1}, \qquad u^{[i]} = \rhki{r}{0}{i+1}.
\end{equation}
Using $m^{[i]} = \rhki{m}{0}{i}$ and $P^{[i]} = \rhkix{P}{0}{i}$, the natural gradient step in~\eqref{eq:natural-gradient-update} recovers the covariance and mean updates from Corollary~\ref{cor:forward-gauss-markov-marginal-update}. The marginal update from Corollary~\ref{cor:reverse-gauss-markov-marginal-update} admits the same interpretation using the reverse marginal $\lhki{q}{\T}{i}(x_{\T})$ and potential function $\lhkixx{V}{\T}{i+1}(x_{\T})$.

\paragraph{Gaussian conditionals.}

For each $0 \leq k < T$, fix $x_{k}$ and let $z=x_{k+1}$. The Gaussian variational distribution is now the forward conditional
\begin{equation}
    q^{[i]}(z) = \rhki{q}{k}{i}(z \given x_{k}) = \mathcal{N}(z \given \rhkix{F}{k}{i} \, x_{k} + \rhki{d}{k}{i}, \rhki{\Sigma}{k}{i}),
\end{equation}
and the corresponding quadratic log-potential is
\begin{align}
    \widehat{g}_{k}^{[i+1]}(z; x_{k}) & = \ell_{f}^{[i]}(z, x_{k}) + \rhkixx{V}{k+1}{i+1}(z) \\
    & = - \frac{1}{2} z^{\top} \left[ C_{\bar x \bar x, k}^{[i]} + \rhki{R}{k+1}{i+1} \right] z + z^{\top} \left[ C_{\bar x x, k}^{[i]} \, x_{k} + c_{\bar x, k}^{[i]} + \rhki{r}{k+1}{i+1} \right] + \const.
\end{align}
The Fourier--Hermite coefficients with respect to $z$ are thus
\begin{equation}
    U_{k}^{[i]} = C_{\bar x \bar x, k}^{[i]} + \rhki{R}{k+1}{i+1}, \qquad u_{k}^{[i]}(x_{k}) = C_{\bar x x, k}^{[i]} \, x_{k} + c_{\bar x, k}^{[i]} + \rhki{r}{k+1}{i+1}.
\end{equation}
Keeping this log-potential fixed and applying~\eqref{eq:natural-gradient-update} in $z=x_{k+1}$ gives
\begin{equation}
    \begin{aligned}
        \left[ \rhki{\Sigma}{k}{i+1} \right]^{-1} & = \beta \left[ \rhki{\Sigma}{k}{i} \right]^{-1} + (1 - \beta) \, U_{k}^{[i]}, \\
        \left[ \rhki{\Sigma}{k}{i+1} \right]^{-1} \left[ \rhkix{F}{k}{i+1} \, x_{k} + \rhki{d}{k}{i+1} \right] & = \beta \left[ \rhki{\Sigma}{k}{i} \right]^{-1} \left[ \rhkix{F}{k}{i} \, x_{k} + \rhki{d}{k}{i} \right] + (1 - \beta) \, u_{k}^{[i]}(x_{k}).
    \end{aligned}
\end{equation}
Substituting the coefficients above and matching the linear and constant terms in $x_{k}$ recovers the updates for $\rhki{\Sigma}{k}{i+1}$, $\rhkix{F}{k}{i+1}$, and $\rhki{d}{k}{i+1}$ in Proposition~\ref{prop:forward-gauss-markov-conditionals-update}. The reverse conditional updates in Proposition~\ref{prop:reverse-gauss-markov-conditionals-update} follow by the same logic, with $x_{k-1}$ as the Gaussian variable and the corresponding reverse quantities.

\newpage

\section{Recursive Bayesian Inference Algorithms}

\vspace*{\fill}

\begin{algorithm}[H]
    \setstretch{1.5}
    \caption{Backward Recursion of the Forward Entropic Variational Smoother}
    \label{alg:forward-gauss-markov-backward-pass}
    \small
    \begin{algorithmic}[1]
        \Require 
             Damping: $\beta$, first Gaussian marginal: $\rhki{m}{0}{i}, \rhkix{P}{0}{i}$,
            \Statex forward affine-Gaussian conditionals: $\rhkix{F}{0:\T-1}{i}, \rhki{d}{0:\T-1}{i}, \rhki{\Sigma}{0:\T-1}{i}$, 
            \Statex log prior $\ell_{p}$: $L_{0}^{[i]}, l_{0}^{[i]}$, log measurement $\ell_{h}$: $L_{1:\T}^{[i]}, l_{1:\T}^{[i]}$, 
            \Statex log dynamics $\ell_{f}$: $C_{\bar x \bar x, 0:\T-1}^{[i]}, C_{\bar x x, 0:\T-1}^{[i]}, C_{xx, 0:\T-1}^{[i]}, c_{\bar x, 0:\T-1}^{[i]}, c_{x, 0:\T-1}^{[i]}$.
        
        \item[] 
        \State $\rhki{R}{\T}{i+1} = L_{\T}^{[i]}, \quad \rhki{r}{\T}{i+1} = l_{\T}^{[i]}$ \Comment{initialize backward recursion}
        \For{$k \gets T-1, \dots, 0$}
            \State $\rhki{G}{\bar x \bar x, k}{i+1} = (1 - \beta) \left[ C_{\bar x \bar x, k}^{[i]} + \rhki{R}{k+1}{i+1} \right] + \beta \left[ \rhki{\Sigma}{k}{i} \right]^{-1}$
            \State $\rhki{G}{xx, k}{i+1} = (1 - \beta) \, C_{xx, k}^{[i]} + \beta \left[ \rhkix{F}{k}{i} \right]^{\top} \left[ \rhki{\Sigma}{k}{i} \right]^{-1} \rhkix{F}{k}{i}$
            \State $\rhki{G}{\bar x x, k}{i+1} = (1 - \beta) \, C_{\bar x x, k}^{[i]} + \beta \left[ \rhki{\Sigma}{k}{i} \right]^{-1} \rhkix{F}{k}{i}$
            \State $\rhki{g}{\bar x, k}{i+1} = (1 - \beta) \left[ c_{\bar x, k}^{[i]} + \rhki{r}{k+1}{i+1} \right] + \beta \left[ \rhki{\Sigma}{k}{i} \right]^{-1} \rhki{d}{k}{i}$
            \State $\rhki{g}{x, k}{i+1} = (1 - \beta) \, c_{x, k}^{[i]} - \beta \left[ \rhkix{F}{k}{i} \right]^{\top} \left[ \rhki{\Sigma}{k}{i} \right]^{-1} \rhki{d}{k}{i}$
            
            \item[]
            \State $\rhki{S}{k}{i+1} = \rhki{G}{xx, k}{i+1} - \left[ \rhki{G}{\bar x x, k}{i+1} \right]^{\top} \left[ \rhki{G}{\bar x \bar x, k}{i+1} \right]^{-1} \rhki{G}{\bar x x, k}{i+1}$ \Comment{update log-normalizers}
            \State $\rhki{s}{k}{i+1} = \rhki{g}{x, k}{i+1} + \left[ \rhki{G}{\bar x x, k}{i+1} \right]^{\top} \left[ \rhki{G}{\bar x \bar x, k}{i+1} \right]^{-1} \rhki{g}{\bar x, k}{i+1}$

            \item[]
            \State $\rhki{R}{k}{i+1} = L_{k}^{[i]} + 1 / (1 - \beta) \, \rhki{S}{k}{i+1}$ \Comment{update potential functions}
            \State $\rhki{r}{k}{i+1} = l_{k}^{[i]} + 1 / (1 - \beta) \, \rhki{s}{k}{i+1}$

            \item[]
            \State $\rhkix{F}{k}{i+1} = \Big[ \rhki{G}{\bar x \bar x, k}{i+1} \Big]^{-1} \, \rhki{G}{\bar x x, k}{i+1}$ \Comment{update conditional posteriors}
            \State $\rhki{d}{k}{i+1} = \Big[ \rhki{G}{\bar x \bar x, k}{i+1} \Big]^{-1} \, \rhki{g}{\bar x, k}{i+1}$
            \State $\rhki{\Sigma}{k}{i+1} = \Big[ \rhki{G}{\bar x \bar x, k}{i+1} \Big]^{-1}$
        \EndFor



        \item[]
        \State $\rhki{m}{0}{i+1} = \rhkix{P}{0}{i+1} \, \left[ (1 - \beta) \, \rhki{r}{0}{i+1} + \beta \left[ \rhkix{P}{0}{i} \right]^{-1} \rhki{m}{0}{i} \right]$ \Comment{update boundary marginal}
        \State $\rhkix{P}{0}{i+1} = \left[ (1 - \beta) \, \rhki{R}{0}{i+1} + \beta \left[ \rhkix{P}{0}{i} \right]^{-1} \right]^{-1}$

        \item[]
        \State \Return $\rhki{m}{0}{i+1}, \rhkix{P}{0}{i+1}$, $\rhkix{F}{0:\T-1}{i+1}, \rhki{d}{0:\T-1}{i+1}, \rhki{\Sigma}{0:\T-1}{i+1}, \rhkix{R}{0:\T}{i+1}, \rhki{r}{0:\T}{i+1}, \rhkix{S}{0:\T-1}{i+1}, \rhki{s}{0:\T-1}{i+1}$.
    \end{algorithmic}
\end{algorithm}

\vspace*{\fill}

\newpage 

\vspace*{\fill}

\begin{algorithm}
    \setstretch{1.5}
    \caption{Forward Recursion of the Forward Entropic Variational Smoother}
    \label{alg:forward-gauss-markov-forward-pass}
    \small
    \begin{algorithmic}[1]
        \Require 
            First Gaussian marginal: $\rhki{m}{0}{i+1}, \rhkix{P}{0}{i+1}$,
            \Statex forward affine-Gaussian conditionals: $\rhkix{F}{0:\T-1}{i+1}, \rhki{d}{0:\T-1}{i+1}, \rhki{\Sigma}{0:\T-1}{i+1}$.

        \For{$k \gets 0, \dots, T-1$}
            \State $\rhki{m}{k+1}{i+1} = \rhkix{F}{k}{i+1} \, \rhki{m}{k}{i+1} + \rhki{d}{k}{i+1}$ \Comment{update marginals}
            \State $\rhkix{P}{k+1}{i+1} = \rhkix{F}{k}{i+1} \, \rhkix{P}{k}{i+1} \, \left[ \rhkix{F}{k}{i+1} \right]^{\top} + \rhki{\Sigma}{k}{i+1}$
        \EndFor

        \State \Return $\rhki{m}{0:\T}{i+1}, \rhkix{P}{0:\T}{i+1}$.
    \end{algorithmic}
\end{algorithm}

\begin{algorithm}[H]
  \setstretch{1.25}
  \caption{Optimal Damping for the Forward Entropic Variational Smoother}
  \label{alg:forward-gauss-markov-optimal-damping}
  \small
  \begin{algorithmic}[1]
      \Require Maximum multiplier: $\alpha_{\text{max}}$, minimum multiplier: $\alpha_{\text{min}}$,
      \Statex initial multiplier: $\alpha_{0}$, first Gaussian marginal: $\rhki{m}{0}{i}, \rhkix{P}{0}
      {i}$,
      \Statex forward affine-Gaussian conditionals: $\rhkix{F}{0:\T-1}{i}, \rhki{d}{0:\T-1}{i},
      \rhki{\Sigma}{0:\T-1}{i}$,
      \Statex log prior $\ell_{p}$, log measurement $\ell_{h}$, log dynamics $\ell_{f}$.

      \State $\alpha \gets \alpha_{0}$
      \Repeat
          \State $\beta \gets \alpha / (1 + \alpha)$
          \vspace{0.25em}
          \State $\rhki{m}{0}{i+1}, \rhkix{P}{0}{i+1}, \rhkix{F}{0:\T-1}{i+1}, \rhki{d}{0:\T-1}{i+1},
          \rhki{\Sigma}{0:\T-1}{i+1}, \mathsf{feasible} \gets \textsc{Forward Conditionals}$
          \Comment{Algorithm~\ref{alg:forward-gauss-markov-backward-pass}}
          \vspace{0.25em}
          \If{$\mathsf{not\,\,feasible}$}
              \State $\alpha_{\text{min}} \gets \alpha$
              \State $\alpha \gets \sqrt{\alpha \cdot \alpha_{\text{max}}}$ \Comment{increase multiplier}
          \Else
              \State $\rhki{m}{0:\T}{i+1}, \rhkix{P}{0:\T}{i+1} \gets \textsc{Forward Marginals}$
              \Comment{Algorithm~\ref{alg:forward-gauss-markov-forward-pass}}
              \vspace{0.25em}
              \State $\Delta_{\mathrm{KL}} \gets \varepsilon - \mathbb{D}_{\mathrm{KL}} \big[ \rhki{q}
              {}{i+1} \ggiven \rhki{q}{}{i} \big]$
              \If{$\Delta_{\mathrm{KL}} > 0$}
                  \State $\alpha_{\text{max}} \gets \alpha$
                  \State $\alpha \gets \sqrt{\alpha \cdot \alpha_{\text{min}}}$ \Comment{reduce
                  multiplier}
              \ElsIf{$\Delta_{\mathrm{KL}} < 0$}
                  \State $\alpha_{\text{min}} \gets \alpha$
                  \State $\alpha \gets \sqrt{\alpha \cdot \alpha_{\text{max}}}$ \Comment{increase
                  multiplier}
              \EndIf
          \EndIf
      \Until{$\mathsf{feasible\,\,and\,\,} \Delta_{\mathrm{KL}} \approx 0$}
      \Comment{gradient vanishes}
      \State $\beta \gets \alpha / (1 + \alpha)$
      \State \Return $\beta$
  \end{algorithmic}
\end{algorithm}

\vspace*{\fill}

\newpage

\vspace*{\fill}

\begin{algorithm}[H]
    \setstretch{1.5}
    \caption{Forward Recursion of the Reverse Entropic Variational Smoother}
    \label{alg:reverse-gauss-markov-forward-pass}
    \small
    \begin{algorithmic}[1]
        \Require 
            Damping: $\beta$, last Gaussian marginal: $\lhki{m}{\T}{i}, \lhkix{P}{\T}{i}$,
            \Statex reverse affine-Gaussian conditionals: $\lhkix{F}{1:\T}{i}, \lhki{d}{1:\T}{i}, \lhki{\Sigma}{1:\T}{i}$, 
            \Statex log prior $\ell_{p}$: $L_{0}^{[i]}, l_{0}^{[i]}$, log measurement $\ell_{h}$: $L_{1:\T}^{[i]}, l_{1:\T}^{[i]}$, 
            \Statex log dynamics $\ell_{f}$: $C_{\bar x \bar x, 0:\T-1}^{[i]}, C_{\bar x x, 0:\T-1}^{[i]}, C_{xx, 0:\T-1}^{[i]}, c_{\bar x, 0:\T-1}^{[i]}, c_{x, 0:\T-1}^{[i]}$.
        
        \item[] 
        \State $\lhki{R}{0}{i+1} = L_{0}^{[i]}, \quad \lhki{r}{0}{i+1} = l_{0}^{[i]}$ \Comment{initialize forward recursion}
        \For{$k \gets 1, \dots, T$}
            \State $\lhki{G}{\bar x \bar x, k}{i+1} = (1 - \beta) \, C_{\bar x \bar x, k-1}^{[i]} + \beta \left[ \lhkix{F}{k}{i} \right]^{\top} \left[ \lhki{\Sigma}{k}{i} \right]^{-1} \lhkix{F}{k}{i}$
            \State $\lhki{G}{xx, k}{i+1} = (1 - \beta) \, \left[ C_{xx, k-1}^{[i]} + \lhki{R}{k-1}{i+1} \right] + \beta \left[ \lhki{\Sigma}{k}{i} \right]^{-1}$
            \State $\lhki{G}{x \bar x, k}{i+1} = (1 - \beta) \, C_{x \bar x, k-1}^{[i]} + \beta \left[ \lhki{\Sigma}{k}{i} \right]^{-1} \lhkix{F}{k}{i}$
            \State $\lhki{g}{\bar x, k}{i+1} = (1 - \beta) \, c_{\bar x, k-1}^{[i]} - \beta \left[ \lhkix{F}{k}{i} \right]^{\top} \left[ \lhki{\Sigma}{k}{i} \right]^{-1} \lhki{d}{k}{i}$
            \State $\lhki{g}{x, k}{i+1} = (1 - \beta) \, \left[ c_{x, k-1}^{[i]} + \lhki{r}{k-1}{i+1} \right] + \beta \left[ \lhki{\Sigma}{k}{i} \right]^{-1} \lhki{d}{k}{i}$
            
            \item[]
            \State $\lhki{S}{k}{i+1} = \lhki{G}{\bar x \bar x, k}{i+1} - \left[ \lhki{G}{x \bar x, k}{i+1} \right]^{\top} \left[ \lhki{G}{xx, k}{i+1} \right]^{-1} \lhki{G}{x \bar x, k}{i+1}$ \Comment{update log-normalizers}
            \State $\lhki{s}{k}{i+1} = \lhki{g}{\bar x, k}{i+1} + \left[ \lhki{G}{x \bar x, k}{i+1} \right]^{\top} \left[ \lhki{G}{xx, k}{i+1} \right]^{-1} \lhki{g}{x, k}{i+1}$

            \item[]
            \State $\lhki{R}{k}{i+1} = L_{k}^{[i]} + 1 / (1 - \beta) \, \lhki{S}{k}{i+1}$ \Comment{update potential functions}
            \State $\lhki{r}{k}{i+1} = l_{k}^{[i]} + 1 / (1 - \beta) \, \lhki{s}{k}{i+1}$

            \item[]
            \State $\lhkix{F}{k}{i+1} = \Big[ \lhki{G}{xx, k}{i+1} \Big]^{-1} \, \lhki{G}{x \bar x, k}{i+1}$ \Comment{update conditional posteriors}
            \State $\lhki{d}{k}{i+1} = \Big[ \lhki{G}{xx, k}{i+1} \Big]^{-1} \, \lhki{g}{x, k}{i+1}$
            \State $\lhki{\Sigma}{k}{i+1} = \Big[ \lhki{G}{xx, k}{i+1} \Big]^{-1}$
        \EndFor



        \item[]
        \State $\lhki{m}{\T}{i+1} = \lhkix{P}{\T}{i+1} \left[ (1 - \beta) \, \lhki{r}{\T}{i+1} + \beta \left[ \lhkix{P}{\T}{i} \right]^{-1} \lhki{m}{\T}{i} \right]$ \Comment{update boundary marginal}
        \State $\lhkix{P}{\T}{i+1} = \left[ (1 - \beta) \, \lhki{R}{\T}{i+1} + \beta \left[ \lhkix{P}{\T}{i} \right]^{-1} \right]^{-1}$

        \item[]
        \State \Return $\lhki{m}{\T}{i+1}, \lhkix{P}{\T}{i+1}, \lhkix{F}{1:\T}{i+1}, \lhki{d}{1:\T}{i+1}, \lhki{\Sigma}{1:\T}{i+1}, \lhkix{R}{0:\T}{i+1}, \lhki{r}{0:\T}{i+1}, \lhkix{S}{1:\T}{i+1}, \lhki{s}{1:\T}{i+1}$.
    \end{algorithmic}
\end{algorithm}

\vspace*{\fill}

\newpage

\vspace*{\fill}

\begin{algorithm}[H]
    \setstretch{1.5}
    \caption{Backward Recursion of the Reverse Entropic Variational Smoother}
    \label{alg:reverse-gauss-markov-backward-pass}
    \small
    \begin{algorithmic}[1]
        \Require 
            Last Gaussian marginal: $\lhki{m}{\T}{i+1}, \lhkix{P}{\T}{i+1}$,
            \Statex reverse affine-Gaussian conditionals: $\lhkix{F}{1:\T}{i+1}, \lhki{d}{1:\T}{i+1}, \lhki{\Sigma}{1:\T}{i+1}$.

        \For{$k \gets T, \dots, 1$}
            \State $\lhki{m}{k-1}{i+1} = \lhkix{F}{k}{i+1} \, \lhki{m}{k}{i+1} + \lhki{d}{k}{i+1}$ \Comment{update marginals}
            \State $\lhkix{P}{k-1}{i+1} = \lhkix{F}{k}{i+1} \, \lhkix{P}{k}{i+1} \, \left[ \lhkix{F}{k}{i+1} \right]^{\top} + \lhki{\Sigma}{k}{i+1}$
        \EndFor
        
        \State \Return $\lhki{m}{0:\T}{i+1}, \lhkix{P}{0:\T}{i+1}$.
    \end{algorithmic}
\end{algorithm}

\begin{algorithm}[H]
  \setstretch{1.25}
  \caption{Optimal Damping for the Reverse Entropic Variational Smoother}
  \label{alg:reverse-gauss-markov-optimal-damping}
  \small
  \begin{algorithmic}[1]
      \Require Maximum multiplier: $\alpha_{\text{max}}$, minimum multiplier: $\alpha_{\text{min}}$,
      \Statex initial multiplier: $\alpha_{0}$, last Gaussian marginal: $\lhki{m}{\T}{i}, \lhkix{P}
      {\T}{i}$,
      \Statex reverse affine-Gaussian conditionals: $\lhkix{F}{1:\T}{i}, \lhki{d}{1:\T}{i},
      \lhki{\Sigma}{1:\T}{i}$,
      \Statex log prior $\ell_{p}$, log measurement $\ell_{h}$, log dynamics $\ell_{f}$.

      \State $\alpha \gets \alpha_{0}$
      \Repeat
          \State $\beta \gets \alpha / (1 + \alpha)$
          \vspace{0.25em}
          \State $\lhki{m}{\T}{i+1}, \lhkix{P}{\T}{i+1}, \lhkix{F}{1:\T}{i+1}, \lhki{d}{1:\T}{i+1},
          \lhki{\Sigma}{1:\T}{i+1}, \mathsf{feasible} \gets \textsc{Reverse Conditionals}$
          \Comment{Algorithm~\ref{alg:reverse-gauss-markov-forward-pass}}
          \vspace{0.25em}
          \If{$\mathsf{not\,\,feasible}$}
              \State $\alpha_{\text{min}} \gets \alpha$
              \State $\alpha \gets \sqrt{\alpha \cdot \alpha_{\text{max}}}$ \Comment{increase
              multiplier}
          \Else
              \State $\lhki{m}{0:\T}{i+1}, \lhkix{P}{0:\T}{i+1} \gets \textsc{Reverse Marginals}$
              \Comment{Algorithm~\ref{alg:reverse-gauss-markov-backward-pass}}
              \vspace{0.25em}
              \State $\Delta_{\mathrm{KL}} \gets \varepsilon - \mathbb{D}_{\mathrm{KL}} \big[ \lhki{q}
              {}{i+1} \ggiven \lhki{q}{}{i} \big]$
              \If{$\Delta_{\mathrm{KL}} > 0$}
                  \State $\alpha_{\text{max}} \gets \alpha$
                  \State $\alpha \gets \sqrt{\alpha \cdot \alpha_{\text{min}}}$ \Comment{reduce
                  multiplier}
              \ElsIf{$\Delta_{\mathrm{KL}} < 0$}
                  \State $\alpha_{\text{min}} \gets \alpha$
                  \State $\alpha \gets \sqrt{\alpha \cdot \alpha_{\text{max}}}$ \Comment{increase
                  multiplier}
              \EndIf
          \EndIf
      \Until{$\mathsf{feasible\,\,and\,\,} \Delta_{\mathrm{KL}} \approx 0$}
      \Comment{gradient vanishes}
      \State $\beta \gets \alpha / (1 + \alpha)$
      \State \Return $\beta$
  \end{algorithmic}
\end{algorithm}

\vspace*{\fill}

\newpage

\vspace*{\fill}

\begin{algorithm}
    \setstretch{1.25}
    \caption{Forward Entropic Variational Smoother}
    \label{alg:forward-gauss-markov-iterated-smoother}
    \small
    \begin{algorithmic}[1]
        \Require 
            First Gaussian marginal: $\rhki{m}{0}{0}, \rhkix{P}{0}{0}$,
            \Statex forward affine-Gaussian conditionals: $\rhkix{F}{0:\T-1}{0}, \rhki{d}{0:\T-1}{0}, \rhki{\Sigma}{0:\T-1}{0}$.
            
        \State $i \gets 0$
        \While{not converged}
            \State $\rhki{m}{0:\T}{i}, \rhkix{P}{0:\T}{i} \gets \textsc{Forward Marginals}$ \Comment{Algorithm~\ref{alg:forward-gauss-markov-forward-pass}}
            \vspace{0.25em}
            \State $\ell_{p}, \ell_{h}, \ell_{f} \gets \textsc{Approximate Model}$ \Comment{Definition~\ref{def:gslr}/\ref{def:fourier-hermite}}
            \vspace{0.25em}
            \State $\beta \gets \textsc{Optimal Damping}$ \Comment{Algorithm~\ref{alg:forward-gauss-markov-optimal-damping}}
            \vspace{0.5em}
            \State $\rhki{m}{0}{i+1}, \rhkix{P}{0}{i+1}, \rhkix{F}{0:\T-1}{i+1}, \rhki{d}{0:\T-1}{i+1}, \rhki{\Sigma}{0:\T-1}{i+1} \gets$ $\textsc{Forward Conditionals}$ \Comment{Algorithm~\ref{alg:forward-gauss-markov-backward-pass}}
            \State $i \gets i + 1$
        \EndWhile
        \State \Return $\rhki{m}{0:\T}{\infty}, \rhkix{P}{0:\T}{\infty}$.
    \end{algorithmic}
\end{algorithm}

\begin{algorithm}
    \setstretch{1.25}
    \caption{Reverse Entropic Variational Smoother}
    \label{alg:reverse-gauss-markov-iterated-smoother}
    \small
    \begin{algorithmic}[1]
        \Require Last Gaussian marginal: $\lhki{m}{\T}{0}, \lhkix{P}{\T}{0}$,
        \Statex reverse affine-Gaussian conditionals: $\lhkix{F}{1:\T}{0}, \lhki{d}{1:\T}{0}, \lhki{\Sigma}{1:\T}{0}$.
            
        \State $i \gets 0$
        \While{not converged}
            \State $\lhki{m}{0:\T}{i}, \lhkix{P}{0:\T}{i} \gets \textsc{Reverse Marginals}$ \Comment{Algorithm~\ref{alg:reverse-gauss-markov-backward-pass}}
            \vspace{0.25em}
            \State $\ell_{p}, \ell_{h}, \ell_{f} \gets \textsc{Approximate Model}$ \Comment{Definition~\ref{def:gslr}/\ref{def:fourier-hermite}}
            \vspace{0.25em}
            \State $\beta \gets \textsc{Optimal Damping}$ \Comment{Algorithm~\ref{alg:reverse-gauss-markov-optimal-damping}}
            \vspace{0.5em}
            \State $\lhki{m}{\T}{i+1}, \lhkix{P}{\T}{i+1}, \lhkix{F}{1:\T}{i+1}, \lhki{d}{1:\T}{i+1}, \lhki{\Sigma}{1:\T}{i+1} \gets \textsc{Reverse Conditionals}$ \Comment{Algorithm~\ref{alg:reverse-gauss-markov-forward-pass}}
            \State $i \gets i + 1$
        \EndWhile
        \State \Return $\lhki{m}{0:\T}{\infty}, \lhkix{P}{0:\T}{\infty}$.
    \end{algorithmic}
\end{algorithm}

\vspace*{\fill}

\newpage

\vspace*{\fill}

\begin{algorithm}
    \setstretch{1.5}
    \caption{Marginals of the Hybrid Entropic Variational Smoother}
    \label{alg:hybrid-gauss-markov-marginals}
    \small
    \begin{algorithmic}[1]
        \Require 
            Damping: $\beta$, previous Gaussian marginal: $m_{1:\T-1}^{[i]}, P_{1:\T-1}^{[i]}$,
            \Statex forward log-normalizing functions: $\rhki{S}{1:\T-1}{i+1}, \rhki{s}{1:\T-1}{i+1}$, 
            \Statex reverse potential functions: $\lhki{R}{1:\T-1}{i+1}, \lhki{r}{1:\T-1}{i+1}$.

        \For{$k \gets 1, \dots, T-1$}
            \State $P_{k}^{[i+1]} = \left[ (1 - \beta) \left[ \lhki{R}{k}{i+1} + \rhki{S}{k}{i+1} \right] + \beta \left[ P_{k}^{[i]} \right]^{-1} \right]^{-1}$ \Comment{update marginals}
            \State $m_{k}^{[i+1]} = P_{k}^{[i+1]} \left[ (1 - \beta) \left[ \lhki{r}{k}{i+1} + \rhki{s}{k}{i+1} \right] + \beta \left[ P_{k}^{[i]} \right]^{-1} m_{k}^{[i]} \right]$
        \EndFor
        
        \State \Return $m_{1:\T-1}^{[i+1]}, P_{1:\T-1}^{[i+1]}$.
    \end{algorithmic}
\end{algorithm}

\begin{algorithm}
    \setstretch{1.35}
    \caption{Hybrid Entropic Variational Smoother}
    \label{alg:hybrid-gauss-markov-iterated-smoother}
    \small
    \begin{algorithmic}[1]
        \Require 
            Initial Gaussian marginals: $m_{0:\T}^{[0]}, P_{0:\T}^{[0]}$,
            \Statex forward affine-Gaussian conditionals: $\rhkix{F}{0:\T-1}{0}, \rhki{d}{0:\T-1}{0}, \rhki{\Sigma}{0:\T-1}{0}$,
            \Statex reverse affine-Gaussian conditionals: $\lhkix{F}{1:\T}{0}, \lhki{d}{1:\T}{0}, \lhki{\Sigma}{1:\T}{0}$.
            
        \State $i \gets 0$
        \While{not converged}
            \State $\ell_{p}, \ell_{h}, \ell_{f} \gets \textsc{Approximate Model}$ \Comment{Definition~\ref{def:gslr}/\ref{def:fourier-hermite}}
            \vspace{0.25em}
            \State $\beta \gets \textsc{Optimal Damping}$ 
            \Comment{Algorithm~\ref{alg:forward-gauss-markov-optimal-damping}/\ref{alg:reverse-gauss-markov-optimal-damping}}
            \vspace{0.25em}
            \vspace{0.25em}
            \State $m_{0}^{[i+1]}, P_{0}^{[i+1]}, \rhkix{F}{0:\T-1}{i+1}, \rhki{d}{0:\T-1}{i+1}, \rhki{\Sigma}{0:\T-1}{i+1}, \rhki{S}{1:\T-1}{i+1}, \rhki{s}{1:\T-1}{i+1} \gets$ $\textsc{Forward Normalizers}$ \Comment{Algorithm~\ref{alg:forward-gauss-markov-backward-pass}}
            \vspace{0.5em}
            \State $m_{\T}^{[i+1]}, P_{\T}^{[i+1]}, \lhkix{F}{1:\T}{i+1}, \lhki{d}{1:\T}{i+1}, \lhki{\Sigma}{1:\T}{i+1}, \lhki{R}{1:\T-1}{i+1}, \lhki{r}{1:\T-1}{i+1} \gets$ $\textsc{Reverse Potentials}$ \Comment{Algorithm~\ref{alg:reverse-gauss-markov-forward-pass}}
            \vspace{0.25em}
            \State $m_{1:\T-1}^{[i+1]}, P_{1:\T-1}^{[i+1]} \gets$ $\textsc{Hybrid Marginals}$ \Comment{Algorithm~\ref{alg:hybrid-gauss-markov-marginals}}
            \State $i \gets i + 1$
        \EndWhile
        \State \Return $m_{0:\T}^{[\infty]}, P_{0:\T}^{[\infty]}$.
    \end{algorithmic}
\end{algorithm}

\vspace*{\fill}

\end{document}